\documentclass[10pt,journal,compsoc]{IEEEtran}
\ifCLASSOPTIONcompsoc
  \usepackage[nocompress]{cite}
\else
  \usepackage{cite}
\fi
\ifCLASSINFOpdf
\else
\fi

\usepackage{threeparttable}

\usepackage[percent]{overpic}

\usepackage{xcolor}
\usepackage{tikz}
\definecolor{color1}{RGB}{16,70,127}
\definecolor{color2}{RGB}{246,178,147}
\definecolor{color3}{RGB}{109,1,31}

\usepackage[linesnumbered,ruled,vlined]{algorithm2e}
\usepackage{graphicx}
\usepackage{arydshln} 
\usepackage{booktabs}
\usepackage{multirow}
\usepackage{pifont}
\usepackage{color}
\usepackage{amssymb}
\usepackage{amsthm}
\usepackage{graphicx}
\usepackage{color}
\usepackage{dsfont}
\usepackage{algorithmicx}
\usepackage{algpseudocode}
\usepackage{tabularx}
\usepackage{multirow}
\usepackage{amsthm,amsmath,amssymb} 
\usepackage{mathrsfs}
\usepackage{array}
\usepackage{booktabs}  %
\usepackage{slashed}
\usepackage{scalerel}
\usepackage[top=0.4in,bottom=0.4in,left=0.4in,textwidth=7.7in]{geometry}
\usepackage{xspace}
\makeatletter 
\DeclareRobustCommand\onedot{\futurelet\@let@token\@onedot}
\def\@onedot{\ifx\@let@token.\else.\null\fi\xspace}

\def\eg{\emph{e.g}\onedot} 
\def\ie{\emph{i.e}\onedot} 
 
\def\etc{\emph{etc}\onedot} 
 
\def\etal{\emph{et al}\onedot}
\makeatother

\usepackage[pagebackref=false,breaklinks=true,linkcolor=red,anchorcolor=blue, citecolor=green,colorlinks,bookmarks=false]{hyperref}
\begin{document}

\title{A Reverse Causal Framework to Mitigate Spurious Correlations for Debiasing Scene Graph Generation}

\author{
Shuzhou Sun, Li Liu, Tianpeng Liu, Shuaifeng Zhi, Ming-Ming Cheng, Janne Heikkilä, Yongxiang Liu 
\IEEEcompsocitemizethanks{
\IEEEcompsocthanksitem This work was partially supported by the National Key Research and Development Program of China No. 2021YFB3100800, the Academy of Finland under grant 331883, Infotech Project FRAGES, and the National Natural Science Foundation of China under Grant 62376283 and 62201603. 
\IEEEcompsocthanksitem Li Liu, Tianpeng Liu, Shuaifeng Zhi, and Yongxiang Liu are with the College of Electronic Science and Technology, National University of Defense Technology (NUDT), Changsha, Hunan, China.  \protect \\
 Li Liu (dreamliu2010@gmail.com), Yongxiang Liu (lyxbible@sina.com), and Tianpeng Liu (everliutianpeng@sina.cn) are the corresponding authors.
\IEEEcompsocthanksitem Shuzhou Sun is with the Department of Computer Science \& Technology, Tsinghua University, Beijing, China, and also with the Center for Machine Vision and Signal Analysis (CMVS), University of Oulu, 90570 Oulu, Finland. Janne Heikkila is with the Center for Machine Vision and Signal Analysis (CMVS), University of Oulu, 90570 Oulu, Finland.
\IEEEcompsocthanksitem Ming-Ming Cheng is with the TKLNDST, College of Computer Science, Nankai University, Tianjin 300350, China.}
}

\markboth{IEEE Transactions on Pattern Analysis and Machine Intelligence}%
{Sun \MakeLowercase{\textit{et al.}}: USGG}

\IEEEtitleabstractindextext{%
\begin{abstract}
Existing two-stage Scene Graph Generation (SGG) frameworks typically incorporate a detector to extract relationship features and a classifier to categorize these relationships; therefore, the training paradigm follows a causal chain structure, where the detector's inputs determine the classifier's inputs, which in turn influence the final predictions. However, such a causal chain structure can yield spurious correlations between the detector's inputs and the final predictions, \ie, the prediction of a certain relationship may be influenced by other relationships. This influence can induce at least two observable biases: tail relationships are predicted as head ones, and foreground relationships are predicted as background ones; notably, the latter bias is seldom discussed in the literature. To address this issue, we propose reconstructing the causal chain structure into a reverse causal structure, wherein the classifier's inputs are treated as the confounder, and both the detector's inputs and the final predictions are viewed as causal variables. Specifically, we term the reconstructed causal paradigm as the \textbf{R}everse \textbf{c}ausal Framework for \textbf{SGG} (\textbf{RcSGG}). RcSGG initially employs the proposed Active Reverse Estimation (ARE) to intervene on the confounder to estimate the reverse causality, \ie the causality from final predictions to the classifier's inputs. Then, the Maximum Information Sampling (MIS) is suggested to enhance the reverse causality estimation further by considering the relationship information. Theoretically, RcSGG can mitigate the spurious correlations inherent in the SGG framework, subsequently eliminating the induced biases. Comprehensive experiments on popular benchmarks and diverse SGG frameworks show the state-of-the-art mean recall rate. 

\end{abstract}

\begin{IEEEkeywords}
Scene graph generation, causal inference, long-tailed distribution
\end{IEEEkeywords}}

\maketitle

\IEEEdisplaynontitleabstractindextext
\IEEEpeerreviewmaketitle

\IEEEraisesectionheading{\section{Introduction}\label{sec:introduction}}

Scene Graph Generation (SGG) is an essential task to bridge computer vision and natural language processing \cite{ImageRetrieval,SGGSurvey,wang2020visual,chen2024scene}. This task primarily focuses on producing a structured semantic representation of a scene, describing both the objects within the scene and their relationships. SGG offers a higher level understanding of a visual scene \cite{Neuralmotifs,RCNNSGG,VG150},  which is crucial for cognitive tasks. It, therefore,  has attracted significant attention due to its great potential for improving various downstream vision tasks, such as image-text retrieval\cite{ImageRetrieval,retrieval}, visual question answering \cite{VQA1,VQA2}, visual captioning \cite{imagecaptioning1,imagecaptioning2}, \etc.

\begin{figure*}
    \footnotesize\centering
    \centerline{\includegraphics[width=1\linewidth]{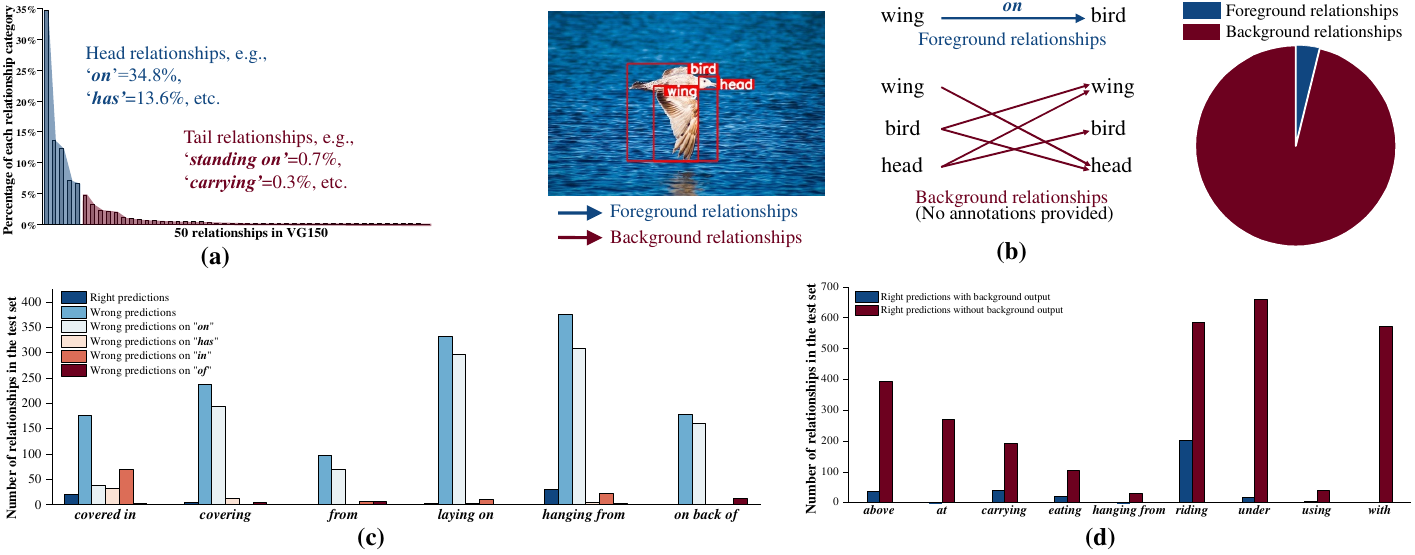}}
        \vspace{-0.3cm}
        \caption{The motivations of RcSGG. Images and count numbers are sourced from VG150. Test results were obtained from the MotifsNet model under the PredCls mode. (a) Distribution of head and tail relationships. Here, we approximate the top 50\% as head relationships (\textit{on}, \textit{has}, \textit{in}, \textit{of}, \textit{wearing}), with the remainder classified as tail relationships. (b) Foreground and background relationships in the scene graph generation task and their respective proportions over the entire dataset. (c) The number of right and wrong predictions, where we also showcase the number of instances incorrectly predicted as partial head relationships. (d) The number of right predictions under two scenarios: considering and not considering background relationships.}
    \label{motivation}
    \vspace{-0.2cm}
\end{figure*}

Despite considerable advancements in SGG task, the challenge of biased predictions persists as a relatively unaddressed concern. Within the body of research, such biased predictions typically refer to predicting fine-grained tail relationships as those belonging to coarse-grained head categories. The prevailing view attributes this to the imbalanced relationship distribution within SGG datasets. As illustrated in Figure \ref{motivation} (a), the relationships are dominated by a few head categories, exemplified by the ``\textit{on}'' relationship, which accounts for an extraordinary 34.8\% of cases. The impact of this skewed distribution is showcased in Figure \ref{motivation} (c), where it's observed that a significant portion of the tail relationships are predicted as head ones; for instance, ``\textit{covered in}'' is predicted as ``\textit{on}''.

For the biased predictions, substantial efforts have been made through various approaches, including resampling \cite{SegG,TransRwt}, reweighting \cite{FGPL,PPDL,GCL,EBMloss}, adjustment \cite{DLFE,TDE}, and hybrid strategies \cite{HML,NICE,RTPB,BPLSA}. The above methods emphasize that imbalanced relationship distributions can mislead the SGG model; however, they overlook relationship features that are not derived through independent and identical distribution (i.i.d.)  sampling within the relationship feature space but are aggregated from the image feature space (please refer to Section \ref{section3.2} and Figure \ref{spurious_correlations} (a) for a detailed analysis). As a result, these strategies inherently fail to address the gap of training relationship-level classifiers using image-level features, which, however, we believe, is pivotal to biased predictions.

In this paper, we go beyond merely skewed relationship distributions, initiating our inquiry from the perspective of the training framework to uncover the roots of biased predictions within SGG task. Conventional SGG frameworks typically employ a sequential arrangement, featuring an object detector followed by a relationship classifier, all within a two-stage training paradigm \cite{Neuralmotifs,RCNNSGG,VG150,VCtree}. In the first stage, an object detector pre-trained on the SGG dataset for object detection tasks, extracts relationship features $R$ from the input images $X$. These features $R$ typically encompass the features of the objects involved in the pairwise relationships and their bounding boxes, often cropped from the complete feature map to maintain uniformity in size, the process can be achieved by RoI Pooling as seen in Faster R-CNN \cite{fasterrcnn}. In the second stage, these extracted features $R$ serve as the foundation for training a relationship classifier. Note that during this classifier's training stage, the detector's parameters are kept frozen. Thus, our goal is to build an optimal mapping between the classifier's inputs (relationship features $R$) and outputs (final predictions $Y$).

However, the aforementioned decoupled SGG framework results in a semantic gap between training features and the target model, \ie, relationship features sampled from the image space are utilized to train the relationship classifier. Focusing on aggregated relationship features within the image domain compromises the independence of relationship predictions, meaning that predictions for one relationship in an image should not be influenced by others. In this paper, we formally define such compromised predictive independence as spurious correlations within the two-stage SGG framework: The relationship features employed in classifier training are aggregated within the image domain, thus causing predictions for a particular relationship to be influenced by other relationships within the same image; this influence is defined in this paper as spurious correlations from the input image to relationship predictions. Clearly, any imbalance in image domain aggregation exacerbates the impact of these spurious correlations. Unfortunately, existing SGG datasets typically exhibit two types of imbalance at the image domain level: 1) Head and tail relationship imbalance, as seen in Figure \ref{motivation}(a), and 2) Foreground-background relationship imbalance, as illustrated in Figure \ref{motivation}(b). Consequently, spurious correlations in SGG tasks lead to two types of prediction biases: 1) Predicting head relationships as tail relationships (Figure \ref{motivation}(c)), and 2) Predicting foreground relationships as background relationships (Figure \ref{motivation}(d)). To distinguish these biases, we refer to them respectively as \textbf{head-tail bias} and \textbf{fore-back bias}. Additionally, it is important to note that in SGG tasks, the image domain exhibits significant relationship co-occurrences. For instance, the probability of encountering ``\textit{mounted on}" and ``\textit{growing on}" within the same image is considerably higher than seeing ``\textit{mounted on}" and ``\textit{flying on}" together. Such co-occurrences can exacerbate the impact of spurious correlations, extensively analyzed in Section \ref{section3.2}. 


For head-tail bias and fore-back bias induced by spurious correlations, we resort to causal inference \cite{pearl2009causality,causalGeneralization,PearlPCH,pearlCBM}, an inference procedure that has demonstrated remarkable efficacy across fields such as statistics, econometrics, and epidemiology and has also gained considerable traction within the deep learning community in recent times. Our primary insight is that causal inference can model non-linear, high-order dependencies between variables \cite{luo2020causal, Bengio2021Toward}. In our approach, we initially build a Structural Causal Model (SCM) \cite{pearlCBM,CausalFairness} tailored for the SGG task. From previous discussions, it is clear that a two-stage training paradigm should adhere to the causal chain structure $X \rightarrow R \rightarrow Y$, where $\rightarrow$ denotes the direction of causality. Thus, the problem simplifies to focusing on the causality $R \rightarrow Y$, which is influenced by spurious correlations between $X$ and $Y$. Intuitively, ensuring statistical independence between $X$ and $Y$ within the chain structure $X \rightarrow R \rightarrow Y$ can effectively eliminate the spurious correlations. Inspired by reverse causality \cite{pearl2009causality}, we reconstruct the causal chain $X \rightarrow R \rightarrow Y$ into a reverse causal structure $X \rightarrow R \leftarrow Y$. In this configuration, the correlation between $X$ and $Y$ is eliminated, effectively preventing spurious correlations in SGG tasks. This restructuring is supported by the two-stage training paradigm where: 1) A static object detector extracts relationship features $R$ from image features $X$, which a relationship classifier then uses to predict labels $Y$. $R$ inherently carries predictive information about $Y$, shaped by the label-guided training process. Thus, the observed causal relationship $R \rightarrow Y$ can effectively be modeled in reverse as $Y \rightarrow R$, making $R$ a mediator between $X$ and $Y$. 2) Reverse causality \cite{pearl2009causality} posits that $Y \rightarrow R$ reconstructs the dependency when the outcome $Y$ can impact the intermediary $R$. In our framework, the distribution of $R$ is continuously refined based on the outcomes from $Y$ during the training of relationship classifier (as detailed in Section \ref{section_ARE}), aligning with the reverse causality's requirements for dependency reconstruction.

For the restructured framework $X \rightarrow R \leftarrow Y$, we need to estimate the reverse causality, \ie, $R \leftarrow Y$. To achieve this, we propose Active Reverse Estimation (ARE), inspired by the model-to-data training paradigm in Active Learning (AL) \cite{activelearning,AL_1,AL_2,EDAL}. AL emphasizes the impact of inference outcomes on training data, aligning with our reconstructed inverse relationship $R \leftarrow Y$. ARE first establishes a query condition to sample from the query set for optimizing the feature space, where the query set constitutes a subset of the entire relationship feature space. Specifically, the query condition iterates based on the loss during training, thereby ensuring adaptive optimization of the relationship feature space. Unfortunately, the SGG task confronts a dual bias dilemma, \ie, head-tail bias and fore-back bias. To address dual imbalance, our proposed Active Reverse Estimation (ARE) encompasses two stages: 1) Borrowing from active learning's model-to-data paradigm, each training iteration within our framework refines the space of foreground relations based on the prior iteration’s results, effectively addressing the challenge of head-tail bias. Additionally, SGG tasks face significant challenges due to dual imbalances: relationship and object imbalances, where object imbalances further lead to disparities among object pairs. Consequently, the ARE sampling tends to converge on prominent object pairs, as elaborated in Section \ref{MIS}. To counteract this issue, we introduce Maximum Information Sampling (MIS), which focuses on maximizing the information yield from sampling outcomes based on the distribution of object pairs. 2) For the optimized foreground relationship space, we selectively discard certain background relationships to mitigate fore-back bias, while our meticulously designed foreground-background preservation ratio enhances model training stability. Notably, our approach is designed to be seamlessly incorporated into any two-stage framework, offering a model-agnostic solution that supports both reverse causality estimation during training phase and forward causality inference during testing phase, as detailed in Section \ref{section3.4.2}.

In summary, the contributions of our work are three-fold:
\begin{itemize}
\item We thoroughly analyze the biased prediction problem in the SGG task, including the widely studied head-tail bias and the less-discussed fore-back bias. Importantly, we identify that both types of bias originate from the same underlying issue, \ie, spurious correlations.

\item We reconstruct the causal chain structure of the standard SGG paradigm into a reverse causal structure, denoted as RcSGG. RcSGG incorporates Active Reverse Estimation for reverse causality estimation, alongside employing Maximum Information Sampling to refine this estimation process. Theoretically, our method eradicates biased predictions induced by spurious correlations between input images and final outputs.
 
\item RcSGG attains state-of-the-art performance in terms of mean recall rate, as demonstrated by extensive experimental results on popular benchmarks and SGG frameworks, while also achieving a better trade-off between recall and mean recall rates.
\end{itemize}

\begin{figure*}
    \footnotesize\centering
    \begin{overpic}[width=1.0\linewidth]{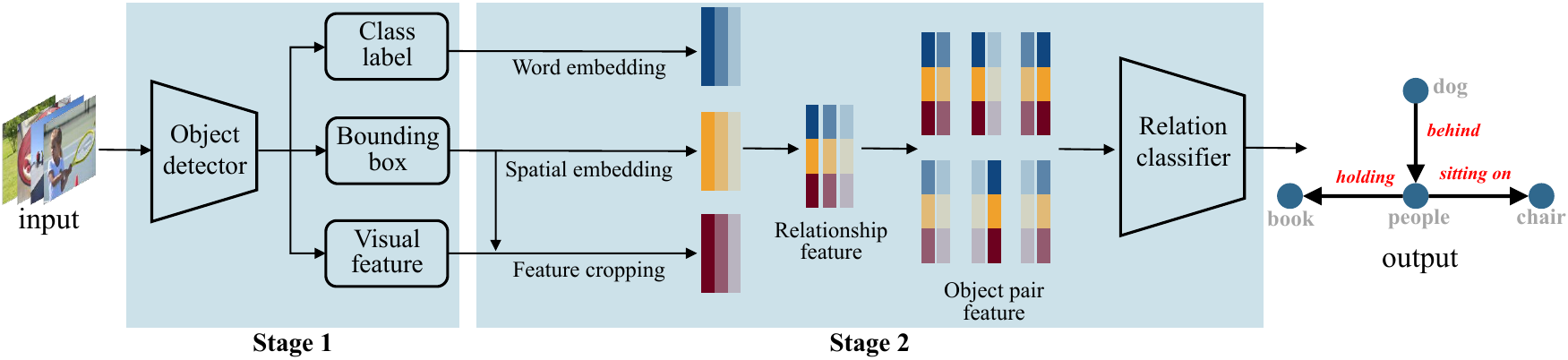}
        \put(1.5,5.5){ \large $X$}
        \put(12,7.5){ \large $f_o$}
        \put(51.5,4){ \large $R$}
        \put(74.5,7.2){ \large $f_c$}
        \put(89.5,3.4){ \large $Y$}
    \end{overpic}
    \vspace{-0.6cm}
        \caption{The standard pipeline of the two-stage scene graph generation framework, where the input image $X$ is processed by a pre-trained object detector to obtain relationship features $R$, followed by a relationship classifier to produce the final output $Y$.}
    \label{framework}
    \vspace{-0.3cm}
\end{figure*}
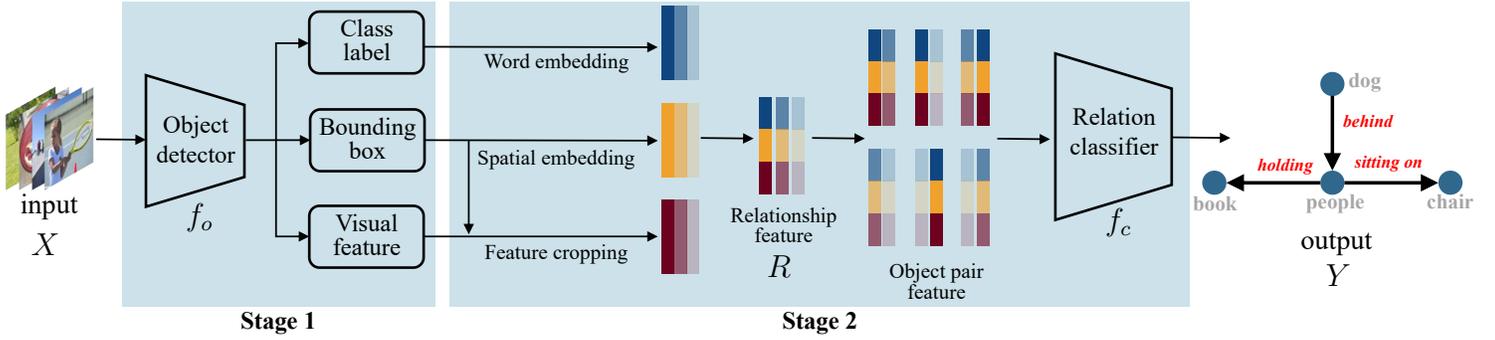

\section{Related works}
\subsection{Unbiased cene Graph Generation}
\label{sec:Related_SGG}
Scene Graph Generation (SGG) assigns a triplet relationship, such as \textit{\textless people, \textbf{sitting on}, chair\textgreater}, to each pair of objects in an image \cite{wang2020visual,chen2024scene}. Hence, compared to image classification \cite{AlexNet,VGG} and object detection \cite{fasterrcnn}, it facilitates a more profound understanding of the given scenes. Benefiting from this, SGG has shown promising utility across a spectrum of downstream applications, such as image-text retrieval \cite{ImageRetrieval,retrieval}, visual question answering \cite{VQA1,VQA2}, and visual captioning \cite{imagecaptioning1,imagecaptioning2}. However, the challenge of data imbalance within SGG datasets \cite{VG}, which can yield biased predictions, has prompted a shift in research focus towards developing debiasing strategies. Consequently, our discussion narrows to the examination of debiasing approaches within the SGG task, highlighting efforts to enhance the influence of tail relationships during model training. These efforts can broadly be categorized into four distinct categories:

\textit {Resampling methods} \cite{SegG,TransRwt,DT2} upsample tail relationships and/or downsample head ones to rebalance the training data distribution. For instance, Decoupled Training for Devil in the Tails (DT2) \cite{DT2} devised a novel sampling method, Alternating Class-Balanced Sampling (ACBS), to capture the interplay between long-tailed entities and the predicate distribution of visual relationships. 

\textit{Reweighting methods} \cite{FGPL,PPDL,GCL,ML-MWN,EBMloss} modify the contributions of various relationships during training, for instance, by weighting prediction losses to improve the model's ability to represent tail categories. Recently, Multi-Label Meta Weight Net (ML-MWN) \cite{ML-MWN} introduced a multi-label meta-learning framework to address biased relationship distributions, learning a meta-weight network for each training sample. 

\textit{Adjustment methods} \cite{DLFE,TDE} adjust the biased models to achieve debiased predictions, for example, by adjusting output logits to enhance the probability of neglected relationships during training. Dynamic Label Frequency Estimation (DLFE) \cite{DLFE} frames SGG as a Learning from Positive and Unlabeled data (PU learning) task and introduces Dynamic Label Frequency Estimation (DLFE) to refine label frequency estimates, leveraging training-time data augmentation over multiple iterations. 

\textit{Hybrid methods} \cite{HML,NICE,RTPB,BPLSA} typically integrate a subset or even all of the aforementioned strategies. For example, NICE \cite{NICE} initially detects noise in samples to reassign them high-quality soft predicate labels and incorporates a dynamic balancing weighting strategy during the knowledge distillation phase to penalize biases across different teacher networks.

Beyond the strategies previously outlined, some recent works \cite{PENET,DHL,MEET} enhance learning for tail relationships through network modifications to pursue debiased predictions. Furthermore, incorporating additional knowledge, especially from large models, shows promise in mitigating biases in SGG models. For instance, CaCao \cite{CaCao} enhances tail relationships using the pretrained language models. Recently, Semantic Granularity Controller (SGC) \cite{chen2023addressing} has been proposed to address predicate overlap by dynamically adjusting predicate granularity. While both SGC and RcSGG aim to mitigate biases in SGG, their focuses differ: SGC primarily targets predicate-level adjustments to resolve annotation inconsistencies, whereas RcSGG addresses spurious correlations by reconstructing the causal structure of the training paradigm. Although debiasing research is rather active in the SGG community, these approaches primarily emphasize the influence of the long-tail distribution on relationship classifier training while neglecting the fundamental source of bias: spurious correlations arising from the current two-stage training paradigm. Furthermore, the foreground-background imbalance is rarely considered in these methods. In contrast, our work traces the model bias from the causal perspective and proposes a debiased scene graph model, RcSGG. Our method estimates the reverse causality and thus can eliminate both head-tail and fore-back biases induced by spurious correlations within the two-stage training paradigm.

\subsection{Causal Inference}
\label{sec:Causal}
Causal inference investigates causal relationships between variables, employing experimental design and observational studies to mitigate confounding factors, and has achieved impressive performance in statistics, econometrics, and epidemiology \cite{pearl2009causality,pearlCBM,PearlPCH,Bengio2021Toward,Wang,DisC}. Causal inference has also attracted significant attention in the SGG community recently. Such approaches focus on the intrinsic features of the relationships through causal interventions, mitigating the effects of confounding variables, and thereby making debiased predictions even in the face of skewed distributions. For instance, TDE \cite{TDE} employs counterfactual estimation for the debiasing of the trained SGG model. DSDI \cite{DSDI} addresses dual imbalances in observed data by incorporating biased resistance loss and a causal intervention tree, consequently enhancing foreground representation and mitigating context bias. More recently, TsCM \cite{TsCM} introduced a two-stage causal learning approach: the first stage focuses on causal representation learning, eliminating the semantic confusion bias in the SGG task, while the second stage emphasizes causal calibration learning, addressing the long-tailed distribution bias. Compared to the aforementioned methods, our approach, RcSGG, emphasizes two key distinctions: 

\textbf{1) Source of bias.} Our work investigates biases from the perspective of model architecture. Specifically, we analyze the spurious correlations inherent in the prevalent two-stage training paradigm and identify two observable biases: head-tail bias and fore-back bias. Notably, the latter has received limited attention in the existing literature. In contrast, methods such as TsCM \cite{TsCM} focus on bias sources primarily at the data distribution level, such as the imbalance in relationship distributions.

\textbf{2) Bias mitigation.} RcSGG mitigates spurious correlations by reconstructing the causal chain into a reverse causal structure, effectively addressing the biases introduced by the two-stage training paradigm. In contrast, existing methods like TsCM \cite{TsCM} are predominantly data-driven, relying on statistical knowledge extracted from the training set, such as relationship populations and adjustment factors. These methods fail to address the root causes of bias and overlook architecture-specific biases.

\section{Methods}
\label{sec3}

\subsection{Overview} 
\label{section3.1}
Popular Scene Graph Generation (SGG) frameworks typically involve two subtasks: object detection and relationship classification \cite{Neuralmotifs,VCtree,TDE,ImageRetrieval,SGGSurvey}. As depicted in Figure \ref{framework}, the object detector, denoted as $f_o$, begins by extracting class labels, bounding boxes and visual features from the images $X$. Since $f_o$ is frozen, such information for an image remains fixed throughout the training process. Class labels and bounding boxes are embedded into vectors of specific dimensions, while visual features are cropped based on bounding boxes to represent vectors of individual objects, typically achieved through RoI pooling. These vectors are then jointly embedded into relationship features $R$. Subsequently, these features $R$ are paired and fed into the relationship classifier $f_c$ to produce the final output $Y$. However, it's worth emphasizing that the detector $f_o$ is typically pre-trained on the SGG dataset with the object detection task and remains frozen during the training of classifier $f_c$. Hence, the central goal of this typical two-stage SGG framework is to develop a high-performing relationship classifier $f_c$, with $f_o$ serving merely to extract relationship features from $X$ necessary for $f_c$'s training.

The typical two-stage SGG framework, refined for debiasing, primarily scrutinizes skewed annotations as the main source of model biases. However, these approaches overlook the model biases induced by the decoupled two-stage training regime, where ``decoupled'' refers to $f_o$ and $f_c$ undergoing separate training phases. Such biases are attributable to $f_c$ being trained with a skewed feature space, yet this non-ideal space is actually extracted from images by a pre-trained and subsequently frozen detector. Evidently, there exists a training gap here: $f_c$'s training input is relationship features. These features do not come as independently and identically distributed samples from the relationship space but are instead extracted from the image space, leading us to argue that this introduces spurious correlations between the input images and the final predictions.

Formally, in Section \ref{section3.2}, we discuss the origins of spurious correlations and how they induce model biases, especially the head-tail bias and fore-back bias. We further analyze why a potential solution, batch balance, is inappropriate for the SGG task. In Section \ref{section3.3}, we introduce a slightly tempered alternative, approximate batch balance; however, as a compensation for the approximation, we suggest leveraging a reverse causal estimation strategy to optimize the feature space in each batch. In Section \ref{section3.4}, we highlight how our method differentiates from reweighting and resampling techniques, and detail its integration of both reverse and forward causal estimations into any SGG framework, enhancing its utility for debiased predictions.

To clarify upcoming sections, we define key notations and symbols here. Let $\mathcal{D}= \{ (X, Y) \} = \{(\mathbf{x}_i, \mathbf{y}_i)\}_{i=1}^{N_{\mathcal{D}}}$ denote the observed data with $N_{\mathcal{D}}$ samples, where $\mathbf{x}_i$ is $i$-th instance in image set $X$; $\mathbf{y}_i \in \mathbb{R}^{N_i \times K}$ represents the $N_i$ relationships in the $i$-th image, and its element $\mathbf{y}_{i j} \in \mathbb{R}^{K}$ is a $K$ dimension one-hot vector denoting the label of $j$-th relationship in $\mathbf{x}_i$. For the observed data $\mathcal{D}$, we denote the extracted relationship features as $R=\{\mathbf{r}_i\}_{i=1}^{N_{\mathcal{D}}}$; $\mathbf{r}_i \in \mathbb{R}^{N_i \times d}$ represents the $N_i$ relationship features extracted from image $\mathbf{x}_i$, while $\mathbf{r}_{ij} \in \mathbb{R}^{d}$ denotes the feature of the $j$-th relationship. Following the batch-based training procedure, each iteration comprises multiple batches of equal size. We define the $b$-th batch in the $t$-th iteration as $\mathbf{B}_{tb}$, where $t \in \mathbb{Z}^{+}$ denotes the iteration index and $N_{\mathbf{B}}$ is the batch size; therefore, $\mathcal{D}$ can be partitioned into $\lceil N_{\mathcal{D}}/N_{\mathbf{B}} \rceil$ batches in each iteration. For simplicity, we assume $N_{\mathcal{D}}/N_{\mathbf{B}} \in \mathbb{Z}^{+}$. Additionally, for brevity in the following text (Section \ref{section3.2}, \ref{section3.3}, and \ref{section3.4}), we define $[K] \doteq {1, 2, \ldots, K}$ to represent the set of category indices, and separately, we use $|\cdot|$ to denote the length/size of the given set/list.

\subsection{Spurious Correlations in SGG Task} 
\label{section3.2}
As elucidated in Section \ref{section3.1}, we focus mainly on the relationship classifier $f_c$ instead of the object detector $f_o$. The latter, being pre-trained, remains frozen during training and solely extracts relationship features from $X$ to support $f_c$ training. However, this training approach compromises the independence of the relationship space for $f_c$ training. Ideally, in an independent and identically distributed (i.i.d.) sampling scenario, relationships should be sampled independently to ensure that the recognition of each is unaffected by others. In reality, however, the typical training procedures sample on a per-image basis rather than per-relationship, resulting in similar or identical relationships within a single image. For example, in an image with many people sitting on chairs, the predominant relationship is "people, sitting on, chair." As a result, the relationship features $f_o$ extracts from a single image are not independent but highly correlated.

\begin{figure}
    \footnotesize\centering
    \centerline{\includegraphics[width=1.0\linewidth]{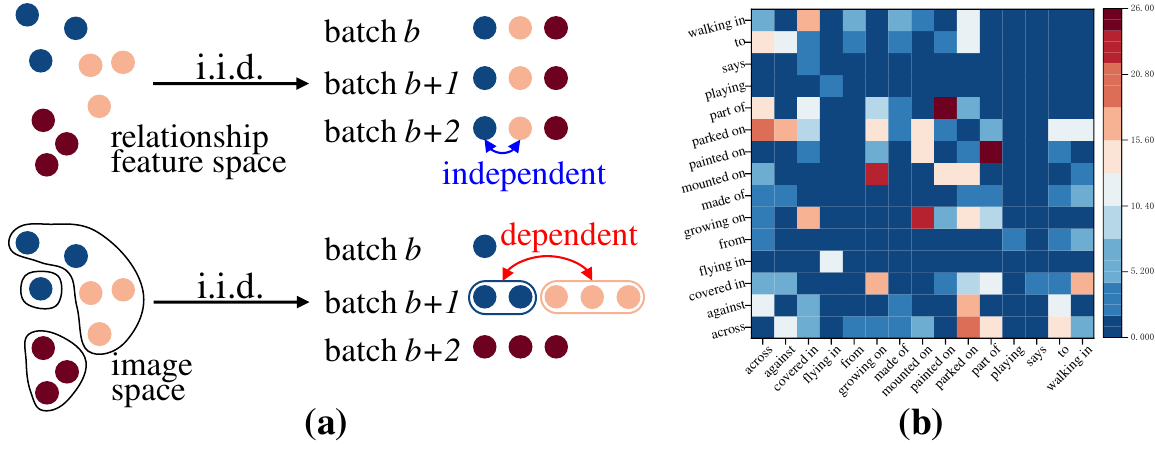}}
    \vspace{-0.3cm}
        \caption{(a) Sampling independence of the relationship feature space and the image space. Solid spheres, \protect\tikz \protect\fill[color1] (0,0) circle (0.6ex);, \protect\tikz \protect\fill[color2] (0,0) circle (0.6ex);, and \protect\tikz \protect\fill[color3] (0,0) circle (0.6ex);, denote different relationships. An irregular closed black curve symbolizes an image. (b) Co-occurrences of relationships in the VG150 train set; only a subset is displayed for clarity.}
    \label{spurious_correlations}
    \vspace{-0.3cm}
\end{figure}

In Figure \ref{spurious_correlations} (a), we depict an example with three class, showing their feature distribution at the relationship level. For simplicity, we assume $N_{\mathcal{D}}/N_{\mathbf{B}} = 3$. Hence, in an ideal scenario, after i.i.d. sampling, the distribution within each batch is expected to align with the original feature space distribution. However, because relationship features are aggregated at the image level in the SGG task, this ideal i.i.d. condition is violated. Relationships like "standing on" and "eating" frequently appear in indoor scenes, while "flying in" is common in outdoor images, as shown in Figure \ref{spurious_correlations} (a). Additionally, there is strong co-occurrence of relationships in SGG tasks, as illustrated in Figure \ref{spurious_correlations} (b). For instance, the probability of "\textit{growing on}" co-occurring with "\textit{mounted on}" is much higher than with "\textit{flying on}" in a scene. As a result, even if the images are i.i.d. sampled, the derived relationship features are dependent. Such dependency results in the identification of a relationship being impacted by other relationships in the image, constituting the spurious correlation between $X$ and $Y$. The detailed proof of spurious correlations between $X$ and $Y$ is provided in \textbf{Appendix A}.

A salient pathway through which spurious correlations clearly arise from transferring bias from $X$ to $R$, consequently misleading $f_c$ and ultimately yielding biased predictions $Y$: 
\begin{equation}
f_{\text {bias }}=\operatorname{argmin}_{f \in \mathcal{F}} \  RCE(f_c),    
\end{equation}
where $RCE$ is the Relationship Classification Error and $RCE(\cdot)$ is a supervised kernel (\eg cross-entropy loss). Recently, Wang \etal \cite{Wang} employed a balanced minibatch sampling strategy to achieve Bayes-optimal classifiers by leveraging the invariance in the underlying causal mechanisms of the data. Inspired by this work, we give two assumptions, Assumptions 1 and 2, for the spurious correlation in SGG tasks from the \textbf{batch} perspective.

\textbf{Assumption 1} (Spurious correlations in a single batch).
\textit{For any relationship feature $\mathbf{r}_{tb}^p$ in batch $\mathbf{B}_{tb}$, its outcome is influenced by another feature $\mathbf{r}_{tb}^q$ in the same batch. Formally, $Y(\mathbf{r}_{tb}^p) \neq Y(\mathbf{r}_{tb}^p \cup \mathbf{r}_{tb}^q)$, where $Y(\cdot)$ represents the prediction of the first instance in the given feature combination.}

\textbf{Assumption 2} (No spurious correlations across batches).
\textit{For any relationship feature $\mathbf{r}_{tb}^p$ in batch $\mathbf{B}_{tb}$, its outcome is unaffected by any feature $\mathbf{r}_{m n}^q$ from a different batch $\mathbf{B}_{m n}$. Formally, $Y(\mathbf{r}_{tb}^p) = Y(\mathbf{r}_{tb}^p \cup \mathbf{r}_{m n}^q)$, where $b \neq m$.}

Assumptions 1 and 2 are foundational to this paper; Assumption 1 posits that relationships within the same batch influence each other, whereas Assumption 2 asserts no such influence across different batches. The detailed mathematical proofs supporting Assumptions 1 and 2, including the analysis of parameter updates and the effects on the loss function, are provided in \textbf{Appendix C}.

\textbf{Theorem 1} (Bayesian optimal classifier induced by epoch balance). \textit {The model $f_c$ trained on $\mathcal{D}$ is a Bayesian optimal classifier $f_c^*$, \ie $\operatorname{argmax}_{y \in[K]} f_c(\mathbf{r})=\operatorname{argmax}_{y \in[K]} f_c^*(\mathbf{r})$, if the following two conditions hold: 1) $N_{\mathcal{D}} = N_{\mathbf{B}}$, 2) $|\mathbf{r}^{m}| = |\mathbf{r}^{n}|$ ($m \in [K], n \in [K], m \neq n$), where $\mathbf{r}^{m}$ is the feature of $m$-th relationship class in $\mathcal{D}$.} 

\textbf{Theorem 2} (Bayesian optimal classifier induced by batch balance). \textit {The model $f_c$ trained on $\mathcal{D}$ is a Bayesian optimal classifier, \ie $\operatorname{argmax}_{y \in[K]} f_c(\mathbf{r})=\operatorname{argmax}_{y \in[K]} f_c^*(\mathbf{r})$, if the following two conditions hold: 1) $N_{\mathcal{D}}/N_{\mathbf{B}} \in \{2, 3, 4 \cdots \}$, 2) $|\mathbf{r}^{bm}| = |\mathbf{r}^{bn}|$ ($b \in [N_{\mathcal{D}}/N_{\mathbf{B}}], m \in [K], n \in [K], m \neq n$), where $\mathbf{r}^{bm}$ represents the features of the $m$-th relationship category in the $b$-th batch.}

Assumptions 1 and 2 highlight the sources of spurious correlations in the SGG task, thereby revealing that we only need to balance within each batch to prevent spurious correlations. Let us consider the simplest full batch training case where $N_{\mathcal{D}}$ equals $N_{\mathbf{B}}$. In this case, we only need to ensure that the entire dataset is category-balanced to obtain the Bayes optimal classifier, which is unaffected by spurious correlations (see Theorem 1). The detailed proof of Theorem 1 can be found in \textbf{Appendix D}.

However, due to practical hardware constraints, most training pipelines conform to batches where $N_{\mathcal{D}}/N_{\mathbf{B}} \in \{2, 3, 4 \cdots \}$. In such cases, achieving a Bayesian optimal classifier merely necessitates balancing data within each batch (see Theorem 2) \cite{lu2019bayes,Wang}. Based on Theorem 1, we can easily derive Theorem 2. Even though $N_{\mathcal{D}} \neq N_{\mathbf{B}}$, the classification remains balanced within each batch, \ie, $|\mathbf{r}^{i m}| = |\mathbf{r}^{i n}|$. This implies that the model's updates on each batch will resemble those made over $\mathcal{D}$, leading to the equation:

\begin{equation}
\nabla_\theta L_{f_c}(\mathbf{B}_t ; \theta) \approx \nabla_\theta L_{f_c}(\mathcal{D}; \theta),
\end{equation}
where $\nabla$ and $\theta$ correspond to the gradient and model parameters, respectively, while $L_{f_c}$ denotes the loss function of the model $f_c$, typically the cross-entropy loss.

Owing to the similarity in the direction of updates, it can be further ensured that after multiple iterations, the model weights $\theta$ will gradually converge to what they would have been had the updates been performed on $\mathcal{D}$. As a result, while Theorem 2 makes use of more approximations, it still arrives at a conclusion analogous to Theorem 1, \ie, $f_c$ is a Bayesian optimal classifier.

Unfortunately, despite the potential of Theorem 2 in facilitating debiased SGG, its implementation (\eg rebalancing strategies \cite{SegG,TransRwt}) faces significant challenges. The serious long-tail distribution in SGG tasks makes achieving balanced relationship categories within each training batch nearly impossible. In this paper, we argue that spurious correlations in SGG are due to the fact that the relationship features $R$ in the standard pipeline inevitably absorb the data bias of input $X$, thereby failing to estimate a satisfactory causality of $R \rightarrow  Y$.


\subsection{Reverse Causal Framework} 
\label{section3.3}
\begin{figure}
    \footnotesize\centering
    \begin{overpic}[width=0.7\linewidth]{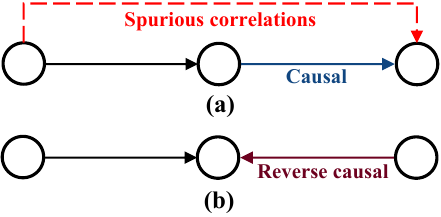}
        \put(1,10.8){ \Large $X$}
        \put(45.5,10.8){ \Large $R$}
        \put(90.5,10.8){ \Large $Y$}
        \put(1,31.8){ \Large $X$}
        \put(45.5,31.8){ \Large $R$}
        \put(90.5,31.8){ \Large $Y$}
    \end{overpic}
    \vspace{-0.3cm}
    \caption{Structural Causal Model (SCM). (a) is the causal chain structure of the typical two-stage SGG method. (b) is the reverse causal structure we propose.}
    \vspace{-0.3cm}
    \label{SCM}
\end{figure}
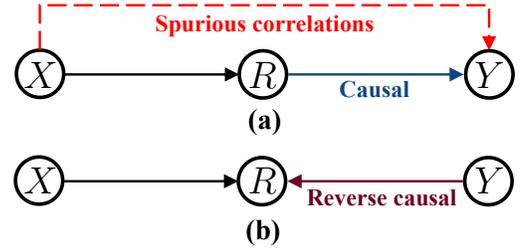

Section \ref{section3.2} underscores the spurious correlations in the SGG task and discusses the inadequacy of the batch balance method in resolving this issue. In this work, we conceptualize spurious correlations as dependencies among causal variables. We first model the Structural Causal Model (SCM) \cite{pearlCBM,CausalFairness} for the two-stage SGG training pipeline, which clearly follows a causal chain structure, $X \rightarrow R \rightarrow Y$, where $\rightarrow$ denotes the causal direction, as shown in Figure \ref{SCM} (a). Hence, the relationship classifier $f_o$ in the standard two-stage SGG framework essentially aims to estimate the causality between $R$ and $Y$, while the spurious correlations pertain to the dependency between $X$ and $Y$:
\begin{align}
P(X, Y) &= \sum_R P(X, R, Y) = \sum_R P(X) P(R \mid X) P(Y \mid R).
\label{eq:XY_dependency}
\end{align}
Equation (\ref{eq:XY_dependency}) highlights the dependency between $X$ and $Y$ in the causal chain $X \rightarrow R \rightarrow Y$, resulting in spurious correlations that negatively affect $f_o$. Inspired by causal inference theory \cite{pearl2009causality}, we model the SGG task with a reverse causal structure $X \rightarrow R \leftarrow Y$; this approach inherently circumvents the spurious correlation between $X$ and $Y$, as shown in Figure \ref{SCM} (b). According to the D-Separation principle, if $R$ is not given, $X$ and $Y$ are independent since $R$ acts as a collider, resulting in $P(X,Y) = P(X)P(Y)$. Additionally, even when $R$ is given, the paths from $X \rightarrow R$ and $R \leftarrow Y$ are decoupled, preventing collider bias and maintaining independence between $X$ and $Y$. Hence, we still have:
\begin{align}
P(X, Y) &= \sum_R P(X, R, Y) \nonumber \\
&= P(X) P(Y) \sum_R P(R \mid X, Y) = P(X) P(Y) .
\label{eq:P(X)P(Y)}
\end{align}
As depicted in Equation (\ref{eq:P(X)P(Y)}), $X$ and $Y$ are independent. This independence does not mean that image features $X$ are unrelated to outcomes $Y$, but that relationship predictions in a scene do not depend on other concurrent relationships.  For instance, consider a scene with $N+1$ relationships; even if $N$ of these relationships are ``\textit{sitting on},'' it does not necessarily follow that the additional one is also ``\textit{sitting on}.'' Unfortunately, in the existing paradigms based on chain causality, $X$ and $Y$ are highly dependent, and in the example given above, another relationship might be predicated as ``\textit{sitting on},'' meaning the prediction of a instance could be influenced by other relationships, which is the spurious correlations we focus on in this paper. Inspired by Equation (\ref{eq:P(X)P(Y)}), to eliminate spurious correlations between $X$ and $Y$, all we need is an appropriate intervention on the confounder $R$ within the reconstructed reverse causal structure \cite{pearl2009causality}. To achieve this goal, our method intervenes on the relationship features provided by $f_o$; these obtained features can, therefore, be regarded as independent and identically distributed samples from the entire relationship feature space. Consequently, RcSGG estimates a clean reverse causality, \ie $R \leftarrow Y$.

To estimate the reverse causality $R \leftarrow Y$, we propose Active Reverse Estimation (ARE), inspired by the Active Learning (AL) paradigm \cite{AL_1,AL_2,activelearning}. Our main insight is that $R \leftarrow Y$ signifies that the relationship features $R$ are derived from the predictions $Y$, which is highly similar to the fundamental concept of AL, \ie \textbf{model-to-data}, where the sampling process in the current AL round is often guided by the model results obtained from the previous rounds. Specifically, AL acquires a limited number of data points from the query set based on the query conditions for model training, thereby alleviating expensive labeling costs. Generally, the query conditions of the current AL round are determined by the model results of the previous rounds. Readers may refer to AL literature \cite{AL_1,AL_2} and its exploration \cite{EDAL} in the SGG task for more information.

\textbf{Theorem 3} (Bayesian optimal classifier induced by approximate batch balance). \textit {The model $f_c$ trained on $\mathcal{D}$ is a Bayesian optimal classifier, \ie $\operatorname{argmax}_{y \in[K]} f_c(\mathbf{r})=\operatorname{argmax}_{y \in[K]} f_c^*(\mathbf{r})$, if the following two conditions hold: 1) $N_{\mathcal{D}}/N_{\mathbf{B}} \in \{2, 3, 4 \cdots \}$, 2) $|\mathbf{r}^{bm}| \approx |\mathbf{r}^{bn}|$ ($b \in [N_{\mathcal{D}}/N_{\mathbf{B}}], m \in [K], n \in [K], m \neq n$).}

In summary, ARE leverages the model-to-data paradigm of AL to simulate reverse causality $R \leftarrow Y$. Although AL sampling results are often not class-balanced because query conditions do not apply equally to all classes, they can achieve rough balance through adaptive adjustments in subsequent rounds \cite{EDAL}. This enables ARE to optimize the relationship feature $R$ effectively, leading to an approximate batch balance. While this may not fully satisfy the strict balance conditions of Theorem 2, it still supports achieving a Bayesian optimal classifier as outlined in Theorem 3, with a detailed proof in \textbf{Appendix E}.

\subsubsection{Active Reverse Estimation}
\label{section_ARE}
Section \ref{section3.2} highlights that the unobserved confounder $Z$ introduces spurious correlations between $X$ and $Y$, and thus, the relationship classifier $f_c$ is actually learned within a perturbed effective parameter space $\mathcal{F}_{\mathbf{R}}$ from the parameter space $\mathcal{F}$. It should be noted that in this paper, effective parameter space and parameter space have distinct meanings. The former refers to the range of parameters learnable given the training data under a specific model, while the latter denotes the set of all possible parameter values within the model. Consequently, $\mathcal{F}_{\mathbf{R}}$ is determined by both $f_c$ and $R$, whereas $\mathcal{F}$ is solely $f_c$ dependent. As a result,  even under ideal training conditions, the $f_c$ cannot evade the impact of these correlations. This prompts us to seek an effective parameter space $\mathcal{F}_{\widetilde{\mathbf{R}}}$ devoid of such spurious correlations, wherein the Bayesian optimal classifier $f_c^*$ can be achieved using a standard classification supervised kernel: 
\begin{equation}
f_c^*=\operatorname{argmin}_{f_c \in \mathcal{F}_{\widetilde{\mathbf{R}}}} \ RCE(f_c).
\label{f_c^*}
\end{equation}
Equation (\ref{f_c^*}) shows that achieving the Bayesian optimal classifier $f_c^*$ necessitates the identification of the effective parameter space $\mathcal{F}_{\widetilde{\mathbf{R}}}$, governed by $f_c$ and $R$. In the two-stage SGG framework, this space is composed of Spurious Correlation Error (SCE) and Overlapping Error (OE), with OE representing the intersection of SCE and RCE; further analysis of $\mathcal{F}_{\widetilde{\mathbf{R}}}$ can be found in \textbf{Appendix B}. To ensure $f_c$ learns within $\mathcal{F}_{\widetilde{\mathbf{R}}}$, we propose Active Reverse Estimation (ARE), which optimizes the relationship feature $R$ by estimating the reverse causality $R \leftarrow Y$. ARE is inspired by AL, which designs query conditions to select high-information samples from the query set to reduce annotation costs \cite{EDAL,activelearning}. 

Similarly, we first establish the query set $\mathcal{Q}$ and design the query conditions $\mathcal{Q}_{cond}$, both of which are conducted within the relationship feature space, inherently avoiding spurious correlations between $X$ and $Y$. Specifically, let $\mathcal{Q}=\{\mathbf{Q}_i\}_{i=1}^{K^{\prime}}$ ($K^{\prime} <K$), where $\mathbf{Q}_{i}$ denotes all the features corresponding to the $i$-th relationship. $\mathcal{Q}$ does not encompass all relationship categories for two primary reasons: 1) The extreme long-tail distribution of the SGG dataset ensures that nearly all images in each batch include head relationship features. 2) While our designed query condition $\mathcal{Q}_{\text{cond}}$ offers some resilience against disturbances in the distribution of the query set $\mathcal{Q}$, it remains crucial to prevent severe long-tail distribution issues within the relationship classes of $\mathcal{Q}$, as attested to in Section 4.3. Intuitively, this can be achieved by down-sampling $\mathcal{Q}$, but this sacrifices some features. Instead, we achieve this aim by adjusting $K^{\prime}$, ensuring that relationship features are fully leveraged and offering a scalable avenue for applying our method to datasets with varying degrees of long-tail distributions. 

Then, we design the query conditions $\mathcal{Q}_{cond}$ to sample from the query set $\mathcal{Q}$ to estimate the reverse causality $R \leftarrow  Y$. Inspired by the model-to-data training mode in AL, we propose a loss-driven estimation approach that spans across batches. Considering the consistency of query conditions across batches, we focus on the query condition $\mathcal{Q}_{cond}^{tb}$  of a specific batch, \ie the $b$-th batch in the $t$-th iteration, as well as its entire process of reverse causality estimation. let $\mathcal{L}_{tb}=\{L_{tb}^k\}_{k=1}^{K^{\prime}}$ and $L_{tb}^k$ denote the average loss of the $k$-th category in query set $\mathcal{Q}$ when training $f_c$ with $\mathbf{B}_{tb}$. We then model the query condition $\mathcal{Q}_{cond}^{tb}$ using the probability distribution $\mathcal{P}_{tb}=\{P_{tb}^k\}_{k=1}^{K^{\prime}}$ of $\mathcal{L}_{tb}$, where 
\begin{equation}
P_{tb}^k = \exp (-\alpha L_{tb}^k) / \sum_{k=1}^{K^{\prime}} \exp (-\alpha L_{tb}^k), 
\label{probability}
\end{equation}
and $\alpha$ here is a weight factor used to adjust the attention to categories within different loss intervals. $\mathcal{Q}_{cond}^{tb}$ is built upon $\mathcal{P}_{tb}$; therefore, $\mathcal{Q}_{cond}^{tb}$ captures the reverse causality $R \leftarrow Y$, as the relationship feature space $R$ for the current batch is determined by the loss of the previous batch, with the loss naturally interpreted as the category-level training outcome $Y$. Based on this observation, we model the reverse causality estimation as sampling $K^{\prime}$ categories from the query set $\mathcal{Q}$ based on the loss probability distribution $\mathcal{P}_{tb}$ and incorporate the sampled results into the current batch's relationship feature space. Specifically, for $\mathcal{Q}_{cond}^{tb}$, it implies that the sampling size for the corresponding category in $\mathcal{Q}$ is $\mathbf{N}_{tb}=\{\widetilde{n}_{tb}^k\}_{k=1}^{K^{\prime}}$, where
\[
    \widetilde{n}_{tb}^k = \lambda P_{tb}^k |\mathbf{Q}_k|,
\]
and $\lambda$ is a hyper-parameter utilized to control the order of magnitude of the sampling process. To ensure coverage of the entire relationship feature space, we employ a sampling without replacement strategy from the query set $\mathcal{Q}$. Once a particular relationship feature is completely sampled, all features of that category in $\mathcal{D}$ are replenished back into $\mathcal{Q}$.  

The reverse causality estimation consists of two steps: \textbf{Step 1}) Calculate the query condition $\mathcal{Q}_{cond}^{tb}$ based on $\mathcal{P}_{tb}$. \textbf{Step 2}) Sample and integrate relationship features $\mathbf{R}_{tb}^{+}=\{\mathbf{r}_{tb}^{k+}\}_{k=1}^{K^{\prime}}$ into the current batch’s feature space based on $\mathcal{Q}_{cond}^{tb}$. These features are derived using the sampling kernel $\xi_{\Delta}(\cdot, \cdot)$, with $\Delta$ denoting the sampling method, focusing on the quantity and source pool for each category. For instance, the sampled relationship feature of $k$-th category in $\mathcal{Q}$ can be calculated as:
\[
    \mathbf{r}_{tb}^{k+} = \xi_{\Delta}(\mathbf{Q}_k, \widetilde{n}_{tb}^k).
\]
Intuitively, $\Delta$ in the above formula can be random sampling (rnd). However, this naive approach often yields low-information samples. Consequently, we propose a Maximum Information Sampling (MIS) approach (detailed in Section \ref{MIS}). Thus, within our framework, $\mathbf{r}_{tb}^{k+}$ is computed as:
\begin{equation}
    \mathbf{r}_{tb}^{k+} = \xi_{MIS}(\mathbf{Q}_k, \widetilde{n}_{tb}^k).
    \label{add_fg}
\end{equation}
It can be observed that $\mathbf{r}_{tb}^{k+}$ is sampled from $\mathcal{Q}$ based on the query condition $\mathcal{Q}_{cond}^{tb}$, and each feature cannot be traced back to its original image $X$, as the reverse causality estimation intervenes in the feature space $R$. As a result, we have $P(X, Y) = P(X) P(Y)$.

\begin{figure}
    \footnotesize\centering
    \begin{overpic}[width=0.9\linewidth]{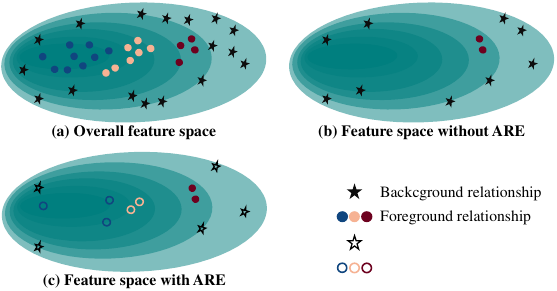}
        \put(67.5,8){ $\widetilde{\mathbf{r}}_{tb}^{bg}$}
        \put(67.5,3){ $\mathbf{R}_{tb}^{+}$}
    \end{overpic}
    \vspace{-0.3cm}
    \caption{(a) Distribution of relationship features across the entire dataset. (b) and (c) depict the distribution of relationship features within a batch; the former represents the original distribution, while the latter illustrates the distribution optimized by our proposed Active Reverse Estimation (ARE).}
    \vspace{-0.2cm}
    \label{ARE}
\end{figure}

Beyond the reverse causality estimation for foreground relationships, we also observed that the SGG dataset is replete with extensive background relationships, as depicted in Figure \ref{motivation} (b). This may cause $f_c$ to misclassify foreground relationships as background ones, \ie, fore-back bias, as discussed in Section \ref{sec:introduction}.

Continuing with the example of the $t$-th iteration's $b$-th batch, $\mathbf{r}_{tb}^{fg}$ and $\mathbf{r}_{tb}^{bg}$ represent the foreground and background relationships within the batch, respectively. For $\mathbf{r}_{tb}^{bg}$, we retain only $min(|\mathbf{r}_{tb}^{bg}|,\widetilde{n}_{t b}^{bg})$ background relationships, where $min(\cdot, \cdot)$ is a computation kernel to take the minimum value of the input at two positions. It is given that:
\[
    \widetilde{n}_{t b}^{bg} = \pi(\{\widetilde{n}_{tb}^k\}_{k=1}^{K^{\prime}} + |\mathbf{r}_{tb}^{fg}|),
\]
and $\pi$ is a hyper-parameter. Herein, $\{\widetilde{n}_{tb}^k\}_{k=1}^{K^{\prime}} + |\mathbf{r}_{tb}^{fg}|$ computes the number of foreground relationships within the batch after reverse causality estimation. Consequently, the aforementioned equation aims to maintain a fixed ratio between background and foreground relationships, specifically at $\pi$:1. Thus, the retained background relationships $\widetilde{\mathbf{r}}_{tb}^{bg}$ can be sampled as: 
\begin{equation}
    \widetilde{\mathbf{r}}_{tb}^{bg} = \xi_{rnd}(\mathbf{r}_{tb}^{bg}, \widetilde{n}_{tb}^{bg}).
\label{bg_rnd}
\end{equation}
Here, we have an interesting observation. MIS demonstrates significant enhancement for foreground relationships, while its effect on background relationships closely resembles that of random sampling, as elaborated in Section \ref{section4.3}. Thus, we employ random sampling $rnd$ in Equation (\ref{bg_rnd}) to maintain high efficiency. We argue that the intervention on background relationships focuses on maintaining its ratio with foreground relationships rather than on the relationships themselves. This assertion can be validated by adjusting $\pi$, also explored in Section \ref{section4.3}. Within the $t$-th iteration, the relationship features $\mathbf{B}_{tb}$ of the $b$-th batch is eventually updated to $\widetilde{\mathbf{B}}_{tb}$, 
\begin{equation}
\widetilde{\mathbf{B}}_{tb} = \widetilde{\mathbf{r}}_{tb}^{bg} \cup \mathbf{r}_{tb}^{fg} \cup \mathbf{R}_{tb}^{+} , 
\end{equation}
serving as the basis for training the desired model $f_c$. As illustrated in Figure \ref{ARE} (b), in a typical training procedure, a batch often covers only a limited number of foreground relationships while being inundated with numerous background relationships. However, our proposed reverse causal estimation, by emphasizing the impact of $Y$ on $R$, significantly optimizes this skewed relationship feature space, as seen in Figure \ref{ARE} (c). It is worth mentioning that $\mathbf{R}_{tb}^{+}$ can achieve approximate balance by adjusting $\lambda$ without sacrificing the properties of reverse causal estimation. The trained model will, therefore, be a Bayesian optimal classifier; see Theorem 3. Moreover, ARE depends on Assumption 2's premise of batch independence. Even in practical scenarios where this assumption is not strictly satisfied, we argue that ARE remains effective due to: 1) The local optimization of feature spaces for each batch, driven by the dynamic condition $\mathcal{Q}_{cond}$, minimizes disruptions between consecutive batches, and 2) The non-replacement strategy prevents the propagation of correlations across batches, enabling ARE to maintain its robustness even when the ideal conditions are not fully met.

In summary, we propose ARE to intervene in the relationship feature space during classifier training. While this method may appear similar to the resampling methods \cite{SegG,TransRwt} reviewed in Section \ref{sec:Related_SGG}, ARE distinguishes itself in three key ways:

\textbf{1) Different sampling drivers.} Resampling methods address imbalances by reweighting class frequencies through oversampling or undersampling, whereas ARE, inspired by AL, is loss-driven and enhances the feature space during training. To our knowledge, ARE is the first method to leverage AL techniques for intervening in causal relationships in SGG tasks.

\textbf{2) Focus on both foreground and background relationships.} Existing resampling methods primarily focus on foreground relationships. ARE, however, incorporates both foreground and background relationships through reverse causal estimation, enabling it to address both head-tail bias and fore-back bias.

\textbf{3) Granularity of sampling.} Resampling methods typically focus solely on relationship distributions, overlooking the impact of object distributions. ARE, guided by the proposed MIS, simultaneously considers both relationship and object attributes, ensuring that the feature space captures the maximum possible information. Details of MIS are provided in Section \ref{MIS}.


\begin{figure}
    \footnotesize\centering
    \centerline{\includegraphics[width=0.9\linewidth]{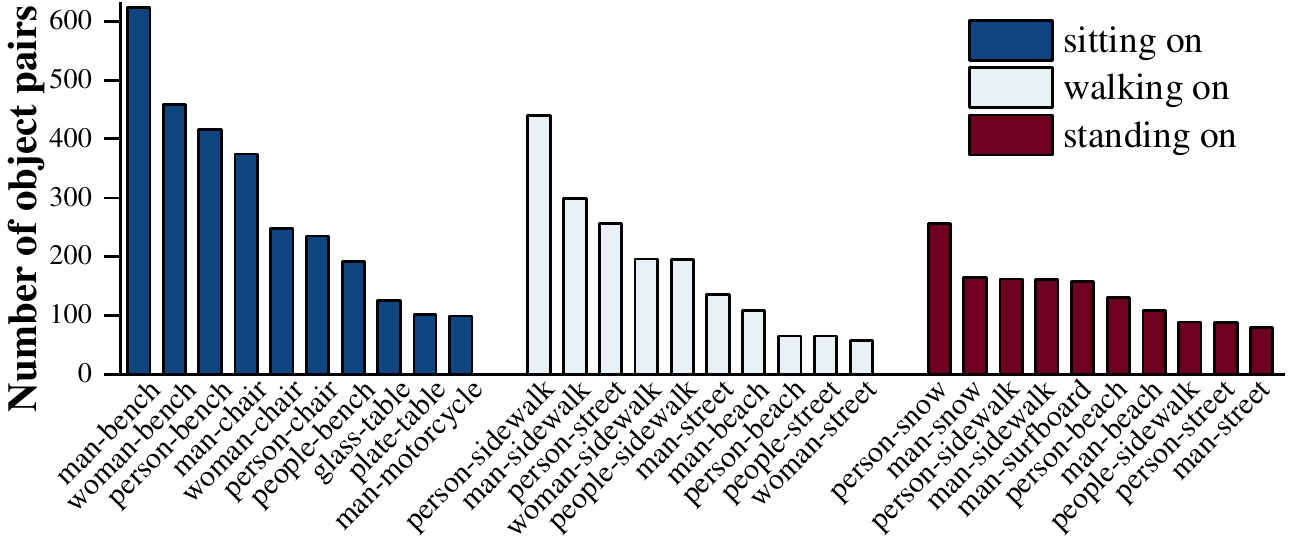}}
    \vspace{-0.3cm}
        \caption{Given the distribution of object pairs under specific relationships, for clarity, only the top 10 object pairs are displayed for each relationship.}
    \label{object_pair}
    \vspace{-0.2cm}
\end{figure}

\subsubsection{Maximum Information Sampling}
\label{MIS}
This subsection introduces Maximum Information Sampling (MIS). The motivation for MIS arises from several unique challenges of integrating AL concepts into SGG tasks:

\textbf{1) Dual imbalances in SGG.} SGG tasks are characterized by both relationship imbalance and object imbalance, which further propagate into object pair imbalance, as illustrated in Figure \ref{object_pair}. For instance, object pairs such as ``\textit {$<$man, \textbf{sitting on}, bench$>$}" dominate the dataset, while rarer combinations are severely underrepresented. This imbalance not only skews the sampled feature space but also exacerbates spurious correlations, limiting the representativeness of training batches.

\textbf{2) Category-to-instance sampling gap.} The query conditions in AL are typically at the instance-level \cite{AL_1,AL_2,EDAL}; however, the relationship query conditions we proposed are at the category-level. This implies that we need to conduct instance-level sampling from $\mathcal{Q}$ using category-level query conditions, which, could result in ambiguous reverse causal estimations.

\textbf{3) Integration with reverse causality.} Effective sampling in SGG must also align with the reverse causal framework introduced. Sampling strategies need to simultaneously maximize mutual information and adaptively refine relationship features ($R$) based on outcomes ($Y$) to ensure the elimination of spurious correlations and maintain causal consistency.

In response to these challenges, we introduce MIS, which assesses relationship feature information using object pairs to maximize diversity within each sampled batch. For the sake of generality, we illustrate MIS using the $k$-th relationship category as an example, where $\widetilde{n}_{tb}^k$ samples are drawn from $\mathbf{Q}_k$.  For this category, assuming that the frequency of the object pair $<o_\mu, o_\nu>$ in the current sampling pool is $p(o_\mu, o_\nu)$, where $o_\mu$ and $o_\nu$ represent the object and subject respectively, we define its information content within $\mathbf{Q}_k$ as $- \log (p(o_\mu, o_\nu))$. Hence, the entropy of $\mathbf{Q}_k$ can be formulated as:
\begin{equation}
H(\mathbf{Q}_k)=-\sum_{(o_\mu, o_\nu) \in \mathbf{Q}_k} p(o_\mu, o_\nu) \log (p(o_\mu, o_\nu)).
\end{equation}
Following sampling without replacement, similarly, the entropy of $\mathbf{Q}_k$ is given by:
\begin{equation}
\begin{aligned}
H(\mathbf{Q}_k \mid \mathbf{r}_{tb}^{k+}) &=\sum_{(o_\mu, o_\nu) \in \mathbf{Q}_k} p(o_\mu, o_\nu) H(\mathbf{Q}_k \mid o_\mu, o_\nu) \\
& =-\sum_{(o_\mu, o_\nu) \in \mathbf{r}_{tb}^{k+}} p(o_\mu, o_\nu) \times \\ 
&\sum_{(o_\zeta, o_\omega) \in \mathbf{Q}_k} p(o_\zeta, o_\omega \mid o_\mu, o_\nu) \log (p(o_\zeta, o_\omega \mid o_\mu, o_\nu)).
\end{aligned}
\end{equation}
We then define the mutual information $I(\mathbf{r}_{tb}^{k+}; \mathbf{Q}_k)$ of $\mathbf{r}_{tb}^{k+}$ and $\mathbf{Q}_k$, \ie, the degree of information sharing between the two sets. $I(\mathbf{r}_{tb}^{k+} ; \mathbf{Q}_k)$ can therefore be calculated as: 
\begin{equation}
I(\mathbf{r}_{tb}^{k+} ; \mathbf{Q}_k)=H(\mathbf{Q}_k)-H(\mathbf{Q}_k \mid \mathbf{r}_{tb}^{k+}). 
\end{equation}
This implies that to achieve maximum information sampling, we merely need to select an object pair $<o_\mu, o_\nu>, \mu \in[K^{\prime}], \nu \in[K^{\prime}]$ from $\mathbf{Q}_k$ without replacement, where $<o_\mu, o_\nu>$ maximizes $I(\mathbf{Q}_k \cup\{<o_\mu, o_\nu>\}; \mathbf{r}_{tb}^{k+})$. Hence, for each sampling iteration, we compute the estimated mutual information gain, $\Delta I\left(o_\mu, o_\nu\right)$, for every object pair that has not been selected yet. If all object pairs have been chosen, we reset them as unselected. $\Delta I\left(o_\mu, o_\nu\right)$ can be calculated as:
\begin{align}
\Delta I(o_\mu, o_\nu) & =  p(o_\mu, o_\nu \mid \mathbf{Q}_k) \log (\frac{p(o_\mu, o_\nu \mid \mathbf{Q}_k)}{p(o_\mu \mid \mathbf{Q}_k) \times p(o_\nu \mid \mathbf{Q}_k)}) \nonumber \\
&+ p(o_\mu, o_\nu \mid \mathbf{r}_{tb}^{k+}) \log (\frac{p(o_\mu, o_\nu \mid \mathbf{r}_{tb}^{k+})}{p(o_\mu \mid \mathbf{r}_{tb}^{k+}) \times p(o_\nu \mid \mathbf{r}_{tb}^{k+})}).
\end{align}
We select the sample $<o_{\mu_{\max }}, o_{\nu_{\max }}>$ with the maximum information content in the current sampling pool, where
\begin{equation}
<o_{\mu_{\max }}, o_{\nu_{\max }}>=\operatorname{argmax}_{o_\mu, o_\nu} \Delta I\left(o_\mu, o_\nu\right).
\label{mutual_information_sample}
\end{equation}

Ultimately, we add $<o_{\mu_{\max}}, o_{\nu_{\max}}>$ to $\mathbf{r}_{tb}^{k+}$ and remove it from $\mathbf{Q}k$, repeating this process until the sample count reaches $\widetilde{n}_{tb}^k$.  Equation (\ref{mutual_information_sample}) seeks to maximize mutual information, essential but computationally intense due to the large number of relationships in SGG tasks. Despite setting $K'$ to limit $\mathcal{Q}$'s size, calculating mutual information directly remains costly. However, maximizing diversified relationship features within each batch is crucial for SGG tasks and aligns with batch balance as indicated in Theorem 3. This observation provides us with an opportunity for optimization. Specifically, we forego detailed frequency considerations of object pairs $p(o_\mu, o_\nu)$, we focus on maximizing information by selecting unique or non-redundant instances from $\mathcal{Q}$. While "maximum information" differs subtly from "maximum mutual information," their effects are similar in SGG contexts. Our optimized method is delineated in Algorithm 1, where we seek maximum information sampling by selecting as diverse object pairs as possible (lines 6-7 in Algorithm 1).   

\begin{algorithm}[t]
\caption{Maximum Information Sampling}\label{algorithm_MIS}
\begin{flushleft}
\textbf{Require:} $\mathcal{Q}=\{\mathbf{Q}_i\}_{i=1}^{K^{\prime}}$, $\mathbf{N}_{tb}=\{\widetilde{n}_{tb}^k\}_{k=1}^{K^{\prime}}$ \\
\textbf{Ensure:} $\mathbf{R}_{tb}^{+}=\{\mathbf{r}_{tb}^{k+}\}_{k=1}^{K^{\prime}}$ \\
1: \textbf{for} {$k$} = 1 \textbf{to} $K^{\prime}$ \\
2: \hspace*{1.5em} $\mathbf{r}_{tb}^{k+} = 0$, $remaining\_pairs$ = $\mathbf{Q}_i$ \\
3: \hspace*{1.5em} \textbf{while} $|\mathbf{r}_{tb}^{k+}| < \widetilde{n}_{tb}^k$ \textbf{do} \\ 
4: \hspace*{3em} get $unique\_pairs$ \\
5: \hspace*{4.5em} \textbf{if} $|\mathbf{r}_{tb}^{k+}| + |unique\_pairs| \leq \widetilde{n}_{tb}^k$ \\
6: \hspace*{6em} \textbf{then} add $unique\_pairs$ to $\mathbf{r}_{tb}^{k+}$ \\
7: \hspace*{4.5em} \textbf{else} $need\_num = \widetilde{n}_{tb}^k - |\mathbf{r}_{tb}^{k+}|$, randomly \hspace*{7.7em} select $need\_num$ instances from \hspace*{7.7em} $unique\_pairs$ and add them to $\mathbf{r}_{tb}^{k+}$ \\
8: \hspace*{3em} update $remaining\_pairs$
\end{flushleft}
\end{algorithm}

\subsection{Discussion} 
\label{section3.4}
\subsubsection{Compared to the Reweighting and Resampling Methods} 
\label{section3.4.1}
As noted in related work, common strategies like resampling and reweighting adjust the input or output to balance each relationship's impact on model training. While there are certain similarities, our approach fundamentally differs from these strategies. We emphasize the following two distinctions:

\textit {Elimination of Spurious Correlations.} Resampling and reweighting adjust the contribution of different classes within a batch. Clearly, these strategies cannot address the spurious correlation issues in the two-stage SGG framework because 1) all extracted relationship features can be directly traced back to their respective images and 2) the relationships derived are solely from a limited set of images (correlated with the batch size), inundated with many scene-specific, similar features. However, our method, by establishing a relationship-level query set, can decouple the relationship from its originating image, eliminating the scene restrictions and batch size limitations within the batch, thereby mitigating the spurious correlation.

\textit {Direction of causality estimation.} Both reweighting and resampling techniques intervene with features typically based on relationship distributions or, more advanced, by enabling the model to learn the intervention procedure adaptively. Regardless, these methods adhere to the intrinsic feature extraction and processing pipeline within the original batch, thereby essentially estimating the forward causality from $R \rightarrow Y$. In contrast, our approach emphasizes leveraging the training results from a preceding batch as query conditions for the relationship feature space of the current batch. As a result, our interventions on the feature space operate based on a reverse causality estimation, specifically, $R \leftarrow Y$.

\subsubsection{Compared the Re-annotate Methods}

Some existing works attribute fore-back bias to the mislabeling of certain foreground relationships as background, and thus, these studies often re-annotate background relationships \cite{fore-back,NICE}. For instance, NICE \cite{NICE} identifies contentious background relationships and reallocates them with soft foreground labels. Clearly, our method is \textbf{orthogonal} to such approaches. Intuitively, applying our method on top of re-annotation approaches could further alleviate the fore-back bias issue, because fore-back bias would still exist even in an ideal state without any missing foreground relationship labels. Combining re-annotation methods with RcSGG is part of our future work.

\subsubsection{Compared to Debiased SGG for Dual Imbalance (DSDI)}

DSDI \cite{DSDI} is the work most similar to our approach, also addressing both fore-back bias and head-tail bias as well. For fore-back bias, DSDI designed BR Loss to decouple background classification from foreground relationship recognition. For head-tail bias, BR Loss uses a weighting mechanism to prioritize tail relationships. Despite focusing on similar challenges, our method significantly differs from DSDI: 1) Unlike DSDI, which processes background and foreground relationships separately, our approach treats background relationships as a regular class, alleviating fore-back bias by maintaining a fixed background to foreground ratio. 2) Inspired by AL, we introduce ARE, which updates the distribution of foreground relationships in each batch to address fore-back bias. Importantly, the updated distribution extends beyond individual batch contexts, helping to disrupt spurious correlations between images and predictions, unlike DSDI which confines its approach to weighting tail relationships within batch boundaries.

\begin{table*}[htbp]
  \centering
  \vspace{-0.3cm}
  \caption{Quantitative results (mR@K) of RcSGG and baseline methods on the MotifsNet, VCTree, and Transformer backbones. $\text{AVG}_{\text{mR}}^{\Delta}$ denotes the mean of mR@20, mR@50, and mR@100, while $\text{AVG}_{\text{mR}}^{\Diamond}$ represents the mean of mR@50 and mR@100. We \underline{underline} the optimal results for emphasis. Please note that the backbone method is provided for reference only and is not included in the ranking.}
  \vspace{-0.3cm}
    \resizebox{19.5cm}{!}{\begin{tabular}{c|l|ccccc|ccccc|ccccc}
    \toprule
          &   & \multicolumn{5}{c|}{PredCls}  & \multicolumn{5}{c|}{SGCls}    & \multicolumn{5}{c}{SGDet} \\
          &   & \multicolumn{1}{c}{mR@20} & \multicolumn{1}{c}{mR@50} & \multicolumn{1}{c}{mR@100} & \multicolumn{1}{c}{$\text{AVG}_{\text{mR}}^{\Delta}$} & \multicolumn{1}{c|}{$\text{AVG}_{\text{mR}}^{\Diamond}$} & \multicolumn{1}{c}{mR@20} & \multicolumn{1}{c}{mR@50} & \multicolumn{1}{c}{mR@100} & \multicolumn{1}{c}{$\text{AVG}_{\text{mR}}^{\Delta}$} & \multicolumn{1}{c|}{$\text{AVG}_{\text{mR}}^{\Diamond}$}  &\multicolumn{1}{c}{mR@20} & \multicolumn{1}{c}{mR@50} & \multicolumn{1}{c}{mR@100} & \multicolumn{1}{c}{$\text{AVG}_{\text{mR}}^{\Delta}$} & \multicolumn{1}{c}{$\text{AVG}_{\text{mR}}^{\Diamond}$} \\
    \midrule
    \midrule
    \multirow{46}{*}{\rotatebox{90}{VG150 \cite{VG150}}} 
    & SSRCNN-SGG \cite{SSRCNN-SGG} \scriptsize \textit {(CVPR’22)}  &  $ - $    &   $ - $    &   $ - $    &  $ - $     & $ - $      &   $ - $    &   $ - $    &   $ - $    &   $ - $    &   $ - $    &  $ 6.1 $     & $ 8.4 $  &   $ 10.0 $    &   $ 8.2 $    & $ 9.2 $ \\
    & HetSGG \cite{HetSGG} \scriptsize \textit {(AAAI’23)}   &  $ - $    &   $ 32.3 $    &   $ 34.5 $    &  $  -$     & $ 33.4 $      &   $ - $    &   $ 15.8 $    &   $ 17.7 $    &   $ - $    &   $ 16.8 $    &  $ - $     & $ 11.5 $  &   $ 13.5 $    &   $ - $    & $  12.5 $ \\
    & SQUAT \cite{SQUAT} \scriptsize \textit {(CVPR’23)}   &  $ - $    &   $ 30.9 $    &   $ 33.4 $    &  $ - $     & $ 32.2  $      &   $ - $    &   $ 17.5 $    &   $  18.8 $    &   $ - $    &   $ 18.2 $    &  $ - $     & $ 14.1 $  &   $ 16.5 $    &   $ - $    & $ 15.3 $ \\
    & CV-SGG \cite{CV-SGG} \scriptsize \textit {(CVPR’23)}   &  $ - $    &   $ 32.6 $    &   $ 36.2 $    &  $ - $     & $ 34.4 $      &   $  -$    &   $ - $    &   $ - $    &   $ - $    &   $ - $    &  $ - $     & $ 14.6 $  &   $ 17.0 $    &   $ - $    & $15.8 $ \\
    & PE-Net \cite{PENET} \scriptsize \textit {(CVPR’23)}   &  $ - $    &   $ 31.5$    &   $ 33.8$    &  $ - $     & $32.7$      &   $  -$    &   $ 17.8 $    &   $18.9$    &   $ - $    &   $ 18.4 $    &  $ - $     & $ 12.4 $  &   $14.5 $    &   $ - $    & $13.5$ \\
    & DSDI \cite{DSDI} \scriptsize \textit {(TPAMI’23)}   &  $ 28.5 $    &   $ 33.6$    &   $ 35.5 $    &  $ 32.5 $     & $ 34.6 $      &   $16.6$    &   $20.4$    &   $21.4$    &   $19.5$    &   $20.9$    &  $ 7.7 $     & $10.3$  &   $12.1$    &   $10.0$    & $11.2$ \\
    \cline{2-17} \noalign{\vskip 1mm}
     & MotifsNet (backbone) \cite{Neuralmotifs}  & $12.2$  & $15.5$  & $16.8$  & $14.8$  & $16.2$  & $7.2$   & $9.0$     & $9.5$   & $8.6$ & $9.3$  & $5.2$   & $7.2$   & $8.5$   & $7.0$ & $7.9$ \\
    & \quad $\text {HML \cite{HML}}$  \scriptsize \textit {(ECCV’22)} & $30.1$  & $36.3$  & $38.7$  & $35.0$ & $37.5$ & $17.1$  & $20.8$  & $22.1$  & $20.0$ & $21.5$  & $10.8$  & $14.6$  & $17.3$  & $14.2$ & $16.0$  \\
    & \quad $\text {FGPL \cite{FGPL}}$  \scriptsize \textit {(CVPR’22)}  & $24.3$  & $33.0$  & $37.5$  & $31.6$ & $35.3$ & $17.1$  & $21.3$  & $22.5$  & $20.3$ & $21.9$ & $11.1$  & $15.4$  & $18.2$  & $14.9$  & $16.8$ \\
    & \quad $\text {TransRwt \cite{TransRwt}}$  \scriptsize \textit {(ECCV’22)}  & $-$     & $35.8$  & $39.1$  & $-$ & $37.5$ & $-$     & $21.5$  & $22.8$  & $-$ & $22.2$ & $-$     & $15.8$  & $18.0$  & $-$ & $16.9$  \\
    & \quad $\text {PPDL \cite{PPDL}}$  \scriptsize \textit {(CVPR’22)}  & $-$     & $32.2$  & $33.3$  & $-$  & $32.8$  & $-$     & $17.5$  & $18.2$  & $-$  & $17.9$  & $-$     & $11.4$  & $13.5$  & $-$ & $12.5$  \\
    & \quad $\text {GCL \cite{GCL}}$  \scriptsize \textit {(CVPR’22)}  & $30.5$  & $36.1$  & $38.2$  & $34.9$ & $37.2$ & $18.0$    & $20.8$  & $21.8$  & $20.2$  & $21.3$  & $12.9$  & $16.8$  & $19.3$  & $16.3$ & $18.1$ \\
    
     & \quad  $\text {NICE \cite{NICE}}$  \scriptsize \textit {(CVPR’22)}  & $-$     & $29.9$    & $32.3$  & $-$ & $31.1$  & $-$     & $16.6$  & $17.9$  & $-$  & $17.3$ & $-$     & $12.2$  & $14.4$  & $-$ & $13.3$  \\
    & \quad $\text {RTPB \cite{RTPB}}$  \scriptsize \textit {(AAAI’22)}  & $28.8$  & $35.3$  & $37.7$  & $33.9$ & $36.5$  & $16.3$  & $19.4$  & $22.6$  & $19.4$  & $21.0$ & $9.7$   & $13.1$  & $15.5$  & $12.8$ & $14.3$ \\
    & \quad  $\text {SGC+B2F \cite{chen2023addressing}}$  \scriptsize \textit {(ICME’23)}  & $-$     & $37.5$    & $40.7$  & $-$ & $39.1$  & $-$     & $20.8$  & $22.1$  & $-$  & $21.5$ & $-$     & $15.0$  & $18.0$  & $-$ & $16.5$  \\
    & \quad $\text {FGPL-A \cite{FGPL-A}}$  \scriptsize \textit {(TPAMI’23)}   &$ 28.4  $&$ 38.0  $&$ \underline{42.4}  $&$ 36.1  $ & $40.0$  &$ \underline{20.5}  $&$ \underline{24.0}  $&$ \underline{25.4}  $&$ \underline{23.3}  $  & $\underline{24.7}$  &$ \underline{13.4} $&$\underline{18.0} $&$ \underline{21.0}$&$ \underline{17.5}$ & $\underline{19.5}$ \\
    & \quad $\text {EICR \cite{EICR}}$  \scriptsize \textit {(ICCV’23)}   &$ -  $&$ 34.9 $&$ 37.0  $&$ -  $ & $36.0$  &$ -  $&$ 20.8  $&$ 21.8 $&$ -  $  & $21.3$  &$ - $&$ 15.5  $&$ 18.2  $&$ -$ & $16.9$ \\
    & \quad $\text {PSCV \cite{PSCV}}$  \scriptsize \textit {(arXiv’23)}   &$ 28.1  $&$ 33.2  $&$ 35.2  $&$ 32.2 $ & $34.2$  &$ 17.1  $&$ 20.8  $&$ 21.9  $&$ 19.9  $ & $21.4$  &$ 6.4  $&$ 10.1  $&$ 13.9  $&$ 10.1$ & $12.0$  \\
    & \quad $\text {CFA \cite{CFA}}$  \scriptsize \textit {(ICCV’23)}   &$ -  $&$ 35.7  $&$ 38.2  $&$ -  $ & $37.0$  &$ -  $&$ 17.0  $&$ 18.4  $&$ -  $  & $17.7$  &$ - $&$ 13.2  $&$ 15.5  $&$ -$ & $14.4$ \\
    & \quad $\text {MEET \cite{MEET}}$  \scriptsize \textit {(ICCV’23)}   &$ -  $&$ 25.3  $&$  33.5  $&$ -  $ & $29.4$  &$ -  $&$ 19.0  $&$ 23.7  $&$ -  $  & $21.4$  &$ - $&$ 8.5  $&$ 11.8   $&$ -$ & $10.2$ \\
    & \quad $\text {HiLo \cite{HiLo}}$  \scriptsize \textit {(ICCV’23)}   &$ -  $&$ -  $&$  -  $&$ -  $ & $-$  &$ -  $&$ -  $&$ -  $&$ -  $  & $-$  &$ - $&$ 14.7  $&$ 17.7   $&$ -$ & $16.2$ \\
    & \quad  $\text {NICEST \cite{NICEST}}$  \scriptsize \textit {(TPAMI’24)}  & $-$     & $29.5$    & $31.6$  & $-$ & $30.6$  & $-$     & $15.7$  & $16.5$  & $-$  & $16.1$ & $-$     & $10.4$  & $12.4$  & $-$ & $11.4$  \\
    & \quad \textbf{RcSGG (ours)}   &  $\underline{32.6}$    &   $\underline{38.8}$    &   $41.4$    &   $\underline{37.6}$    &   $\underline{40.1}$    &   $19.8$    &   $23.2$    &   $24.1$    &   $22.4$    &   $23.7$    &   $13.3$    & $16.9$  & $19.9$  & $16.7$  & $18.4$  \\
    \cline{2-17} \noalign{\vskip 1mm}
    & VCTree (backbone) \cite{VCtree} &$12.4$ 	&$15.4$ 	&$16.6$ 	&$14.8$ &$16.0$ &$6.3$ 	&$7.5$ 	&$8.0$ 	&$7.3$  &$7.8$ &$4.9$ 	&$6.6$ 	&$7.7$ 	&$6.4$ &$7.2$  \\
    & \quad $\text {HML \cite{HML}}$  \scriptsize \textit {(ECCV’22)}  &$31.0 	$&$36.9 	$&$39.2 	$&$35.7 	$ &$30.1$ &$20.5 	$&$25.0 	$&$26.8 	$&$24.1 	$ &$25.9$ &$10.1 	$&$13.7 	$&$16.3 	$&$13.4$ &$15.0$  \\
    & \quad $\text {FGPL \cite{FGPL}}$  \scriptsize \textit {(CVPR’22)}  &$30.8 	$&$37.5 	$&$40.2 	$&$36.2 	$ &$38.9$ &$21.9 	$&$26.2 	$&$27.6 	$&$25.2 	$ &$26.9$ &$11.9 	$&$16.2 	$&$19.1 	$&$15.7$ &$17.7$   \\
    & \quad $\text {TransRwt \cite{TransRwt}}$  \scriptsize \textit {(ECCV’22)} &$- 	$&$37.0 	$&$39.7 	$&$- 	$ &$38.4$ &$- 	$&$19.9 	$&$21.8 	$&$- 	$ &$20.9$ &$- 	$&$12.0 	$&$14.9 	$&$-$  &$13.5$ \\
    & \quad $\text {PPDL \cite{PPDL}}$  \scriptsize \textit {(CVPR’22)} &$- 	$&$33.3 	$&$33.8 	$&$- 	$ &$33.6$ &$- 	$&$21.8 	$&$22.4 	$&$- 	$ &$22.1$ &$- 	$&$11.3 	$&$13.3 	$&$-$ &$12.3$  \\
    & \quad $\text {GCL \cite{GCL}}$  \scriptsize \textit {(CVPR’22)} &$31.4 	$&$37.1 	$&$39.1 	$&$35.9 	$ &$37.5$ &$19.5 	$&$22.5 	$&$23.5 	$&$21.8 	$ &$23.0$ &$11.9 	$&$15.2 	$&$17.5 	$&$14.9$ &$16.4$   \\
     & \quad $\text {NICE \cite{NICE}}$  \scriptsize \textit {(CVPR’22)} &$- 	$&$30.7 	$&$33.0 	$&$- 	$ &$31.9$ &$- 	$&$19.9 	$&$21.3 	$&$- 	$ &$20.6$ &$- 	$&$11.9 	$&$14.1 	$&$-$ &$13.0$    \\
    & \quad $\text {RTPB \cite{RTPB}}$  \scriptsize \textit {(AAAI’22)}  &$27.3 	$&$33.4 	$&$35.6 	$&$32.1 	$ &$34.5$ &$20.6 	$&$24.5 	$&$25.8 	$&$23.6 	$ &$25.2$ &$9.6 	$&$12.8 	$&$15.1 	$&$12.5$ &$14.0$  \\
    & \quad  $\text {SGC+B2F \cite{chen2023addressing}}$  \scriptsize \textit {(ICME’23)}  & $-$     & $36.2$    & $40.4$  & $-$ & $38.3$  & $-$     & $22.5$  & $25.1$  & $-$  & $23.8$ & $-$     & $14.6$  & $17.6$  & $-$ & $16.1$  \\
    & \quad $\text {FGPL-A \cite{FGPL-A}}$  \scriptsize \textit {(TPAMI’23)}   &$ \underline{34.5} $&$ \underline{41.6} $&$ \underline{44.3} $&$ \underline{40.1} $ & $\underline{43.0}$  &$ \underline{24.1}$&$ \underline{28.8}  $&$ \underline{30.6} $&$ \underline{27.8} $  & $\underline{29.7}$  &$ 12.2 $&$ 16.9 $&$ 20.0  $&$ 16.4$ & $18.5$ \\
    & \quad $\text {EICR \cite{EICR}}$  \scriptsize \textit {(ICCV’23)}   &$ -  $&$ 35.6  $&$ 37.9  $&$ -  $ & $36.8$  &$ -  $&$ 26.2  $&$ 27.4 $&$ -  $  & $26.8$  &$ - $&$ 15.2  $&$ 17.5  $&$ -$ & $16.4$ \\
     & \quad $\text {PSCV \cite{PSCV}}$  \scriptsize \textit {(arXiv’23)}   &$ 30.2  $&$ 34.8  $&$ 37.2  $&$ 34.1 $  &$36.0$  &$ 22.1  $&$ 25.8  $&$ 27.1  $&$ 25.0  $  &$26.5$  &$ 6.5  $&$ 9.9  $&$ 13.3  $&$ 9.9$ &$11.6$ \\
    & \quad $\text {CFA \cite{CFA}}$  \scriptsize \textit {(ICCV’23)}   &$ -  $&$ 34.5  $&$ 37.2  $&$ -  $ & $35.6$  &$ -  $&$ 19.1  $&$ 20.8  $&$ -  $  & $20.0$  &$ - $&$ 13.1  $&$ 15.5  $&$ -$ & $14.3$ \\
    & \quad $\text {MEET \cite{MEET}}$  \scriptsize \textit {(ICCV’23)}   &$ -  $&$ 25.5  $&$   34.5  $&$ -  $ & $30.0$  &$ -  $&$ 14.5  $&$  18.6  $&$ -  $  & $16.6$  &$ - $&$ 8.2  $&$ 11.5   $&$ -$ & $9.9$ \\
    & \quad $\text {HiLo \cite{HiLo}}$  \scriptsize \textit {(ICCV’23)}   &$ -  $&$ -  $&$  -  $&$ -  $ & $-$  &$ -  $&$ -  $&$ -  $&$ -  $  & $-$  &$ - $&$ 12.9  $&$ 15.2   $&$ -$ & $14.1$ \\
    & \quad  $\text {NICEST \cite{NICEST}}$  \scriptsize \textit {(TPAMI’24)}  & $-$     & $30.6$    & $32.9$  & $-$ & $31.8$  & $-$     & $18.9$  & $20.0$  & $-$  & $19.5$ & $-$     & $10.2$  & $11.9$  & $-$ & $11.1$  \\
    & \quad \textbf{RcSGG (ours)}   &  $33.7$     &   $38.9$    &    $42.4$   &    $38.3$   &   $40.7$    &    $24.3$   &   $27.4$    &   $29.6$    &  $27.1$     &   $28.5$    &   $\underline{12.6}$    & $\underline{17.3}$ &   $\underline{20.4}$ & $\underline{16.8}$ &   $\underline{18.9}$ \\
    \cline{2-17} \noalign{\vskip 1mm}
    & Transformer (backbone) \cite{transformer} &$ 12.4  $&$ 16.0  $&$ 17.5  $&$ 15.3  $ &$16.8$ &$ 7.7   $&$ 9.6     $&$ 10.2   $&$ 9.2   $ &$9.9$ &$ 5.3   $&$ 7.3   $&$ 8.8   $&$ 7.1$ &$8.1$  \\
    & \quad $\text {HML \cite{HML}}$  \scriptsize \textit {(ECCV’22)} &$ 30.1  $&$ 36.3  $&$ 38.7  $&$35.0  $ &$37.5$ &$ 17.1  $&$ 20.8  $&$ 22.1  $&$ 20.0  $ &$21.5$ &$ 10.8  $&$ 14.6  $&$ 17.3  $&$ 14.2$ &$16.0$  \\
    & \quad $\text {FGPL \cite{FGPL}}$  \scriptsize \textit {(CVPR’22)}  &$ 24.3  $&$ 33.0  $&$ 37.5  $&$ 31.6  $ &$35.3$ &$ 17.1  $&$ 21.3  $&$ 22.5  $&$ 20.3  $ &$21.9$ &$ 11.1  $&$ 15.4  $&$ 18.2  $&$ 14.9$ &$16.8$  \\
    & \quad $\text {TransRwt \cite{TransRwt}}$  \scriptsize \textit {(ECCV’22)}  &$ -     $&$ 35.8  $&$ 39.1  $&$ -  $ &$37.5$ &$ -     $&$ 21.5  $&$ 22.8  $&$ -  $ &$22.2$ &$ -     $&$ 15.8  $&$ 18.0  $&$ -$ &$16.9$  \\
    & \quad $\text {FGPL-A \cite{FGPL-A}}$  \scriptsize \textit {(TPAMI’23)}   &$ 27.2 $&$ 36.3 $&$  40.7 $&$ 34.7 $ & $38.5$  &$ \underline{19.9} $&$ \underline{23.2}  $&$ \underline{24.5}  $&$ \underline{22.5}  $  & $\underline{23.9}$  &$ 12.5 $&$ 17.0  $&$ 19.8  $&$ 16.4$ & $18.4$ \\
    & \quad $\text {EICR \cite{EICR}}$  \scriptsize \textit {(ICCV’23)}   &$ -  $&$ 36.9 $&$ 39.1  $&$ -  $ & $38.0$  &$ -  $&$ 21.6  $&$ 22.4 $&$ -  $  & $22.0$  &$ - $&$ 17.3  $&$ 19.7  $&$ -$ & $18.5$ \\
    & \quad $\text {PSCV \cite{PSCV}}$  \scriptsize \textit {(arXiv’23)}   &$ 30.2  $&$ 34.8  $&$ 36.4  $&$ 33.8  $ &$35.6$ &$ 18.6  $&$ 21.3  $&$ 22.2  $&$ 20.7  $ &$21.8$ &$ 7.3  $&$ 11.1  $&$ 14.5  $&$ 11.0$ &$12.8$  \\
    & \quad $\text {CFA \cite{CFA}}$  \scriptsize \textit {(ICCV’23)}   &$ -  $&$ 30.1  $&$ 33.7  $&$ -  $ & $31.9$  &$ -  $&$ 15.7  $&$ 17.2  $&$ -  $  & $16.5$  &$ - $&$ 12.3  $&$ 14.6  $&$ -$ & $13.5$ \\
    & \quad $\text {HiLo \cite{HiLo}}$  \scriptsize \textit {(ICCV’23)}   &$ -  $&$ -  $&$  -  $&$ -  $ & $-$  &$ -  $&$ -  $&$ -  $&$ -  $  & $-$  &$ - $&$ 14.6  $&$ 17.6   $&$ -$ & $16.1$ \\
    & \quad \textbf{RcSGG (ours)}   &  $\underline{32.4}$    &   $\underline{39.6}$    &   $\underline{41.3}$    &  $\underline{37.8}$     & $\underline{40.5}$      &   $\underline{19.9}$    &   $22.3$    &   $24.4$    &   $22.2$    &   $23.4$    &  $\underline{13.3}$     & $\underline{18.2}$  &   $\underline{20.9}$    &   $\underline{17.5}$    & $\underline{19.6}$ \\

    \midrule
    \midrule
    \multirow{13}{*}{\rotatebox{90}{GQA \cite{GQAdataset}}} & MotifsNet (backbone) \cite{Neuralmotifs}  &$11.7$& $15.4$ & $16.3$  & $14.5$  & $15.9$  & $6.8$   & $8.6$     & $9.4$   & $8.3$ & $9.0$  & $4.4$   & $6.8$   & $8.6$   & $6.6$ & $7.7$ \\
    & \quad GCL \cite{GCL} \scriptsize \textit {(CVPR’22)} &-  & $36.7$  & $38.1$  & -  & $37.4$  & -   & $\underline{17.3}$     & $18.1$   & - & $\underline{17.7}$  & -   & $\underline{16.8}$   & $\underline{18.8}$   & - & $\underline{17.8}$ \\
    & \quad CFA \cite{CFA} \scriptsize \textit {(ICCV’23)}  &-  & $31.7$ & $33.8$  & -  & $32.8$  & -   & $14.2$     & $15.2$   & - & $14.7$  & -   & $11.6$   & $13.2$   & - & $12.4$ \\
    & \quad EICR \cite{EICR} \scriptsize \textit {(ICCV’23)}  &-  & $ 36.3$ & $38.0$  & -  & $37.2$  & -   & $17.2$     & $18.2$   & - & $\underline{17.7}$  & -   & $ 16.0$   & $18.0$   & - & $17.0$ \\
    & \quad \textbf{RcSGG (ours)}   &$\underline{30.3} $ &$\underline{36.9} $ &$ \underline{38.5}$&$ \underline{35.2}$ & $ \underline{37.7}$  & $ \underline{12.2}$   & $16.7$     & $\underline{18.4}$   & $\underline{15.8}$ & $17.6$  & $\underline{11.3}$   & $15.8$   & $17.6$   & $\underline{14.9}$ & $16.7$ \\
    \cline{2-17} \noalign{\vskip 1mm}
    & VCTree (backbone) \cite{VCtree} &$11.9$  & $14.3$ & $16.2$  & $14.1$  & $15.3$  & $6.8$   & $7.3$     & $7.8$   & $7.3$ & $7.6$  & $3.8$   & $6.1$   & $7.4$   & $5.7$ & $6.8$ \\
    & \quad GCL \cite{GCL} \scriptsize \textit {(CVPR’22)}  &-  & $35.4$ & $36.7$  & -  & $36.1$  & -   & $17.3$     & $18.0$   & - & $17.7$  & -   & $15.6$   & $17.8$   & - & $16.7$ \\
    & \quad CFA \cite{CFA} \scriptsize \textit {(ICCV’23)} &-  & $33.4$ & $35.1$  & -  & $34.3$  & -   & $14.1$     & $15.0$   & - & $14.6$  & -   & $10.8$   & $12.6$   & - & $11.7$ \\
    & \quad EICR \cite{EICR} \scriptsize \textit {(ICCV’23)}  &-  & $35.9$ & $37.4$  & -  & $36.7$  & -   & $17.8$     & $18.6$   & - & $18.2$  & -   & $ 14.7$   & $16.3$   & - & $15.5$ \\
    & \quad \textbf{RcSGG (ours)}   &$\underline{32.1}$ & $\underline{37.8}$ & $\underline{40.2}$  & $\underline{36.7}$  & $\underline{39.0}$  & $\underline{12.8}$   & $\underline{18.2}$     & $\underline{19.4}$   & $\underline{16.8}$ & $\underline{18.8}$  & $\underline{11.9}$   & $\underline{16.8}$   & $\underline{18.3}$   & $\underline{15.7}$ & $\underline{17.6}$ \\
    \cline{2-17} \noalign{\vskip 1mm}
    & Transformer (backbone) \cite{transformer} &$12.3$  & $15.2$ & $17.7$  & $15.1$  & $16.5$  & $7.3$   & $8.4$     & $9.6$   & $8.4$ & $9.0$  & $5.1$   & $6.6$   & $7.9$   & $6.5$ & $7.3$ \\
    & \quad CFA \cite{CFA} \scriptsize \textit {(ICCV’23)}  &-  & $27.8$ & $29.4$  & -  & $28.6$  & -   & $16.2$     & $16.9$   & - & $16.6$  & -   & $13.4$   & $15.3$   & - & $14.4$ \\
    & \quad \textbf{RcSGG (ours)}   &$\underline{31.6}$  & $\underline{37.2}$ & $\underline{39.3}$  & $\underline{36.0}$  & $\underline{38.3}$  & $\underline{13.3}$   & $\underline{18.1}$     & $\underline{19.2}$   & $\underline{16.9}$ & $\underline{18.7}$  & $\underline{11.4}$  & $\underline{16.3}$   & $\underline{17.4}$   & $\underline{15.0}$ & $\underline{16.9}$ \\
    \midrule
    \midrule
    \end{tabular}}%
  \label{mRK}%
  \vspace{-0.2cm}
\end{table*}%

\subsubsection{Training and Testing}
\label{section3.4.2}
\textit {Training.} This section discusses how to incorporate reverse causal estimation into any SGG framework to eliminate spurious correlations from $X$ to $Y$. Importantly, within RcSGG, the reverse estimation and the training of classifier $f_c$ are decoupled at the batch level. For relationship features $R$ and predictions $Y$,  we denote symbols from batch $b$ as $R_b$ and $Y_b$, and symbols from batch $b+1$ as $R_{b+1}$ and $Y_{b+1}$. At the batch granularity, the causal intervention and relationship classifier training occur within separate data flows; their finer-grained structure can be described as $Y_b \rightarrow R_{b+1} \rightarrow Y_{b+1}$. Wherein, $Y_b \rightarrow R_{b+1}$ emphasizes the reverse causal estimation, while $R_{b+1} \rightarrow Y_{b+1}$ denotes training the classifier $f_c$ along the forward causality direction. Thus, for any SGG framework, by refining the feature space $R$ at the batch level using ARE and then training $f_c$ in the standard procedure; typically, we can use cross-entropy loss to supervise the training of the model. In this way, we can effectively eliminate the spurious correlation between $X$ and $Y$.  

\textit {Testing.} The reverse causality estimation requires relationship labels; however, it's important to note that this estimation process only occurs during the training phase. In other words, for the ARE-based SGG framework, during the testing phase, it suffices to follow the regular data flow, $X \rightarrow R \rightarrow Y$, to obtain predictions devoid of spurious correlations. This is because the trained classifier $f_c^*$ is a Bayesian optimal classifier, learned within a parameter space $\mathcal{F}_{\widetilde{\mathbf{R}}}$ that is already free of spurious correlations, as shown in Equation (\ref{f_c^*}).


\section{Experiments}
\subsection{Experiment Setup}
\label{section4.1}

\textit {Implementations.} We conducted extensive experiments across three datasets, VG150 \cite{VG150}, GQA \cite{GQAdataset}, Open Images V6 (OI V6) \cite{openimages}, and \textbf{PSG}, with details outlined below.

\textbf{VG150} \cite{VG150} is a subset of the VG dataset \cite{VG} comprising 150 object categories and 50 relationship categories. We follow the split in \cite{TDE}, \ie, 62k training images, 5k validation images, and 26k test images. 

\textbf{GQA} \cite{GQAdataset} is a large-scale visual question answering dataset with images from the VG dataset and balanced question-answer pairs. Following \cite{GCL}, we selected the top-200 object classes and top-100 relationship classes, adhering to its split of 70\% for training (including a 5k validation set) and 30\% for testing. 

\textbf{OI V6} \cite{openimages} is a large dataset supporting image classification, object detection, and visual relationship tasks. Following the settings in \cite{BGNN}, it includes 601 object categories and 30 relationship categories, comprising 126K training images, 2K validation images, and 5K test images.

\textbf{PSG} \cite{PSG} is a panoptic SGG dataset designed to overcome the limitations of detection-based approaches in existing SGG datasets. Unlike VG150, PSG utilizes pixel-level annotations to achieve fine-grained object localization and reduce noise from object overlaps. PSG dataset comprises 49k images, 133 object categories, and 56 relationship categories, supporting segmentation-based scene graph tasks.

For all the datasets mentioned here, all experiments are conducted following \cite{TDE} and most of its settings: 1) The detector in the framework is the Faster R-CNN \cite{fasterrcnn} with the ResNeXt-101-FPN backbone \cite{resnet}, pre-trained on the VG training set and achieving 28.14 mAP on the VG test set. The detector is frozen during the training of the SGG relationship classifier. 2) Three popular SGG frameworks are taken including MotifsNet \cite{Neuralmotifs}, VCTree \cite{VCtree}, and Transformer \cite{transformer}. For MotifsNet and VCTree, the batch size and initial learning rate are set to 12 and 0.01, respectively, while for Transformer, they are set to 16 and 0.001. 3) For training epochs and learning rate adjustments, the maximum number of iterations for MotifsNet and VCTree is set to 50k, with learning rates decaying at the 30k-th and 45k-th iterations. For Transformer, the maximum number of iterations is set to 16k, with learning rates decaying at the 10k-th and 16k-th iterations. 4) The hyper-parameter settings in RcSGG are $\pi=3$, $K^{\prime} =44$, $\alpha=0.2$, and $\lambda=0.01$. Detailed ablation studies for these parameters are presented in Section \ref{section4.3}.

\textit {Backbones and baselines.} Experiments are conducted on three classical SGG frameworks: 1) MotifsNet \cite{Neuralmotifs}, 2) VCTree \cite{VCtree}, and 3) Transformer \cite{transformer}. Baseline debiasing methods covering resampling methods (\eg TransRwt \cite{TransRwt}), reweighting methods (\eg PPDL \cite{PPDL}), adjustment methods (\eg RTPB \cite{RTPB}), and hybrid methods (\eg NICE \cite{NICE}).

\textit {Evaluation modes and metrics.} For the \textbf{VG150} \cite{VG150} and \textbf{GQA} \cite{GQAdataset}, we use three standard evaluation modes from \cite{Neuralmotifs,VCtree,TDE}: 1) Predicate Classification (PredCls), predicting relationships given object labels and bounding boxes. 2) Scene Graph Classification (SGCls), predicting both object labels and relationships given bounding boxes. 3) Scene Graph Detection (SGDet), predicting object labels, bounding boxes, and relationships. We employ three evaluation metrics: 1) Recall rate (R@K), measuring overall relationship recall; 2) Mean recall rate (mR@K), averaging recall across different relationship categories; 3) Mean of R@K and mR@K (MR@K), which calculates the average of R@K and mR@K.  In these metrics, "K" signifies the number of relationship predictions for an individual image, commonly set to 20, 50, or 100. Typically, biased SGG scores high on R@K but falls short on mR@K, largely due to its predilection for a few predominant categories, which occupy a substantial portion; however, debiased SGG shows the inverse trend \cite{TDE,NICE,CFA,BPLSA}. Thus, MR@K helps evaluate the model's ability to balance between frequent and rare categories.

For \textbf{OI V6} \cite{openimages}, we report $\text{R@50}$ and $\text{mR@50}$, along with weighted mean AP of relationships ($\text{wmAP}_{rel}$) and weighted mean AP of phrase. Following standard evaluation procedures, we also report $\text{score}_{wtd}$, whose computation is defined as: $\text{score}_{wtd} = 0.2 \times \text{R@50} + 0.4 \times \text{wmAP}_{rel} + 0.4 \times \text{wmAP}_{phr}$.

\begin{table*}[htbp]
  \centering
  \vspace{-0.3cm}
  \caption{Quantitative results (R@K) of RcSGG and baseline methods on the MotifsNet, VCTree, and Transformer backbones. $\text{AVG}_{\text{R}}^{\Delta}$ denotes the mean of R@20, R@50, and R@100, while $\text{AVG}_{\text{R}}^{\Diamond}$ represents the mean of R@50 and R@100.}
  \vspace{-0.3cm}
    \resizebox{19.5cm}{!}{\begin{tabular}{c|l|ccccc|ccccc|ccccc}
    \toprule
          &   & \multicolumn{5}{c|}{PredCls}  & \multicolumn{5}{c|}{SGCls}    & \multicolumn{5}{c}{SGDet} \\
          &   & \multicolumn{1}{c}{R@20} & \multicolumn{1}{c}{R@50} & \multicolumn{1}{c}{R@100} & \multicolumn{1}{c}{$\text{AVG}_{\text{R}}^{\Delta}$} & \multicolumn{1}{c|}{$\text{AVG}_{\text{R}}^{\Diamond}$} & \multicolumn{1}{c}{R@20} & \multicolumn{1}{c}{R@50} & \multicolumn{1}{c}{R@100} & \multicolumn{1}{c}{$\text{AVG}_{\text{R}}^{\Delta}$} & \multicolumn{1}{c|}{$\text{AVG}_{\text{R}}^{\Diamond}$} & \multicolumn{1}{c}{R@20} & \multicolumn{1}{c}{R@50} & \multicolumn{1}{c}{R@100} & \multicolumn{1}{c}{$\text{AVG}_{\text{R}}^{\Delta}$} & \multicolumn{1}{c}{$\text{AVG}_{\text{R}}^{\Diamond}$} \\
    \midrule
    \midrule
    \multirow{28}{*}{\rotatebox{90}{VG150 \cite{VG150}}} 
    & SSRCNN-SGG \cite{SSRCNN-SGG} \scriptsize \textit {(CVPR’22)}  &  $ - $    &   $ - $    &   $ - $    &  $ - $     & $ - $      &   $ - $    &   $ - $    &   $ - $    &   $ - $    &   $ - $    &  $ 25.8 $     & $ 32.7 $  &   $ 36.9 $    &   $ 31.8 $    & $ 34.8 $ \\
    & HetSGG \cite{HetSGG} \scriptsize \textit {(AAAI’23)}  &  $ - $    &   $ 57.1 $    &   $ 59.4 $    &  $ - $     & $ 58.3 $      &   $ - $    &   $ 37.6 $    &   $ 38.5 $    &   $ - $    &   $ 38.1 $    &  $ - $     & $ 30.2 $  &   $ 34.5 $    &   $ - $    & $ 32.4 $ \\
    & SQUAT \cite{SQUAT} \scriptsize \textit {(CVPR’23)}   &  $ - $    &   $ 55.7 $    &   $ 57.9 $    &  $ - $     & $ 56.8  $      &   $ - $    &   $ 33.1 $    &   $ 34.4 $    &   $ - $    &   $  33.8 $    &  $ - $     & $ 24.5 $  &   $ 28.9 $    &   $ - $    & $ 26.7 $ \\
    & CV-SGG \cite{CV-SGG} \scriptsize \textit {(CVPR’23)}   &  $ - $    &   $ 58.2 $    &   $ 62.4 $    &  $ - $     & $ 60.3 $      &   $  -$    &   $ - $    &   $ - $    &   $ - $    &   $ - $    &  $ - $     & $ 27.8 $  &   $ 32.0 $    &   $ - $    & $29.9 $ \\
    & PE-Net \cite{PENET} \scriptsize \textit {(CVPR’23)}   &  $ - $    &   $64.9$    &   $67.2$    &  $ - $     & $66.1$      &   $  -$    &   $ 39.4 $    &   $40.7$    &   $ - $    &   $40.1$    &  $ - $     & $ 30.7 $  &   $ 35.2 $    &   $ - $    & $33.0$ \\
    & DSDI \cite{DSDI} \scriptsize \textit {(TPAMI’23)}   &  $ 47.5$    &   $54.8$    &   $56.6$    &  $53.0$     & $55.7$      &   $30.0$    &   $35.2$    &   $36.3$    &   $33.8$    &   $35.8$    &  $20.4$     & $27.2$  &   $31.6$    &   $26.4$    & $29.4$ \\
    \cline{2-17}\noalign{\vskip 1mm}
    & MotifsNet (backbone) \cite{Neuralmotifs}  & $59.5$ &$66.0$ &$67.9$ &$64.5$ &$67.0$ &$35.8$ &$39.1$ &$39.9$ &$38.3$ &$39.5$ &$25.1$ &$32.1$ &$36.9$ &$31.4$ &$34.5$ \\
    & \quad $\text {TransRwt \cite{TransRwt}}$  \scriptsize \textit {(ECCV’22)} &$-$&$48.6	$&$50.5	$&$- 	$ &$49.6$ &$-	$&$29.4	$&$30.2	$&$- 	$ &$29.8$ &$-	$&$23.5	$&$27.2 	$&$-$ &$25.4$  \\
    & \quad $\text {PPDL \cite{PPDL}}$  \scriptsize \textit {(CVPR’22)} &$- $&$47.2 $&$47.6 $&$- $ &$47.4$ &$- $&$28.4 $&$29.3 $&$- $ &$28.9$ &$- $&$21.2 $&$23.9 $&$-$ &$22.6$ \\
    & \quad $\text {NICE \cite{NICE}}$  \scriptsize \textit {(CVPR’22)}  &$- $&$55.1 $&$57.2 $&$- $ &$56.2$ &$- $&$33.1 $&$34.0 $&$- $ &$33.6$ &$- $&$27.8 $&$31.8 $&$-$ &$29.8$ \\
    & \quad $\text {EICR \cite{EICR}}$  \scriptsize \textit {(ICCV’23)}   &$ -  $&$ 55.3  $&$ 57.4  $&$ -  $ & $56.4$  &$ -  $&$ 34.5  $&$ 35.4 $&$ -  $  & $35.0$  &$ - $&$ 27.9  $&$ 32.2  $&$ -$ & $30.1$ \\
    & \quad $\text {PSCV \cite{PSCV}}$  \scriptsize \textit {(arXiv’23)}  &$- $&$- $&$- $&$- $ &$-$ &$- $&$32.5 $&$33.6 $&$- $ &$30.1$ &$- $&$- $&$- $&$-$ &$-$   \\
    & \quad $\text {CFA \cite{CFA}}$  \scriptsize \textit {(ICCV’23)}   &$ -  $&$ 54.1  $&$ 56.6  $&$ -  $ & $55.4$  &$ -  $&$ \underline{34.9} $&$ \underline{36.1}  $&$ -  $  & $\underline{35.5}$  &$ - $&$ 24.7  $&$31.8  $&$ -$ & $29.6$ \\
    & \quad $\text {HiLo \cite{HiLo}}$  \scriptsize \textit {(ICCV’23)}   &$ -  $&$ -  $&$  -  $&$ -  $ & $-$  &$ -  $&$ -  $&$ -  $&$ -  $  & $-$  &$ - $&$ 26.2  $&$ 30.3   $&$ -$ & $28.3$ \\
    & \quad  $\text {NICEST \cite{NICEST}}$  \scriptsize \textit {(TPAMI’24)}  &$- $&$\underline{59.1} $&$\underline{61.0} $&$- $ &$\underline{60.1}$ &$- $&$34.4 $&$35.2 $&$- $ &$34.8$ &$- $&$\underline{28.0} $&$\underline{32.4} $&$-$ &$\underline{30.2}$ \\
    & \quad \textbf{RcSGG (ours)}   &  $\underline{44.6}$     &   $54.3$    &   $57.3$    &    $\underline{52.1}$   &    $55.8$   &    $\underline{26.4}$   &   $32.8$    &  $33.1$     &  $\underline{30.8}$     &     $33.0$  &   $\underline{22.2}$    &  $24.8$   &   $27.9$    &    $\underline{25.0}$   &  $26.4$ \\
    \cline{2-17}\noalign{\vskip 1mm}
    & VCTree (backbone) \cite{VCtree}  & $59.8$ &$66.2$ &$68.1$ &$64.7$ &$67.2$ &$37.0$ &$40.5$ &$41.4$ &$39.6$ &$41.0$ &$24.7$ &$31.5$ &$36.2$ &$30.8$ &$33.9$ \\
    & \quad $\text {TransRwt \cite{TransRwt}}$  \scriptsize \textit {(ECCV’22)} &$-	$&$48.0	$&$49.9	$&$- 	$ &$49.0$ &$-	$&$30.0	$&$30.9	$&$- 	$ &$30.5$ &$-	$&$23.6	$&$27.8	$&$-$ &$25.7$  \\
    & \quad $\text {PPDL \cite{PPDL}}$  \scriptsize \textit {(CVPR’22)} &$-$&$47.6$&$48	$&$- $ &$47.8$ &$-$&$32.1	$&$33$&$- $ &$32.6$ &$-	$&$20.1	$&$22.9	$&$-$ &$21.5$ \\

    & \quad $\text {NICE \cite{NICE}}$  \scriptsize \textit {(CVPR’22)}  &$- $&$55.0 $&$56.9 $&$- $ &$56.0$ &$- $&$37.8 $&$39.0 $&$- $ &$38.4$ &$- $&$27.0 $&$30.8 $&$-$ &$28.9$ \\
    & \quad $\text {EICR \cite{EICR}}$  \scriptsize \textit {(ICCV’23)}   &$ -  $&$ 56.0  $&$ 57.9  $&$ -  $ & $57.0$  &$ -  $&$ 39.4  $&$ 40.5 $&$ -  $  & $40.0$  &$ - $&$ 26.0  $&$ 30.1  $&$ -$ & $28.1$ \\
    & \quad $\text {PSCV \cite{PSCV}}$  \scriptsize \textit {(arXiv’23)}    &$- $&$- $&$- $&$- $ &$-$ &$- $&$34.2 $&$35.6 $&$- $ &$34.9$ &$- $&$- $&$- $&$-$ &$-$ \\
    & \quad $\text {CFA \cite{CFA}}$  \scriptsize \textit {(ICCV’23)}   &$ -  $&$ 54.7  $&$ 57.2  $&$ -  $ & $ 56.1 $  &$ -  $&$ \underline{42.4} $&$ \underline{43.5}  $&$ -  $  & $\underline{43.0}$  &$ - $&$ 27.1  $&$ 31.2 $&$ -$ & $29.2$ \\
    & \quad $\text {HiLo \cite{HiLo}}$  \scriptsize \textit {(ICCV’23)}   &$ -  $&$ -  $&$  -  $&$ -  $ & $-$  &$ -  $&$ -  $&$ -  $&$ -  $  & $-$  &$ - $&$ 27.1  $&$ 29.8   $&$ -$ & $28.5$ \\
    & \quad  $\text {NICEST \cite{NICEST}}$  \scriptsize \textit {(TPAMI’24)}  &$- $&$\underline{59.1} $&$\underline{60.9} $&$- $ &$\underline{60.0}$ &$- $&$38.4 $&$39.4 $&$- $ &$38.9$ &$- $&$\underline{29.0} $&$\underline{32.7} $&$-$ &$\underline{30.9}$ \\
    & \quad \textbf{RcSGG (ours)}   &    $\underline{43.3}$   &   $51.2$    &   $56.8$    &   $\underline{50.4}$    &   $54.0$    &   $\underline{25.7}$    &   $31.1$    &  $32.6$     &   $\underline{29.8}$    &   $31.9$    &   $\underline{21.6}$    &  $23.1$  &    $26.2$   &   $\underline{23.6}$    & $24.7$ \\
    \cline{2-17}\noalign{\vskip 1mm}
    & Transformer (backbone) \cite{transformer}  & $58.5$ &$65.0$ &$66.7$ &$63.4$ &$66.0$ &$35.6$ &$38.9$ &$39.8$ &$38.1$ &$39.4$ &$24.0$ &$30.3$ &$33.3$ &$29.2$ &$31.8$ \\
    & \quad $\text {TransRwt \cite{TransRwt}}$  \scriptsize \textit {(ECCV’22)} &$-$&$49.0	$&$50.8	$&$- $ &$49.9$ &$-	$&$29.6	$&$30.5	$&$- 	$ &$30.1$ &$-	$&$23.1	$&$27.1	$&$-$  &$25.1$ \\
    & \quad $\text {EICR \cite{EICR}}$  \scriptsize \textit {(ICCV’23)}   &$ -  $&$ 52.8  $&$ 54.7  $&$ -  $ & $53.8$  &$ -  $&$ 31.4  $&$ 32.4 $&$ -  $  & $31.9$  &$ - $&$ 23.7  $&$ 27.7  $&$ -$ & $25.7$ \\
    & \quad $\text {CFA \cite{CFA}}$  \scriptsize \textit {(ICCV’23)}   &$ -  $&$ \underline{59.2}  $&$ \underline{61.5}  $&$ -  $ & $\underline{60.4}$  &$ -  $&$ \underline{36.3} $&$ \underline{37.3}  $&$ -  $  & $\underline{36.8}$  &$ - $&$ \underline{27.7}  $&$ \underline{32.1}  $&$ -$ & $\underline{29.9}$ \\
    & \quad $\text {HiLo \cite{HiLo}}$  \scriptsize \textit {(ICCV’23)}   &$ -  $&$ -  $&$  -  $&$ -  $ & $-$  &$ -  $&$ -  $&$ -  $&$ -  $  & $-$  &$ - $&$ 25.6  $&$ 27.9   $&$ -$ & $26.8$ \\
    & \quad \textbf{RcSGG (ours)}   &   $\underline{42.7}$    &     $52.5$  &  $56.6$     &   $\underline{50.6}$    &   $54.6$    &   $\underline{24.7}$    &   $28.9$    &   $31.5$    &  $\underline{28.4}$     &  $30.2$     &  $\underline{21.3}$     & $24.8$  &  $27.4$     &  $\underline{24.5}$     & $26.1$ \\
    \midrule
    \midrule
    \end{tabular}}%
  \label{RK}%
  \vspace{-0.2cm}
\end{table*}%

\begin{table*}[htbp]
  \centering
  \caption{Quantitative results (MR@K) of RcSGG and baseline methods on the MotifsNet, VCTree, and Transformer backbones. ${\text{MR@K}}^{\Delta}$ denotes the mean of $\text{AVG}_{\text{mR}}^{\Delta}$ and $\text{AVG}_{\text{R}}^{\Delta}$, while ${\text{MR@K}}^{\Diamond}$ represents the mean of $\text{AVG}_{\text{mR}}^{\Diamond}$ and $\text{AVG}_{\text{R}}^{\Diamond}$.}
  \vspace{-0.3cm}
    \resizebox{19.5cm}{!}{\begin{tabular}{c|l|cccccc|cccccc|cccccc}
    \toprule
          &   & \multicolumn{6}{c|}{PredCls}  & \multicolumn{6}{c|}{SGCls}    & \multicolumn{6}{c}{SGDet} \\
          &   & \multicolumn{1}{c}{$\text{AVG}_{\text{mR}}^{\Delta}$} & \multicolumn{1}{c}{$\text{AVG}_{\text{mR}}^{\Diamond}$} & \multicolumn{1}{c}{$\text{AVG}_{\text{R}}^{\Delta}$} & \multicolumn{1}{c}{$\text{AVG}_{\text{R}}^{\Diamond}$} & \multicolumn{1}{c}{${\text{MR@K}}^{\Delta}$} & \multicolumn{1}{c|}{${\text{MR@K}}^{\Diamond}$} & \multicolumn{1}{c}{$\text{AVG}_{\text{mR}}^{\Delta}$} & \multicolumn{1}{c}{$\text{AVG}_{\text{mR}}^{\Diamond}$} & \multicolumn{1}{c}{$\text{AVG}_{\text{R}}^{\Delta}$} & \multicolumn{1}{c}{$\text{AVG}_{\text{R}}^{\Diamond}$} & \multicolumn{1}{c}{${\text{MR@K}}^{\Delta}$} & \multicolumn{1}{c|}{${\text{MR@K}}^{\Diamond}$} &\multicolumn{1}{c}{$\text{AVG}_{\text{mR}}^{\Delta}$} & \multicolumn{1}{c}{$\text{AVG}_{\text{mR}}^{\Diamond}$} & \multicolumn{1}{c}{$\text{AVG}_{\text{R}}^{\Delta}$} & \multicolumn{1}{c}{$\text{AVG}_{\text{R}}^{\Diamond}$} & \multicolumn{1}{c}{${\text{MR@K}}^{\Delta}$} & \multicolumn{1}{c}{${\text{MR@K}}^{\Diamond}$} \\
    \midrule
    \midrule
    \multirow{28}{*}{\rotatebox{90}{VG150 \cite{VG150}}} 
    & SSRCNN-SGG \cite{SSRCNN-SGG} \scriptsize \textit {(CVPR’22)}   &   $- $    &    $- $   &    $- $   &   $ -$    &   $ -$    &    $ -$   &   $- $    &   $-$    &  $ -$     &   $- $    &   $ -$    & $-$  &   $8.2 $    &    $ 9.2$   &    $31.8 $   &   $34.8$    &   $ 20.0$    &    $22.0$  \\
    & HetSGG \cite{HetSGG} \scriptsize \textit {(AAAI’23)}   &   $- $    &    $33.4 $   &    $- $   &   $ 58.3$    &   $ -$    &    $ 45.9$   &   $- $    &   $16.8 $    &  $ -$     &   $38.1 $    &   $ -$    & $ 27.5$  &   $- $    &    $ 12.5$   &    $- $   &   $32.4 $    &   $ -$    &    $22.5 $  \\
    & SQUAT \cite{SQUAT} \scriptsize \textit {(CVPR’23)}   &   $ -$    &    $32.2 $   &    $- $   &   $ 56.8$    &   $- $    &    $ 44.5$   &   $ -$    &   $18.2 $    &  $ -$     &   $33.8 $    &   $ -$    & $ 26.0$  &   $ -$    &    $15.3 $   &    $- $   &   $26.7 $    &   $ -$    &    $ 21.0$  \\
    & CV-SGG \cite{CV-SGG} \scriptsize \textit {(CVPR’23)}   &   $ -$    &    $ 34.4$   &    $ -$   &   $60.3 $    &   $ -$    &    $47.4 $   &   $ -$    &   $- $    &  $- $     &   $- $    &   $ -$    & $- $  &   $- $    &    $15.8 $   &    $- $   &   $29.9 $    &   $ -$    &    $22.9 $  \\
    & PE-Net \cite{PENET} \scriptsize \textit {(CVPR’23)}   &   $- $    &    $32.7$   &    $- $   &   $ 66.1$    &   $ -$    &    $49.4$   &   $- $    &   $18.4 $    &  $ -$     &   $40.1 $    &   $ -$    & $29.3$  &   $- $    &    $ 13.5$   &    $- $   &   $33.0$    &   $ -$    &    $23.3$  \\
    & DSDI \cite{DSDI} \scriptsize \textit {(TPAMI’23)}   &  $  32.5 $    &   $ 34.6$   & $ 53.0$      &   $55.7$    &   $42.8$    &   $45.2$    &   $19.5$    &   $20.9$    &  $ 33.8$     & $35.8$  &   $26.7$    &   $28.4$    & $10.0$  & $11.2$  & $26.4$ & $29.4$ & $18.2$ & $20.3$\\
    \cline{2-20}\noalign{\vskip 1mm}
    & MotifsNet (backbone) \cite{Neuralmotifs} & $14.8$ & $16.2$ & $64.5$ & $67.0$ & $39.7$ & $41.6$ & $8.6$ & $9.3$ & $38.3$ & $39.5$ & $23.5$ & $24.4$ & $7.0$ & $7.9$ & $31.4$ & $34.5$ & $19.2$ & $21.2$  \\
    & \quad $\text {TransRwt \cite{TransRwt}}$  \scriptsize \textit {(ECCV’22)} &$-$ &$37.5$ &$-$ &$49.6$ &$-$ &$43.6$  &$-$ &$22.2$ &$-$ &$29.8$ &$-$ &$26.0$ &$-$ &$16.9$ &$-$ &$25.4$ &$-$ &$21.2$    \\
    & \quad $\text {PPDL \cite{PPDL}}$  \scriptsize \textit {(CVPR’22)} &$-$ &$32.8$ &$-$ &$47.4$ &$-$ &$40.1$ &$-$ &$17.9$ &$-$ &$28.9$ &$-$ &$23.4$ &$-$ &$12.5$ &$-$ &$22.6$ &$-$ &$17.6$ \\

    & \quad $\text {NICE \cite{NICE}}$  \scriptsize \textit {(CVPR’22)} &$-$ &$31.1$ &$-$ &$56.2$ &$-$ &$43.7$ &$-$ &$17.3$ &$-$ &$33.6$ &$-$ &$25.5$ &$-$ &$13.3$ &$-$ &$29.8$ &$-$ &$21.6$ \\
    & \quad $\text {EICR \cite{EICR}}$  \scriptsize \textit {(ICCV’23)}   &$ -  $&$ 36.0  $&$ -  $&$ 56.4  $ & $-$  &$ 46.2  $&$ - $&$ 21.3  $&$ -  $  & $35.0$  &$ - $&$ 28.2  $&$ -  $&$ 16.9$ & $-$ & $30.1$ & $-$ & $23.5$\\
    & \quad $\text {PSCV \cite{PSCV}}$  \scriptsize \textit {(arXiv’23)}  &$32.2$ &$34.2$ &$-$ &$-$ &$-$ &$-$ &$19.9$ &$21.4$ &$-$ &$30.1$ &$-$ &$25.8$ &$10.1$ &$12.0$ &$-$ &$-$ &$-$ &$-$   \\
    & \quad $\text {CFA \cite{CFA}}$  \scriptsize \textit {(ICCV’23)}   &$ -  $&$ 37.0  $&$ -  $&$ 55.4  $ & $-$  &$ 46.2  $&$ - $&$ 17.7  $&$ -  $  & $\underline{35.5}$  &$ - $&$ 26.6  $&$ -  $&$ 14.4$ & $-$ & $\underline{29.6}$ & $-$ & $22.0$\\
    & \quad $\text {HiLo \cite{HiLo}}$  \scriptsize \textit {(ICCV’23)}   &$ -  $&$ - $&$ -  $&$ -  $ & $-$  &$ -  $&$ - $&$ -  $&$ -  $  & $-$  &$ - $&$ - $&$ -  $&$16.2$ & $-$ & $28.3$ & $-$ & $22.3$\\
    & \quad  $\text {NICEST \cite{NICEST}}$  \scriptsize \textit {(TPAMI’24)} &$-$ &$31.8$ &$-$ &$\underline{60.1}$ &$-$ &$46.0$ &$-$ &$16.1$ &$-$ &$\underline{34.8}$ &$-$ &$25.5$ &$-$ &$11.4$ &$-$ &$\underline{30.2}$ &$-$ &$20.8$ \\
    & \quad \textbf{RcSGG (ours)} & $\underline{37.6}$ & $\underline{40.1}$ & $\underline{52.1}$ & $55.8$ & $\underline{44.9}$ & $\underline{48.0}$ & $\underline{22.4}$ & $\underline{23.7}$ & $\underline{30.8}$ & $33.0$ & $\underline{26.6}$ & $\underline{28.4}$ & $\underline{16.7}$ & $\underline{18.4}$ & $\underline{25.0}$ & $26.4$ & $\underline{20.9}$ & $\underline{22.4}$ \\
    \cline{2-20}\noalign{\vskip 1mm}
    & VCTree (backbone) \cite{VCtree} & $14.8$ & $16.0$ & $64.7$ & $67.2$ & $39.8$ & $41.6$ & $7.3$ & $7.8$ & $39.6$ & $41.0$ & $23.5$ & $24.4$ & $6.4$ & $7.2$ & $30.8$ & $33.9$ & $18.6$ & $20.6$ \\
    & \quad $\text {TransRwt \cite{TransRwt}}$  \scriptsize \textit {(ECCV’22)} &$-$ &$38.4$ &$-$ &$49.0$ &$-$ &$43.7$ &$-$ &$20.9$ &$-$ &$30.5$ &$-$ &$25.7$ &$-$ &$13.5$ &$-$ &$25.7$ &$-$ &$19.6$  \\
    & \quad $\text {PPDL \cite{PPDL}}$  \scriptsize \textit {(CVPR’22)} &$-$ &$33.6$ &$-$ &$47.8$ &$-$ &$40.7$ &$-$ &$22.1$ &$-$ &$32.6$ &$-$ &$27.4$ &$-$ &$12.3$ &$-$ &$21.5$ &$-$ &$16.9$ \\

    & \quad $\text {NICE \cite{NICE}}$  \scriptsize \textit {(CVPR’22)} &$-$ &$31.9$ &$-$ &$56.0$ &$-$ &$44.0$ &$-$ &$20.6$ &$-$ &$38.4$ &$-$ &$29.5$ &$-$ &$13.0$ &$-$ &$28.9$ &$-$ &$21.0$ \\ 
    & \quad $\text {EICR \cite{EICR}}$  \scriptsize \textit {(ICCV’23)}   &$ -  $&$ 36.8  $&$ -  $&$ 57.0  $ & $-$  &$ 46.9  $&$ - $&$ 26.8  $&$ -  $  & $40.0$  &$ - $&$ 33.4  $&$ -  $&$ 16.4$ & $-$ & $28.1$ & $-$ & $22.3$\\
    & \quad $\text {PSCV \cite{PSCV}}$  \scriptsize \textit {(arXiv’23)}  &$34.1$ &$36.0$ &$-$ &$-$ &$-$ &$-$ &$25.0$ &$26.5$ &$-$ &$34.9$ &$-$ &$30.7$ &$9.9$ &$11.6$ &$-$ &$-$ &$-$ &$-$ \\
    & \quad $\text {CFA \cite{CFA}}$  \scriptsize \textit {(ICCV’23)}   &$ -  $&$ 35.6  $&$ -  $&$ 56.1  $ & $-$  &$ 45.9  $&$ - $&$ 20.0  $&$ -  $  & $\underline{43.0}$  &$ - $&$ \underline{31.5}  $&$ -  $&$ 14.3$ & $-$ & $29.2$ & $-$ & $\underline{21.8}$\\
    & \quad $\text {HiLo \cite{HiLo}}$  \scriptsize \textit {(ICCV’23)}   &$ -  $&$ - $&$ -  $&$ -  $ & $-$  &$ -  $&$ - $&$ -  $&$ -  $  & $-$  &$ - $&$ - $&$ -  $&$14.1$ & $-$ & $28.5$ & $-$ & $21.3$\\
    & \quad  $\text {NICEST \cite{NICEST}}$  \scriptsize \textit {(TPAMI’24)} &$-$ &$31.8$ &$-$ &$\underline{60.0}$ &$-$ &$45.9$ &$-$ &$19.5$ &$-$ &$38.9$ &$-$ &$29.2$ &$-$ &$11.1$ &$-$ &$\underline{30.9}$ &$-$ &$21.0$ \\
    & \quad \textbf{RcSGG (ours)} & $\underline{38.3}$ & $\underline{40.7}$ & $\underline{50.4}$ & $54.0$ & $\underline{44.4}$ & $\underline{47.4}$ & $\underline{27.1}$ & $\underline{28.5}$ & $\underline{29.8}$ & $31.9$ & $\underline{28.5}$ & $30.2$ & $\underline{16.8}$ & $\underline{18.9}$ & $\underline{23.6}$ & $24.7$ & $\underline{20.2}$ & $\underline{21.8}$ \\
    \cline{2-20}\noalign{\vskip 1mm}
    & Transformer (backbone) \cite{transformer} & $15.3$ & $16.8$ & $63.4$ & $66.0$ & $39.4$ & $41.4$ & $9.2$ & $9.9$ & $38.1$ & $39.4$ & $23.7$ & $24.7$ & $7.1$ & $8.1$ & $29.2$ & $31.8$ & $18.2$ & $20.0$ \\
    & \quad $\text {TransRwt \cite{TransRwt}}$  \scriptsize \textit {(ECCV’22)} &$-$ &$37.5$ &$-$ &$49.9$ &$-$ &$43.7$ &$-$ &$22.2$ &$-$ &$30.1$ &$-$ &$26.2$ &$-$ &$16.9$ &$-$ &$25.1$ &$-$ &$21.0$ \\
    & \quad $\text {EICR \cite{EICR}}$  \scriptsize \textit {(ICCV’23)}   &$ -  $&$ 38.0  $&$ -  $&$ 53.8  $ & $-$  &$ 45.9  $&$ - $&$ 22.0  $&$ -  $  & $31.9$  &$ - $&$ 27.0  $&$ -  $&$ 18.5$ & $-$ & $25.7$ & $-$ & $22.1$\\
    & \quad $\text {CFA \cite{CFA}}$  \scriptsize \textit {(ICCV’23)}   &$ -  $&$ 31.9  $&$ -  $&$ \underline{60.4}  $ & $-$  &$ 46.2  $&$ - $&$ 16.5  $&$ -  $  & $\underline{36.8}$  &$ - $&$ 26.7  $&$ -  $&$ 13.5$ & $-$ & $\underline{29.9}$ & $-$ & $21.7$\\
    & \quad $\text {HiLo \cite{HiLo}}$  \scriptsize \textit {(ICCV’23)}   &$ -  $&$ - $&$ -  $&$ -  $ & $-$  &$ -  $&$ - $&$ -  $&$ -  $  & $-$  &$ - $&$ - $&$ -  $&$16.1$ & $-$ & $26.8$ & $-$ & $21.5$\\
    & \quad \textbf{RcSGG (ours)} & $\underline{37.8}$ & $\underline{40.5}$ & $\underline{50.6}$ & $54.6$ & $\underline{44.2}$ & $\underline{47.6}$ & $\underline{22.2}$ & $\underline{23.4}$ & $\underline{28.4}$ & $30.2$ & $\underline{25.3}$ & $\underline{26.8}$ & $\underline{17.5}$ & $\underline{19.6}$ & $\underline{24.5}$ & $26.1$ & $\underline{21.0}$ & $\underline{22.9}$ \\
    \bottomrule
    \end{tabular}}%
  \label{M-RK}%
  \vspace{-0.2cm}
\end{table*}%

\begin{table}[htbp]
  \centering
  \vspace{-0.3cm}
  \caption{Quantitative results of RcSGG and baseline methods on OI V6 dataset.}
    \begin{tabular}{lccccc}
    \toprule
    \multicolumn{1}{c}{\multirow{2}[4]{*}{Models}} & \multicolumn{1}{r}{\multirow{2}[4]{*}{mR@50}} & \multicolumn{1}{r}{\multirow{2}[4]{*}{R@50}} & \multicolumn{2}{c}{wmAP} & \multicolumn{1}{c}{\multirow{2}[4]{*}{$\text{score}_{wtd}$}} \\
\cmidrule{4-5}          &       &       & \multicolumn{1}{l}{$rel$} & \multicolumn{1}{l}{$phr$} &  \\
    \midrule
    BGNN \cite{BGNN}  \scriptsize \textit {(CVPR’21)}   &   40.5   & 75.0     & 33.5     & 34.2     & 42.1 \\
    PCL \cite{PCL} \scriptsize \textit {(TIP’22)}     &   41.6   & 74.8     & 34.7     & 35.0     & 42.8 \\
    RU-Net \cite{Runet} \scriptsize \textit {(CVPR’22)} &   -   & 76.9     & 35.4     & 34.9     & 43.5 \\
    PE-NET \cite{PENET} \scriptsize \textit {(CVPR’23)} &   -   & 76.5     & 36.6     & 37.4     & 44.9 \\
    \scriptsize {SSRCNN-SGG} \cite{SSRCNN-SGG} \scriptsize \textit {(CVPR’22)} & 42.8   & 76.7     & 41.5     & 43.6     & 49.4 \\
    FGPL-A \cite{FGPL-A} \scriptsize \textit {(TPAMI’23)} &   -  & 73.4     & 35.9     & 36.8     & 43.8 \\
    HetSGG \cite{HetSGG} \scriptsize \textit {(AAAI’23)} &   43.2  & 74.8     & 33.5     & 34.5     & 42.2 \\
    SQUAT \cite{SQUAT} \scriptsize \textit {(CVPR’23)} &   -   & 75.8     & 34.9     & 35.9    & 43.5 \\
    \midrule
    \textbf{RcSGG (ours)} &   $43.4$   & $77.1$     & $39.7$     & $42.8$    & $48.4$ \\
    
    \bottomrule
    \end{tabular}%
  \label{OpenImagesV6}%
  \vspace{-0.3cm}
\end{table}%

\subsection{Main Results and Analysis}
\label{section4.2}
This section shows quantitative and qualitative results. The quantitative results are categorized into \textit {quantitative results excluding the background relationships} and \textit {quantitative results considering the background relationships}. The former adopts a standard evaluation procedure, where outputs related to background relationships in the final predictions are masked, highlighting our method's proficiency in addressing the head-tail bias. In contrast, the latter considers background relationships as a typical category, emphasizing our method's capability to mitigate the fore-back bias.

\textit {Quantitative results excluding the background relationships.} We report the results via different metrics in Tables~\ref{mRK},~\ref{RK},~\ref{M-RK}, and~\ref{OpenImagesV6}, respectively. The results show that: 

1) Across three popular SGG datasets, VG150 \cite{VG150}, GQA \cite{GQAdataset}, and OI V6 \cite{openimages}, RcSGG achieves state-of-the-art performance on the most unbiased SGG metric mR@K, surpassing debiasing methods built on well-known two-stage SGG backbones, as reported in Table~\ref{mRK} and ~\ref{OpenImagesV6}. For VG150, using Transformer as the backbone, our method leads by 0.6\% (FGPL-A \cite{FGPL-A}) to 7.6\% (CFA \cite{CFA}) on the mR@100 metric under the PredCls mode. For GQA, with MotifsNet as the backbone, the improvement ranges from 0.2\% (EICR \cite{EICR}) to 3. 2\% (CFA \cite{CFA}) in the SGCls mode with the same metric. For OI V6, our method outperforms others by 0.2\% (HetSGG \cite{HetSGG}) to 2.9\% (BGNN \cite{BGNN}) on the mR@50 metric. Moreover, compared to the recently popular one-stage SGG backbones, our method continues to achieve state-of-the-art performance. For instance, in the SGDet mode, RcSGG based on a Transformer backbone outperforms these one-stage approaches by 3.9\% (CV-SGG \cite{CV-SGG}) to 10.9\% (SSRCNN-SGG \cite{SSRCNN-SGG}) on the mR@100 metric. Due to partial data unavailability from some baseline methods, additional metrics $\text{AVG}_{\text{mR}}^{\Delta}$ and $\text{AVG}_{\text{mR}}^{\Diamond}$ are introduced in Table~\ref{mRK}, with RcSGG still maintaining the best performance. These results further enhance the effectiveness and robustness of our method for debiasing in SGG tasks.

\begin{figure}[t]
    \footnotesize\centering
    \centerline{\includegraphics[width=0.9\linewidth]{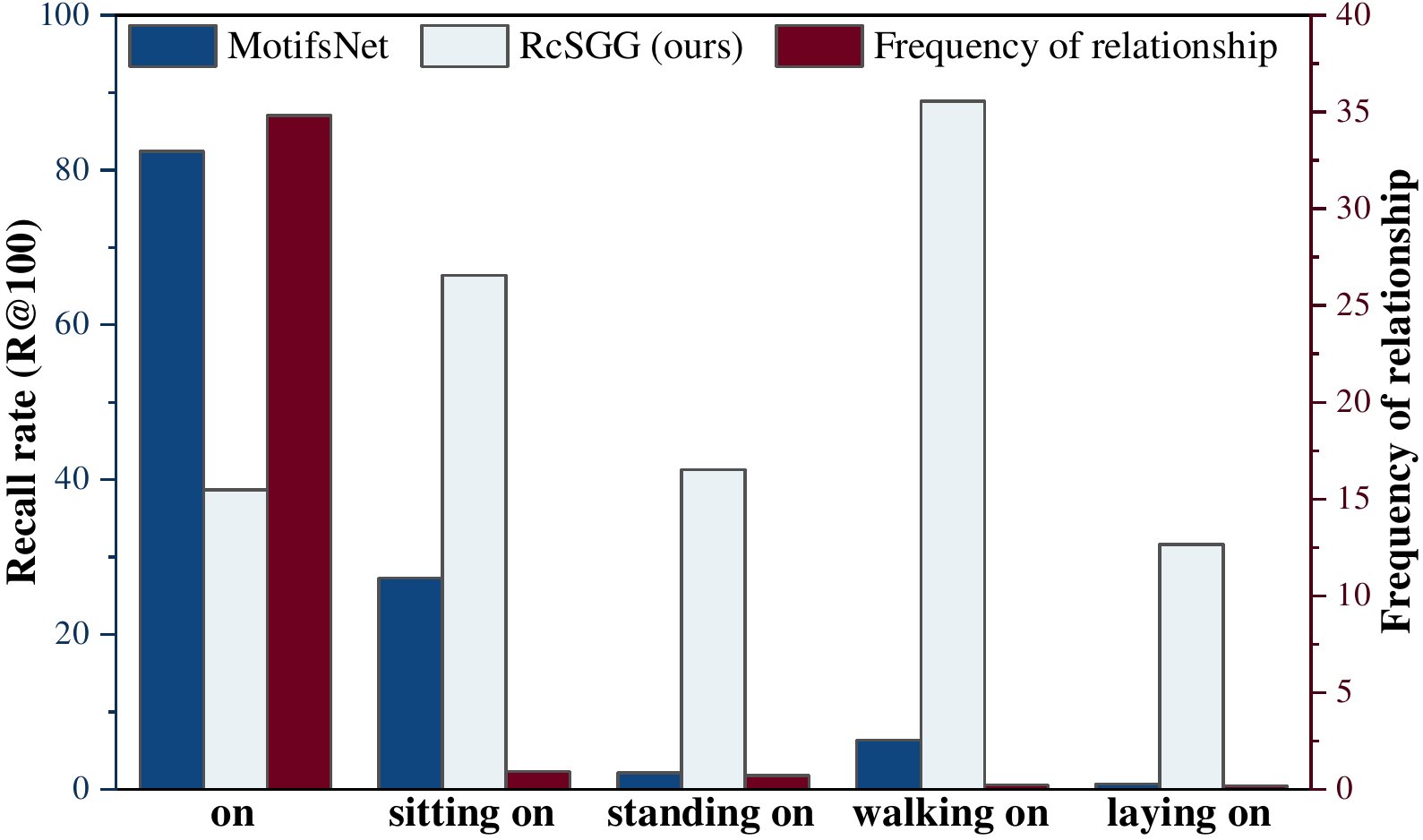}}
    \vspace{-0.3cm}
        \caption{Recall rate (R@100) for representative head and tail relationships, along with the frequency of these relationships.}
    \label{RelationshipComparison}
    \vspace{-0.3cm}
\end{figure}

\begin{table*}[htbp]
  \centering
  \vspace{-0.3cm}
  \caption{Results considering the background relationships. For differentiation, $\text{MotifsNet}^{\dagger}$, $\text{VCTree}^{\dagger}$, and $\text{Transformer}^{\dagger}$ denote the model's final predictions that take into account the output of background relationships.}
  \vspace{-0.3cm}
    \resizebox{19.5cm}{!}{\begin{tabular}{l|ccccc|ccccc|cc}
    \toprule
          & \multicolumn{12}{c}{PredCls}   \\
          & \multicolumn{1}{c}{R@20} & \multicolumn{1}{c}{R@50} & \multicolumn{1}{c}{R@100} & \multicolumn{1}{c}{$\text{AVG}_{\text{mR}}^{\Delta}$} & \multicolumn{1}{c|}{$\text{AVG}_{\text{mR}}^{\Diamond}$} & \multicolumn{1}{c}{mR@20} & \multicolumn{1}{c}{mR@50} & \multicolumn{1}{c}{mR@100} & \multicolumn{1}{c}{$\text{AVG}_{\text{mR}}^{\Delta}$} & \multicolumn{1}{c|}{$\text{AVG}_{\text{mR}}^{\Diamond}$}  &\multicolumn{1}{c}{${\text{MR@K}}^{\Delta}$} & \multicolumn{1}{c}{${\text{MR@K}}^{\Diamond}$}  \\
    \midrule
    \midrule
    $\text{MotifsNet}^{\dagger}$ (backbone) \cite{Neuralmotifs}  & $5.4$  & $10.7$  & $18.4$  & $8.1$  & $11.5$  & $0.9$   & $2.1$     & $4.3$   & $1.5$ & $2.4$  & $4.8$   & $7.0$    \\
    \quad \textbf{RcSGG (ours)}   &  $7.4$  & $12.6$  & $21.5$  & $10.0$  & $13.8$  & $5.0$   & $8.8$     & $14.9$   & $6.9$ & $9.6$  & $8.5$   & $11.7$     \\
    \midrule
    \midrule
    $\text{VCTree}^{\dagger}$ (backbone) \cite{VCtree} &$5.2$  & $9.8$  & $18.1$  & $7.5$  & $11.0$  & $1.1$   & $2.4$     & $4.7$   & $1.8$ & $2.7$  & $4.7$   & $6.9$    \\
    \quad \textbf{RcSGG (ours)}   &  $6.8$  & $10.3$  & $19.8$  & $8.6$  & $12.3$  & $5.7$   & $9.3$     & $15.7$   & $7.5$ & $10.2$  & $8.1$   & $11.3$    \\
    \midrule
    \midrule
    $\text{Transformer}^{\dagger}$ (backbone) \cite{transformer} &$5.5$  & $11.4$  & $19.2$  & $8.5$  & $12.0$  & $0.9$   & $2.2$     & $4.6$   & $1.6$ & $2.6$  & $5.1$   & $7.3$     \\
    \quad \textbf{RcSGG (ours)}   &  $6.7$  & $11.7$  & $20.3$  & $9.2$  & $12.9$  & $4.8$   & $8.4$     & $14.4$   & $6.6$ & $9.2$  & $7.9$   & $11.1$    \\
    \bottomrule
    \end{tabular}}%
  \label{relust_with_bg}%
  \vspace{-0.2cm}
\end{table*}%

\begin{figure*}[t]
    \footnotesize\centering
    \centerline{\includegraphics[width=1\linewidth]{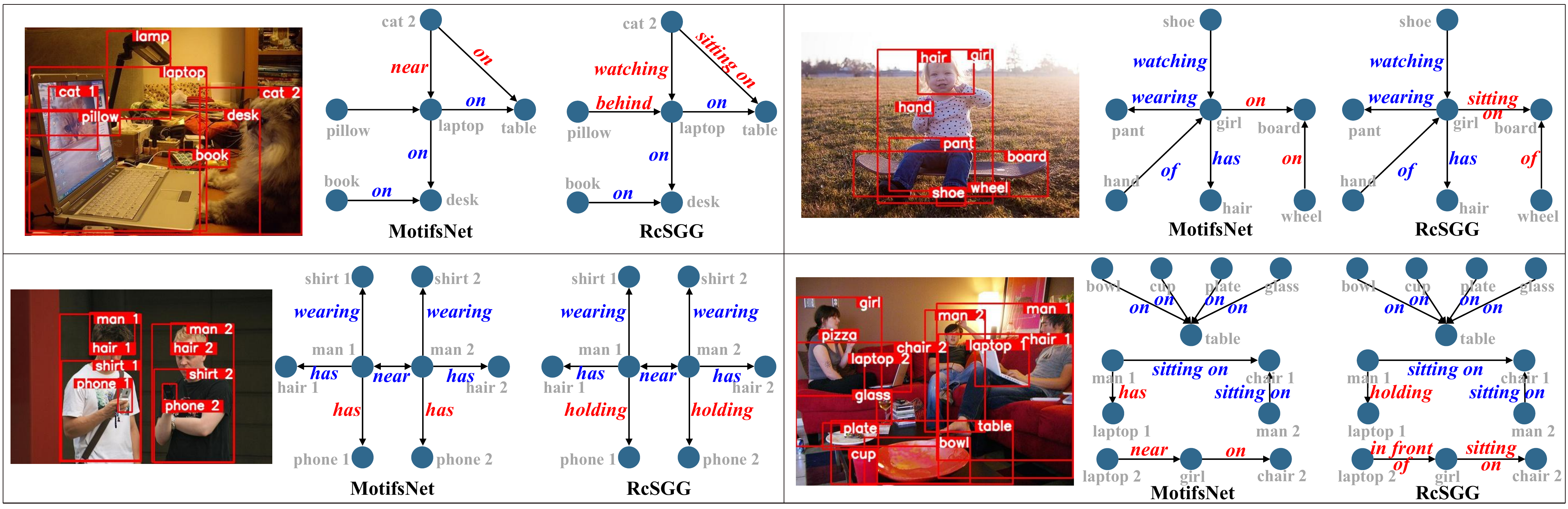}}
    \vspace{-0.3cm}
        \caption{Qualitative results of our method and baseline approach (MotifsNet \cite{Neuralmotifs}). Blue and red highlight consistent and inconsistent predictions, respectively.}
    \label{Qualitative}
    \vspace{-0.2cm}
\end{figure*}

2) RcSGG maintains competitive performance on the category-agnostic traditional metric, R@K, as shown in Table~\ref{RK}. As discussed in \cite{Neuralmotifs,VCtree,TDE,NICE}, mR@K and R@K often have an inverse relationship, due to the pursuit of mR@K inevitably impairing the performance of head relationships. Despite this, compared to other debiasing methods, RcSGG maintains a minimal decline in the R@K metric on the VG150 dataset. For instance, under the PredCls mode for the MotifsNet backbone, our method experienced only a 10.6\% decrease, whereas baseline methods PPDL \cite{PPDL} and TransRwt \cite{TransRwt} saw declines of 20.3\% and 17.4\%, respectively.

3) RcSGG demonstrates a superior balance between head and tail relationships, effectively managing the limitations of both mR@K and R@K metrics, as evidenced by the new MR@K metric—averaging R@K and mR@K—detailed in Table~\ref{M-RK}. This metric confirms that RcSGG not only enhances the balance but also significantly outperforms baselines like MotifsNet, TransRwt, and PPDL. For example, for the baseline method MotifsNet, in PredCls mode, our method outperforms TransRwt \cite{TransRwt} and PPDL \cite{PPDL} by 4.4\% and 7.9\%, respectively. 

4) RcSGG focuses on resampling foreground and background relationships, and Section \ref{section3.4.1} outlines how our method stands apart from typical reweighting and resampling techniques. Reweighting methods such as FGPL \cite{FGPL}, PPDL \cite{PPDL}, and GCL \cite{GCL}, and resampling techniques like TransRwt \cite{TransRwt}, NICE \cite{NICE}, and PSCV \cite{PSCV} are compared. Results show RcSGG significantly outperforms such approaches. We attribute this enhancement to RcSGG's ability to diversify each batch with more relationships than the batch size and dynamically adjust category influence via reverse causality. In contrast, traditional reweighting and resampling are confined to manipulating relationships within fixed batch sizes, often limited by clustered scene relationships.

\textit {Impact of background relationships on model performance.} Existing SGG evaluation protocols often overlook outputs related to background relationships. Specifically, while model predictions generate a comprehensive vector that includes both background and foreground relationships, only the foreground relationships are retained for evaluation. This strategy aims to minimize the distorting effects of the numerically dominant background relationships on performance metrics. However, this evaluation practice can obscure a significant issue: models might overly rely on background relationships, potentially undermining the learning of foreground relationships. To uncover the potential negative influence of background relationships on SGG model performance, we performed an additional evaluation that includes these outputs, as shown in Table~\ref{relust_with_bg}.

Table~\ref{relust_with_bg} demonstrates a significant drop in SGG model performance when background relationship outputs are included, confirming their disruptive influence on model learning. For example, in the PredCls evaluation mode, $\text{MotifsNet}^{\dagger}$'s mR@100 decreased from 16.8\% in the standard evaluation to 4.3\%, and $\text{Transformer}^{\dagger}$'s dropped from 17.5\% to 4.6\%. This indicates that the performance of existing models under standard evaluation protocols is actually contingent on the exclusion of background relationships. In contrast, our method significantly enhances the model’s ability to learn foreground relationships through interventions in the feature space. Even under more stringent evaluation conditions that include background relationships, our method maintains a considerable advantage, such as surpassing $\text{VCTree}^{\dagger}$ by 11.0\% on mR@100, highlighting the necessity to mitigate the impact of background relationships in SGG model training.

Incorporating background relationship outputs, our experimental findings highlight the inherent limitations of current SGG evaluation protocols. This underscores the motivation for our approach, which addresses the typical oversight of background relationships in existing models. These models might unknowingly over-rely on background relationships, thereby obscuring the presence of fore-back bias. Our method mitigates this bias through strategic interventions in the feature space, enhancing model performance under more realistic evaluation conditions. This not only confirms the significance of fore-back bias but also demonstrates the robustness of our approach in addressing this challenge.

\textit{Analyzing the Trade-off between R@K and mR@K.} Tables~\ref{mRK},~\ref{RK}, and ~\ref{M-RK} illustrate that RcSGG significantly improves mR@K across various datasets and frameworks, showcasing its strong debiasing capability. However, this improvement is accompanied by a decline in R@K compared to baseline methods. This phenomenon has been observed in many existing debiasing studies, as they inherently shift focus from high-frequency head relationships to low-frequency tail relationships, a trade-off designed to address the long-tailed distribution of relationships in SGG datasets. To better understand this trade-off, we conduct a detailed comparison of the recall rates (R@100) for several representative head and tail relationships, as shown in Figure~\ref{RelationshipComparison}. From Figure~\ref{RelationshipComparison}, we observe that the baseline method (MotifsNet) achieves higher recall for the head relationship "on." This high recall is partially due to dataset annotation practices, where annotators often simplify nuanced relationships such as "sitting on" or "laying on" into the coarse-grained relationship "on." This introduces noise into the dataset and inflates the prominence of "on." Consequently, models like MotifsNet, trained to optimize R@K, disproportionately favor such coarse-grained, high-frequency relationships.

In contrast, RcSGG demonstrates significantly higher recall on fine-grained tail relationships, such as "sitting on," "standing on," "walking on," and "laying on," despite their low frequencies. This improvement arises from RcSGG’s ARE and MIS, which redistribute the model’s attention from coarse-grained, dominant relationships to nuanced, underrepresented ones. By refining the feature space and mitigating the influence of spurious correlations, RcSGG enables more semantically accurate predictions for tail relationships, enhancing mR@K at the cost of R@K.

\textit {Qualitative results.} We show certain qualitative results in Figure \ref{Qualitative}, which demonstrate that: 

1) Our method can predict more informative tail relationships, such as \textit {$<$cat, \textbf{watching}, laptop$>$} vs. \textit {$<$cat, \textbf{near}, laptop$>$}, \textit {$<$cat, \textbf{sitting on}, table$>$} vs. \textit {$<$cat, \textbf{on}, table$>$}, and \textit {$<$man 1, \textbf{holding}, phone 1$>$} vs. \textit {$<$man 1, \textbf{has}, phone 1$>$}. We attribute this to RcSGG's proactively allowing more tail relationships to supervise model training (see Equation (\ref{add_fg})), thus reducing the model's bias towards head relationships. 

2) Our method is capable of predicting relationships that are typically overlooked, such as \textit {$<$pillow, \textbf{behind}, laptop$>$}. In the baseline methods, the prediction for this relationship surpasses the recall threshold set for all relationships, and as a result, it can be considered neglected. This misclassification often arises from the model's strong response to the background relationship, which tends to average out its response to the foreground relationships. Thus, we argue that this improvement stems from removing a large number of background relationships during training (see Equation (\ref{bg_rnd})), thereby reducing the model's bias towards background relationships. We also argue that the enhancements in the above two observations share a common origin: RcSGG's reverse causality estimation eliminates spurious correlations from $X$ to $Y$, thus reducing the two typical biases, head-tail bias and fore-back bias, presented in Section \ref{sec:introduction}.

3) Our method effectively mitigates the influence of frequent head relationships on less frequent relationships within the same scene. For example, in Figure \ref{Qualitative}, the baseline method predicts the relationship for $<$girl, chair2$>$ as \textit{$<$girl, on, chair2$>$}, influenced by the numerous "on" relationships in the scene, such as $<$bowl, on, table$>$, $<$cup, on, table$>$, $<$plate, on, table$>$, and $<$glass, on, table$>$. These frequent "on" relationships dominate the feature space, causing their features to cluster and bias the model's prediction. In contrast, RcSGG mitigates this issue by refining the feature space through ARE. ARE dynamically re-samples less frequent relationships, ensuring their features remain distinguishable. Consequently, the refined feature space prevents the $<$girl, chair2$>$ pair from being erroneously drawn toward the "on" cluster and allows for the correct prediction of the nuanced relationship \textit{$<$girl, sitting on, chair2$>$}. This improvement highlights RcSGG's capacity to overcome head-tail bias and predict fine-grained relationships accurately.


\begin{table}[t]
    \centering
    \vspace{-0.3cm}
    \caption{The impact of Parameter $\pi$.}
    \vspace{-0.3cm}
        \begin{tabular}{l|ccc}
        \toprule
              & \multicolumn{3}{c}{PredCls} \\
              & R/mR@20    & R/mR@50    & R/mR@100 \\
        \midrule
        \midrule
        $\pi$=1.0  & $50.7$ / $6.8$      &   $60.8$ / $10.2$    & $64.9$ / $12.7$ \\
        $\pi$=1.5  &  $51.3$ / $6.9$     &   $60.9$ / $10.0$    & $65.0$ / $12.4$ \\
        $\pi$=2.0  & $60.5$ / $14.0$      & $66.7$ / $17.9$      &  $68.1$ / $19.1$\\
        $\pi$=2.5  &  $60.7$ / $14.0$     &   $66.8$ / $17.7$    & $68.3$ / $19.0$ \\
        $\pi$=3.0  &  $61.0$ / $14.2$     &   $66.9$ / $18.5$    & $68.4$ / $19.6$ \\
        $\pi$=3.5  & $61.1$ / $14.6$      &  $66.8$ / $18.1$     & $68.4$ / $19.4$ \\
        \bottomrule
        \end{tabular}%
        \label{Ablate_pi}%
        \vspace{-0.2cm}
\end{table}%

\begin{table}[t]
    \centering
    \caption{The impact of Parameter $K^{\prime} $. The "\%" denotes the proportion of instances from the selected $K^{\prime}$ categories in the total number of relationships.}
    \vspace{-0.3cm}
        \begin{tabular}{l|ccc}
        \toprule
              & \multicolumn{3}{c}{PredCls} \\
              & R/mR@20    & R/mR@50    & R/mR@100 \\
        \midrule
        \midrule
        $K^{\prime}$=31 (5.0\%)  & $46.7$ / $19.3$      &  $59.3$ / $26.7$     & $61.4$ / $29.5$ \\
        $K^{\prime}$=39 (10.0\%)  & $45.9$ / $25.2$     & $58.1$ / $33.8$      & $60.2$ / $37.4$ \\
        $K^{\prime}$=42 (15.2\%)  & $45.7$ / $25.4$      &  $57.7$ / $34.1$     &  $59.6$ / $37.7$\\
        $K^{\prime}$=44 (20.8\%)  &  $45.7$ / $25.7$     &  $57.6$ / $33.9$     & $59.9$ / $37.8$ \\
        $K^{\prime}$=45 (32.1\%)  &  $48.6$ / $23.8$     &  $60.1$ / $32.7$     & $63.1$ / $35.8$ \\
        $K^{\prime}$=46 (39.2\%)  &  $48.9$ / $22.4$     &   $60.7$ / $32.1$    & $63.8$ / $34.2$  \\
        \bottomrule
        \end{tabular}%
        \label{Ablate_K}%
        \vspace{-0.2cm}
\end{table}%
            
\begin{table}[t]
    \centering
    \caption{The impact of Parameter $\alpha$.}
    \vspace{-0.3cm}
        \begin{tabular}{l|ccc}
        \toprule
              & \multicolumn{3}{c}{PredCls} \\
              & R/mR@20    & R/mR@50    & R/mR@100 \\
        \midrule
        \midrule
        $\alpha$=1.0  & $45.9$ / $25.7$    & $56.8$ / $34.2$     &  $60.2$ / $38.1$\\
        $\alpha$=0.9  & $45.9$ / $25.6$   & $57.4$ / $34.5$     &  $59.8$ / $38.2$ \\
        $\alpha$=0.8  & $46.3$ / $25.3$    & $57.1$ / $34.6$     &  $59.6$ / $38.4$\\
        $\alpha$=0.3  & $44.7$ / $32.9$   & $54.1$ / $37.4$      & $57.2$ / $41.0$ \\
        $\alpha$=0.2  & $44.6$ / $32.6$   & $54.3$ / $38.8$      & $57.3$ / $41.4$ \\
        $\alpha$=0.1  & $44.3$ / $32.5$   & $54.8$ / $39.1$      & $57.6$ / $41.2$ \\
        \bottomrule
        \end{tabular}%
    \label{Ablate_alpha}%
    \vspace{-0.2cm}
\end{table}

\subsection{Hyper-parameters and Ablations}
\label{section4.3}

\begin{figure}[t]
	\footnotesize\centering
	\centerline{\includegraphics[width=0.9\linewidth]{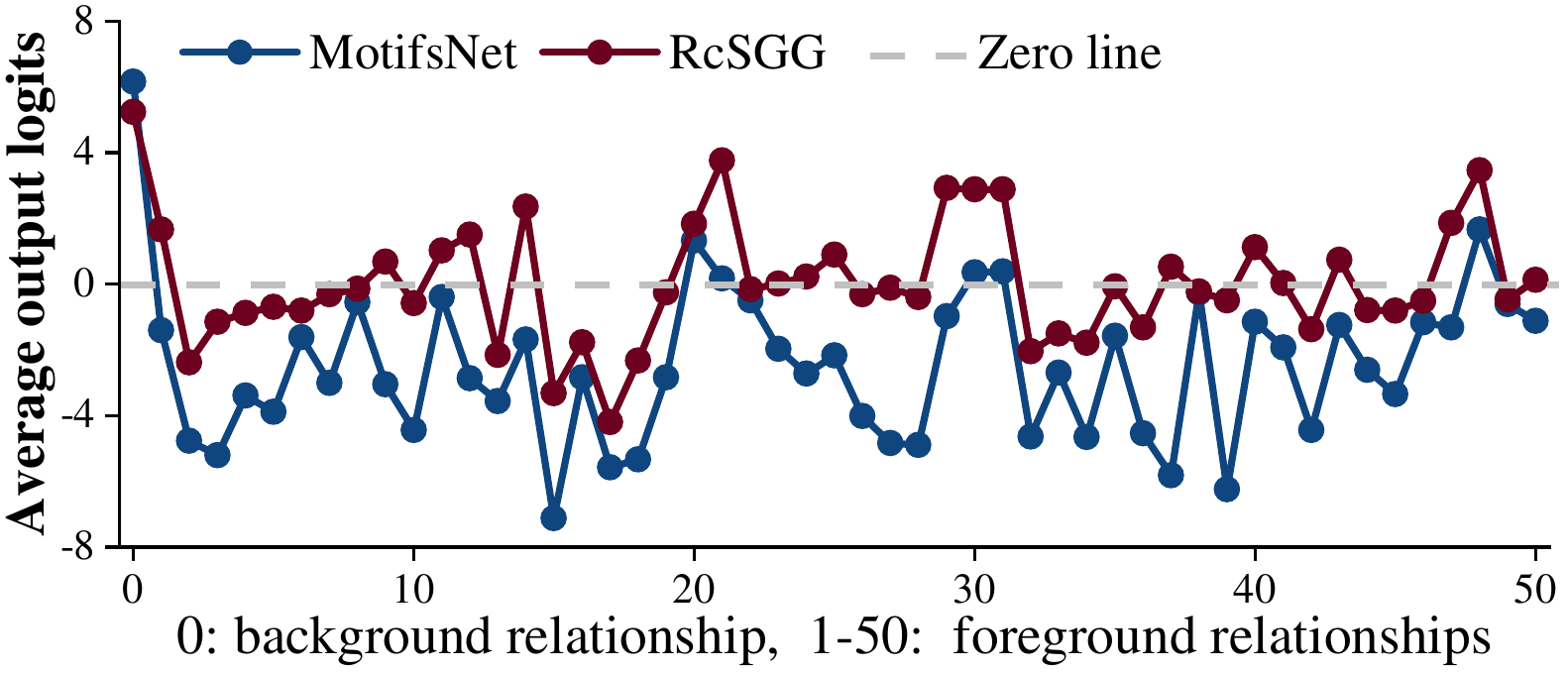}}
     \vspace{-0.3cm}
        \caption{Average output logits.}
	\label{logits}
 \vspace{-0.2cm}
\end{figure}

\textit{Ablate on hyper-parameters.} Our method involves four hyper-parameters: $\pi$, $K^{\prime}$, $\alpha$, and $\lambda$. $\lambda$ is solely for controlling the order of magnitude of sampling, determined by the approximate number of  $K^{\prime}$ relationship categories in the query set $\mathcal{Q}$, and is set to 0.01 in this paper. $\pi$, $K^{\prime}$, and $\alpha$ are critical hyper-parameters affecting RcSGG performance. We conducted extensive ablation studies in Table~\ref{Ablate_pi}, Table~\ref{Ablate_K}, Table~\ref{Ablate_alpha}, with the following observations: 

1) Table~\ref{Ablate_pi} shows that when $\pi=3$, \ie the ratio of background to foreground relationships is 3:1, the model achieves optimal performance. This indicates that the large number of background relationships due to sparse SGG annotations weakens the model's foreground relationship recognition ability, which is mitigated by removing a certain proportion of background relationships. To substantiate this, we visualized the output logits of our approach and a baseline method in Figure \ref{logits}. The results demonstrate that background relationship logits in the baseline method overwhelmingly dominate, weakening the model's response to foreground relationships (almost all negative). In contrast, our proposed approach significantly improves this unhealthy logit distribution, enhancing the model's response and recognition ability for foreground relationships. Such results further indicate that fore-back bias is a fundamental source of bias in SGG models, and that addressing only the head-tail bias is inadequate for completely eliminating model biases.

2) Table~\ref{Ablate_K} indicates that the model achieves optimal performance when $K^{\prime}=44$, \ie the number of categories in $\mathcal{Q}$ is 44. Interestingly, increasing $K^{\prime}$ further results in decline in model performance, \eg $K^{\prime}=44$ vs. $K^{\prime}=45$. We posit that this phenomenon occurs because, despite the selection of an additional category, there is an 11.3\% increase in the proportion of relationship quantities, subsequently instigating new imbalance issues within the query set $\mathcal{Q}$. 

3) Table~\ref{Ablate_alpha} reveals that the model achieves optimal performance when $\alpha=0.2$. In Equation (\ref{probability}), $\alpha$ controls the bias degree of sampling distribution; specifically, the larger $\alpha$ is, the more biased the sampling is toward high-loss categories, and vice versa. An appropriate $\alpha$ balances classification loss and sampling distribution, avoiding excessive imbalance in sampling results (\eg $\alpha=1$). 

\begin{figure}[t]
	\footnotesize\centering
	\centerline{\includegraphics[width=0.9\linewidth]{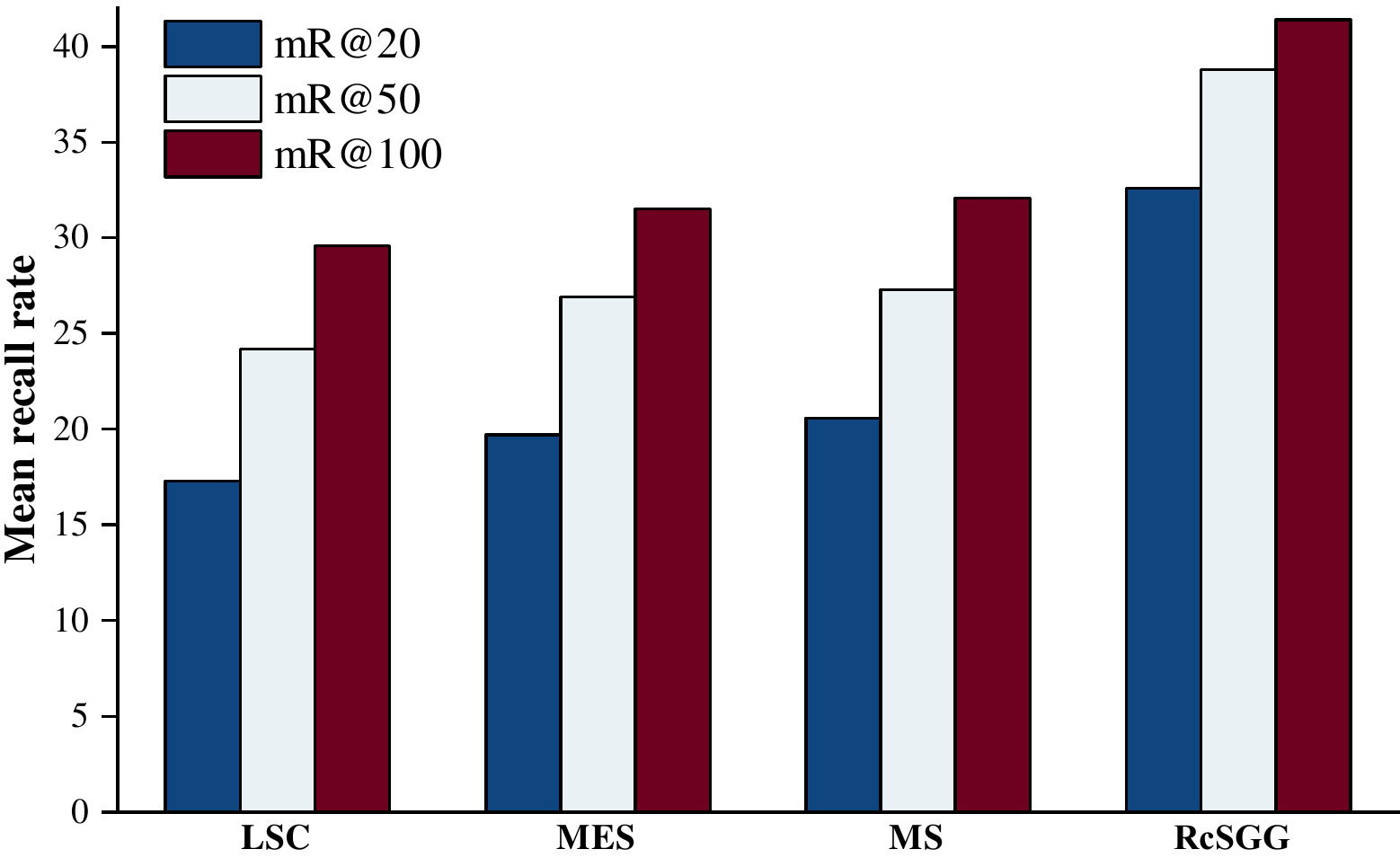}}
        \vspace{-0.3cm}
        \caption{Compared with classical uncertainty-driven active learning techniques, LSC, MES, and MS correspond to Least Confidence Sampling, Maximum Entropy Sampling, and Margin Sampling, respectively.}
	\label{AL}
 \vspace{-0.2cm}
\end{figure}

Note that due to computational resource limitations, we adopted a one-by-one parameter selection approach, with the order in Table~\ref{Ablate_pi} -~\ref{Ablate_alpha} representing the parameter tuning order. Table~\ref{Ablate_pi} is set to remove background relationships proportionally without interfering with foreground relationships. Table~\ref{Ablate_K} is set to $\pi=3$, $\alpha=1$. Table~\ref{Ablate_alpha} is set to $\pi=3$, $K^{\prime} =44$. To facilitate future generalization, we offer the following recommendations for hyperparameter selection: (1) For $\pi$, we recommend a value of $3.0$ (3:1 background-to-foreground ratio), which ensures sufficient context from background relationships without overwhelming the model's focus on foreground relationships. (2) For $K^{\prime}$, we suggest choosing a value such that the selected relationship instances account for approximately 20\% of the total instances, striking a balance between mitigating the long-tail distribution and preserving a diverse set of relationships. (3) For $\alpha$, a small value such as 0.2 is preferred, as it effectively emphasizes high-loss categories while maintaining diversity in the sampling distribution to avoid overfitting.

\textit{Ablate of core components.} RcSGG includes three key designs: Intervention on background relationships (denoted as $\xi_{\sigma}^{bg}$ for brevity), interventions on foreground relationships (denoted as $\xi_{\sigma}^{fg}$), and MIS. We investigated the contributions of these designs by controlling their hyper-parameters, with results presented in Table~\ref{Ablation_2}. The experiment demonstrates that each key design in RcSGG contributes to performance improvement. $\xi_{\sigma}^{bg}$ can simultaneously enhance both R@K and mR@K, demonstrating that interventions in background relationships can improve the recognition performance of foreground relationships, including both head and tail relationships. This improvement is attributed to our method's emphasis on learning features specific to foreground relationships. $\xi_{\sigma}^{fg}$ slightly sacrifices R@K while achieving a significant improvement in mR@K, indicating the necessity and helpfulness of approximately balanced prospect relationship within each batch, which is also emphasized in our Theorem 3. Lastly, MIS consistently helps improve mR@K, suggesting that, even for the same relationship category, enhancing sampling informativeness by focusing on object-level long-tail distribution benefits the SGG task.

\begin{table}[t]
  \centering
  \caption{Ablation of $\xi_{\sigma}^{bg}$, $\xi_{\sigma}^{fg}$, and MIS. $\sigma$ is rnd. $K$ equals 44 with $\xi_{\sigma}^{fg}$ and 0 without it, respectively.}
  \vspace{-0.3cm}
    \begin{tabular}{ccc|ccc}
    \toprule
          &       &     & \multicolumn{3}{c}{PredCls} \\
         $\xi_{\sigma}^{bg}$    & $\xi_{\sigma}^{fg}$    & MIS    & \multicolumn{1}{c}{R/mR@20} & \multicolumn{1}{c}{R/mR@50} & \multicolumn{1}{c}{R/mR@100} \\
    \midrule
    \midrule
     \ding{55}  & \ding{55}  & \ding{55}  & $59.5$ / $12.2$      &   $66.0$ / $15.5$    & $67.9$ / $16.8$  \\
    \ding{51}  & \ding{55}  & \ding{55}  &  $61.0$ / $14.2$     &   $66.9$ / $18.5$    & $68.4$ / $19.6$  \\
    \ding{55}  & \ding{51}  & \ding{55}  &  $47.1$ / $19.8$     &  $58.1$ / $27.6$     &  $61.3$ / $31.7$ \\
    \ding{55}  & \ding{51}  & \ding{51}  &   $48.8$ / $17.9$    &  $59.3$ / $25.5$     & $62.4$ / $27.4$ \\
    \ding{51}  & \ding{51}  & \ding{51}  &   $44.6$ / $32.6$    &  $54.3$ / $38.8$     & $57.3$ / $41.4$ \\
    \bottomrule
    \end{tabular}%
    \label{Ablation_2}%
    \vspace{-0.2cm}
\end{table}

\textit{Comparison with active learning techniques.}
The core component, ARE, within RcSGG, is inspired by active learning.  Active learning primarily evaluates and selects high-value unlabeled data for annotation to minimize labeling costs, as described in references \cite{EDAL,AL_1,AL_2,activelearning}. In contrast, ARE focuses on optimizing the feature space by sampling relationship features based on query conditions, with each category sampled independently and feature annotations remaining clear. While their goals and methods vary, both ARE and active learning share the same learning paradigm, model-to-data. Once ARE's query conditions are set, traditional active learning techniques like uncertainty sampling could replace MIS. To achieve this, one simply needs to replace $MIS$ with $AL_{U}$ in Equation (\ref{add_fg}):
$\mathbf{r}_{tb}^{k+} = \xi_{AL_{U}}(\mathbf{Q}_k, \widetilde{n}_{tb}^k),$
where $AL_{U}$ symbolizes the uncertainty sampling method in active learning, such as Least Confidence Sampling \cite{LCS}, Maximum Entropy Sampling \cite{MES}, and Margin Sampling \cite{MS}. The outcomes of this substitution are presented in Figure \ref{AL}. Upon replacing $MIS$ with $AL_{U}$, we observe a noticeable degradation in model performance. We argue that this decline stems from the inherent data bias issues of uncertainty-based active learning methods. Specifically, these methods tend to oversample underrepresented categories, an issue extensively discussed in the literature \cite{EDAL,AL_1,AL_2,activelearning}. Given the significant object-pair imbalance in scene graph generation tasks (see Figure \ref{object_pair}), this exacerbates the data skewness introduced by $AL_{U}$. However, our proposed method $MIS$ effectively circumvents such biases by maximizing the information of the sampling results, taking into account the distribution of object pairs during the sampling process.

\textit{Evaluation on panoptic scene graph generation dataset.} 
Recently, \cite{PSG} introduced the Panoptic Scene Graph Generation (PSG) dataset. Unlike previous datasets like VG150, PSG supports segmentation-based SGG tasks, utilizing pixel-level annotations to locate objects. To further assess the robustness of our proposed method, we followed the PSG dataset splitting as outlined in \cite{PSG} and utilized three baseline frameworks: MotifsNet, VCTree, and Transformer. The experimental parameters remained consistent with those detailed in Section \ref{section4.1}, and the results are displayed in Table~\ref{PSG}. Table~\ref{PSG} shows that: 1) Our method demonstrates strong debiasing capabilities on the PSG dataset, with increases of 9.6\%, 9.9\%, and 9.7\% on the mR@100 metric across the typical SGG frameworks MotifsNet \cite{Neuralmotifs}, VCTree \cite{VCtree}, and Transformer \cite{transformer}, respectively; 2) Our method exhibits only a minor decline on the R@K metric, specifically decreasing by 1.3\%, 1.1\%, and 1.4\% across these three typical SGG frameworks, respectively. This further demonstrates that our approach does not sacrifice the performance of head relationship category to achieve unbiased predictions; 3) Compared to other debiasing approaches, our method remains competitive, surpassing recent works PSGFormer \cite{PSG} and ADTrans \cite{ADTrans} on the mR@K metric. Although our method underperforms HiLo on the PSG dataset, it significantly surpasses HiLo on the VG150 dataset (see Table~\ref{mRK}). Crucially, HiLo optimizes the model framework, whereas our debiasing approach is model-agnostic and can complement any existing framework. Therefore, integrating RcSGG with the HiLo could potentially improve performance further. Additionally, while RcSGG was originally tailored for detection-based SGG datasets and showed strong generalization on segmentation-based datasets, adapting it specifically for the PSG dataset could further boost its effectiveness.

\begin{table}[t]
    \centering
    \vspace{-0.3cm}
    \caption{Quantitative results of RcSGG and baseline methods on PSG dataset.}
        \begin{tabular}{l|ccc}
        \toprule
              & \multicolumn{2}{c}{SGDet} \\
              & R@20 / 50 / 100  & mR@20 / 50 / 100   \\
        \midrule
        \midrule
        $\text {PSGFormer \cite{PSG}}$  \scriptsize \textit {(ECCV’22)} &$18.0/19.6/20.1$ & $14.8/17.0/17.6$ \\
        $\text {HiLo \cite{HiLo}}$  \scriptsize \textit {(ICCV’23)} &$34.1/40.7/43.0$ & $23.7/30.3/33.1$   \\
        \midrule
        MotifsNet (backbone) \cite{Neuralmotifs}  & $20.0/21.7/22.0$ & $9.1 / 9.6 / 9.7$  \\
        \quad ADTrans \cite{ADTrans} \scriptsize \textit {(AAAI’24)} & $17.1/18.6/19.0$  & $17.1/18.0/18.5$     \\
        \quad \textbf{RcSGG (ours)} &$19.4/20.3/20.7$ & $18.6/19.2/19.3$     \\
        \midrule
        VCTree (backbone) \cite{VCtree}  & $20.6/22.1/22.5$ & $9.7/10.2/10.2$  \\
        \quad ADTrans \cite{ADTrans} \scriptsize \textit {(AAAI’24)} & $17.9/19.5/19.9$ & $18.0/18.9/19.0$    \\
        \quad \textbf{RcSGG (ours)} & $20.7/21.3/21.4$& $18.8/19.4/20.1$    \\
        \midrule
        Transformer (backbone) \cite{transformer}  & $20.7/21.3/21.4$& $10.1/10.3/10.6$  \\
        \quad \textbf{RcSGG (ours)} &$21.3/22.5/22.8$ & $18.9/19.4/20.3$      \\
        \bottomrule
        \end{tabular}%
    \label{PSG}%
    \vspace{-0.2cm}
\end{table}

\begin{table}[t]
    \centering
    \vspace{-0.3cm}
    \caption{Quantitative results of RcSGG with one-stage SGG framework on VG150 dataset.}
    \vspace{-0.2cm}
        \begin{tabular}{l|ccc}
        \toprule
              & \multicolumn{3}{c}{SGDet} \\
              & mR@20/50/100  & $\text{AVG}_{\text{mR}}^{\Delta}$ & $\text{AVG}_{\text{mR}}^{\Diamond}$   \\
        \midrule
        \midrule
        SSR-CNN \cite{SSR-CNN} \scriptsize \textit {(CVPR’22)} & $6.1/8.4/10.0$ & $8.2$  & $9.2$  \\
        \quad \textbf{RcSGG (ours)} & $12.2/17.9/19.3$& $16.5$ &$18.6$    \\
        \midrule
        EGTR (backbone) \cite{EGTR} \scriptsize \textit {(CVPR’24)}  & $5.5/7.9/10.1$& $7.8$ & $9.0$ \\
        \quad \textbf{RcSGG (ours)} &$12.5/18.3/19.8$ & $16.9$ & $19.1$     \\
        \bottomrule
        \end{tabular}%
    \label{onestage}%
    \vspace{-0.2cm}
\end{table}

\textit{Evaluation within end-to-end SGG framework.} 
This paper addresses the issue of spurious correlations in two-stage SGG frameworks, which induce two typical biases: head-tail bias and fore-back bias. To mitigate these issues, we introduce RcSGG, restructuring the causal chain into a reverse causal structure to eliminate spurious correlations. Our approach, ARE, optimizes relationship feature spaces for unbiased predictions. To assess the generalizability of RcSGG, we have extended its application to the end-to-end SGG paradigm, which combines object detection and relationship classification in a unified framework. Although end-to-end frameworks eliminate the explicit separation between object detection and relationship classification, they still implicitly construct relationship feature spaces during their inference processes. This allows RcSGG to optimize these spaces effectively. Experimental results in Table~\ref{onestage} demonstrate that integrating RcSGG into end-to-end frameworks improves the performance of unbiased metrics. For example, applying RcSGG to SSR-CNN \cite{ST-SGG} and EGTR \cite{EGTR} led to substantial gains in the unbiased metric $\text{AVG}_{\text{mR}}^{\Diamond}$, achieving improvements of 9.4\% and 10.1\%, respectively. These results validate the generalization capability of RcSGG across different SGG paradigms. While end-to-end frameworks do not explicitly exhibit the spurious correlations addressed in this paper, they inherit biased feature spaces from the underlying datasets. RcSGG effectively mitigates these biases, highlighting its robustness and applicability beyond the two-stage paradigm.

\begin{table}[t]
    \centering
    \caption{Model inference time of our proposed RcSGG and baseline frameworks on VG150 dataset. The time reported here is s/img per device.}
    \vspace{-0.3cm}
        \begin{tabular}{l|ccc}
        \toprule
              & \multicolumn{2}{c}{PredCls} \\
              & mR@20/50/100  & testing  & training   \\
        \midrule
        \midrule
        MotifsNet (backbone) \cite{Neuralmotifs}  & $12.2 / 15.5 / 16.8$ & $0.11825$ & $0.11943$ \\
        \quad \textbf{RcSGG (ours)} &$32.6 / 38.8 / 41.4$ & $0.11792$   & $0.12522$   \\
        \bottomrule
        \midrule
        VCTree (backbone) \cite{VCtree}  & $12.4 / 15.4 / 16.6$ & $0.15410$ & $0.15526$ \\
        \quad \textbf{RcSGG (ours)} &$33.7 / 38.9 / 42.4$ & $0.15273$   & $0.16347$   \\
        \bottomrule
        \midrule
        Transformer (backbone) \cite{transformer}  & $12.4/16.0/17.5$& $0.18456$ & $0.18522$ \\
        \quad \textbf{RcSGG (ours)} &$32.4 / 39.6 / 41.3$ & $0.18498$   & $0.19570$   \\
        \bottomrule
        \end{tabular}%
    \label{inference_time}%
    \vspace{-0.3cm}
\end{table}

\textit{Efficiency Analysis.} 
This section evaluates RcSGG's computational efficiency by analyzing training and inference times, assessing the feasibility of integrating ARE and MIS into SGG frameworks (Table~\ref{inference_time}). The evaluation results show RcSGG increases training time by about 5\% over baseline methods, but this modest overhead leads to significant debiasing performance gains. We consider this trade-off justified given the substantial improvement in unbiased predictions. Despite the additional computational steps introduced by ARE and MIS during training, our method maintains high efficiency for several reasons. First, the query set $\mathcal{Q}$ is efficiently stored as a byproduct during the initial training epoch, incurring no extra computational cost. Second, we limit the size of the query set $\mathcal{Q}$ to approximately 20\% of the total relationship features, significantly reducing the sampling overhead. Finally, to enhance efficiency further, we optimize Maximum Mutual Information Sampling into Maximum Information Sampling, simplifying the process while preserving its effectiveness. These design choices ensure that the additional computational cost remains minimal, even as RcSGG performs sophisticated feature refinements to mitigate spurious correlations.

Unlike the training phase, the testing phase of RcSGG incurs no additional computational overhead. During testing, ARE is not applied, and the process follows the standard SGG flow ($X \rightarrow R \rightarrow Y$). The classifier operates within the refined parameter space $\mathcal{F}_{\widetilde{\mathcal{R}}}$, which was optimized during training to eliminate spurious correlations. Table~\ref{inference_time} demonstrates that the inference time per image for RcSGG is nearly identical to that of the baselines. Minor variations observed during testing are likely due to system noise or hardware-related fluctuations, rather than inherent differences in the computational requirements of the methods. This aligns with the theoretical expectation that RcSGG adds no additional steps to the inference process.


\subsection{Discussion and Future Work}
\label{section4.4}
As shown in Tables~\ref{mRK} and~\ref{RK}, our method achieves state-of-the-art results in terms of mR@K. However, there is a trade-off with R@K compared to baseline SGG frameworks. We argue this can be primarily attributed to two reasons: 1) By intervening in the relationship feature space, our method boosts the influence of tail categories on model training, preventing the model from predominantly converging to head categories, which inevitably impacts the performance of head relationships. 2) The SGG datasets contain noisy annotations, typically labeling fine-grained tail relationships as coarse-grained head relationships and including semantically ambiguous categories.

Due to space constraints, a thorough analysis of these points is provided in \textbf{Appendix F}. Importantly, the presence of noise in SGG datasets offers crucial insights for our future work, such as reprocessing dataset annotations; a preliminary outlook on this is also discussed in \textbf{Appendix F}.

\section{Conclusion}
\label{section5}
In this paper, we analyze how the decoupled two-stage SGG framework leads to spurious correlation issues, subsequently inducing at least two observable biases: head-tail bias and fore-back bias. To address these issues, we reconstruct the causal chain structure $X \rightarrow R \rightarrow Y$ of the SGG model into a reverse causal structure $X \rightarrow R \leftarrow Y$. By estimating the reverse causality, $R \leftarrow Y$, our restructured model mitigates spurious correlations, thereby eradicating the induced model biases. Furthermore, given the dual imbalance challenges in the SGG task, we propose a Maximum Information Sampling technique to enhance reverse causality estimation. Our method achieves state-of-the-art performance on the primary metric for debiased scene graphs, specifically the mean recall rate mR@K, while remaining competitive in recall rate R@K. This indicates that our method achieves a better trade-off, allowing for refined predictions in tail relationships without overly compromising the performance of head classifications. 


As a preliminary exploration for future work, in this paper, we analyze the semantic confusions and coarse-grained annotations in the SGG task and discuss potential solutions. We believe that addressing these aspects is crucial for achieving a deep understanding of scenes.

\ifCLASSOPTIONcaptionsoff
  \newpage
\fi

\bibliographystyle{IEEEtran}
\bibliography{biblio}

\begin{thebibliography}{10}
\providecommand{\url}[1]{#1}
\csname url@samestyle\endcsname
\providecommand{\newblock}{\relax}
\providecommand{\bibinfo}[2]{#2}
\providecommand{\BIBentrySTDinterwordspacing}{\spaceskip=0pt\relax}
\providecommand{\BIBentryALTinterwordstretchfactor}{4}
\providecommand{\BIBentryALTinterwordspacing}{\spaceskip=\fontdimen2\font plus
\BIBentryALTinterwordstretchfactor\fontdimen3\font minus \fontdimen4\font\relax}
\providecommand{\BIBforeignlanguage}[2]{{%
\expandafter\ifx\csname l@#1\endcsname\relax
\typeout{** WARNING: IEEEtran.bst: No hyphenation pattern has been}%
\typeout{** loaded for the language `#1'. Using the pattern for}%
\typeout{** the default language instead.}%
\else
\language=\csname l@#1\endcsname
\fi
#2}}
\providecommand{\BIBdecl}{\relax}
\BIBdecl

\bibitem{ImageRetrieval}
J.~Johnson, R.~Krishna, M.~Stark, L.-J. Li, D.~Shamma, M.~Bernstein, and L.~Fei-Fei, ``Image retrieval using scene graphs,'' in \emph{Proc. IEEE Conf. Comput. Vis. Pattern Recognit.}, 2015, pp. 3668--3678.

\bibitem{SGGSurvey}
X.~Chang, P.~Ren, P.~Xu, Z.~Li, X.~Chen, and A.~Hauptmann, ``A comprehensive survey of scene graphs: Generation and application,'' \emph{IEEE Trans. Pattern Anal. Mach. Intell.}, vol.~45, no.~1, pp. 1--26, 2023.

\bibitem{wang2020visual}
T.~Wang, J.~Huang, H.~Zhang, and Q.~Sun, ``Visual commonsense r-cnn,'' in \emph{Proc. IEEE Conf. Comput. Vis. Pattern Recognit.}, 2020, pp. 10\,760--10\,770.

\bibitem{chen2024scene}
G.~Chen, J.~Li, and W.~Wang, ``Scene graph generation with role-playing large language models,'' \emph{Int. Conf. Neural Inf. Process. Syst.}, 2024.

\bibitem{Neuralmotifs}
R.~Zellers, M.~Yatskar, S.~Thomson, and Y.~Choi, ``Neural motifs: Scene graph parsing with global context,'' in \emph{Proc. IEEE Conf. Comput. Vis. Pattern Recognit.}, 2018, pp. 5831--5840.

\bibitem{RCNNSGG}
J.~Yang, J.~Lu, S.~Lee, D.~Batra, and D.~Parikh, ``Graph r-cnn for scene graph generation,'' in \emph{Proc. Eur. Conf. Comput. Vis.}, 2018, pp. 670--685.

\bibitem{VG150}
D.~Xu, Y.~Zhu, C.~B. Choy, and L.~Fei-Fei, ``Scene graph generation by iterative message passing,'' in \emph{Proc. IEEE Conf. Comput. Vis. Pattern Recognit.}, 2017, pp. 5410--5419.

\bibitem{retrieval}
V.~Ramanathan, C.~Li, J.~Deng, W.~Han, Z.~Li, K.~Gu, Y.~Song, S.~Bengio, C.~Rosenberg, and L.~Fei-Fei, ``Learning semantic relationships for better action retrieval in images,'' in \emph{Proc. IEEE Conf. Comput. Vis. Pattern Recognit.}, 2015, pp. 1100--1109.

\bibitem{VQA1}
W.~Norcliffe-Brown, S.~Vafeias, and S.~Parisot, ``Learning conditioned graph structures for interpretable visual question answering,'' in \emph{Proc. 32nd Int. Conf. Neural Inf. Process. Syst.}, vol.~31, 2018.

\bibitem{VQA2}
J.~Shi, H.~Zhang, and J.~Li, ``Explainable and explicit visual reasoning over scene graphs,'' in \emph{Proc. IEEE Conf. Comput. Vis. Pattern Recognit.}, 2019, pp. 8376--8384.

\bibitem{imagecaptioning1}
X.~Yang, K.~Tang, H.~Zhang, and J.~Cai, ``Auto-encoding scene graphs for image captioning,'' in \emph{Proc. IEEE Conf. Comput. Vis. Pattern Recognit.}, 2019, pp. 10\,685--10\,694.

\bibitem{imagecaptioning2}
J.~Gu, S.~Joty, J.~Cai, H.~Zhao, X.~Yang, and G.~Wang, ``Unpaired image captioning via scene graph alignments,'' in \emph{Proc. IEEE/CVF Int. Conf. Comput. Vis.}, 2019, pp. 10\,323--10\,332.

\bibitem{SegG}
S.~Khandelwal, M.~Suhail, and L.~Sigal, ``Segmentation-grounded scene graph generation,'' in \emph{Proc. IEEE/CVF Int. Conf. Comput. Vis.}, 2021, pp. 15\,879--15\,889.

\bibitem{TransRwt}
A.~Zhang, Y.~Yao, Q.~Chen, W.~Ji, Z.~Liu, M.~Sun, and T.-S. Chua, ``Fine-grained scene graph generation with data transfer,'' in \emph{Proc. Eur. Conf. Comput. Vis.}, 2022.

\bibitem{FGPL}
X.~Lyu, L.~Gao, Y.~Guo, Z.~Zhao, H.~Huang, H.~T. Shen, and J.~Song, ``Fine-grained predicates learning for scene graph generation,'' in \emph{Proc. IEEE Conf. Comput. Vis. Pattern Recognit.}, 2022, pp. 19\,467--19\,475.

\bibitem{PPDL}
W.~Li, H.~Zhang, Q.~Bai, G.~Zhao, N.~Jiang, and X.~Yuan, ``{PPDL}: Predicate probability distribution based loss for unbiased scene graph generation,'' in \emph{Proc. IEEE Conf. Comput. Vis. Pattern Recognit.}, 2022, pp. 19\,447--19\,456.

\bibitem{GCL}
X.~Dong, T.~Gan, X.~Song, J.~Wu, Y.~Cheng, and L.~Nie, ``Stacked hybrid-attention and group collaborative learning for unbiased scene graph generation,'' in \emph{Proc. IEEE Conf. Comput. Vis. Pattern Recognit.}, 2022, pp. 19\,427--19\,436.

\bibitem{EBMloss}
M.~Suhail, A.~Mittal, B.~Siddiquie, C.~Broaddus, J.~Eledath, G.~Medioni, and L.~Sigal, ``Energy-based learning for scene graph generation,'' in \emph{Proc. IEEE Conf. Comput. Vis. Pattern Recognit.}, 2021, pp. 13\,936--13\,945.

\bibitem{DLFE}
M.-J. Chiou, H.~Ding, H.~Yan, C.~Wang, R.~Zimmermann, and J.~Feng, ``Recovering the unbiased scene graphs from the biased ones,'' in \emph{Proc. 29th ACM Int. Conf. Multi.}, 2021, pp. 1581--1590.

\bibitem{TDE}
K.~Tang, Y.~Niu, J.~Huang, J.~Shi, and H.~Zhang, ``Unbiased scene graph generation from biased training,'' in \emph{Proc. IEEE Conf. Comput. Vis. Pattern Recognit.}, 2020, pp. 3716--3725.

\bibitem{HML}
Y.~Deng, Y.~Li, Y.~Zhang, X.~Xiang, J.~Wang, J.~Chen, and J.~Ma, ``Hierarchical memory learning for fine-grained scene graph generation,'' in \emph{Proc. Eur. Conf. Comput. Vis.}, 2022.

\bibitem{NICE}
L.~Li, L.~Chen, Y.~Huang, Z.~Zhang, S.~Zhang, and J.~Xiao, ``The devil is in the labels: Noisy label correction for robust scene graph generation,'' in \emph{Proc. IEEE Conf. Comput. Vis. Pattern Recognit.}, 2022, pp. 18\,869--18\,878.

\bibitem{RTPB}
C.~Chen, Y.~Zhan, B.~Yu, L.~Liu, Y.~Luo, and B.~Du, ``Resistance training using prior bias: toward unbiased scene graph generation,'' in \emph{Proc. AAAI Conf. Artif. Intell.}, 2022.

\bibitem{BPLSA}
Y.~Guo, L.~Gao, X.~Wang, Y.~Hu, X.~Xu, X.~Lu, H.~T. Shen, and J.~Song, ``From general to specific: Informative scene graph generation via balance adjustment,'' in \emph{Proc. IEEE/CVF Int. Conf. Comput. Vis.}, 2021, pp. 16\,383--16\,392.

\bibitem{VCtree}
K.~Tang, H.~Zhang, B.~Wu, W.~Luo, and W.~Liu, ``Learning to compose dynamic tree structures for visual contexts,'' in \emph{Proc. IEEE Conf. Comput. Vis. Pattern Recognit.}, 2019, pp. 6619--6628.

\bibitem{fasterrcnn}
S.~Ren, K.~He, R.~Girshick, and J.~Sun, ``Faster {R-CNN}: Towards real-time object detection with region proposal networks,'' \emph{IEEE Trans. Pattern Anal. Mach. Intell.}, vol.~39, no.~6, pp. 1137--1149, 2017.

\bibitem{pearl2009causality}
{Pearl, Judea}, ``Causality,'' \emph{Cambridge university press}, 2009.

\bibitem{causalGeneralization}
R.~Christiansen, N.~Pfister, M.~E. Jakobsen, N.~Gnecco, and J.~Peters, ``A causal framework for distribution generalization,'' \emph{IEEE Trans. Pattern Anal. Mach. Intell.}, vol.~44, no.~10, pp. 6614--6630, 2022.

\bibitem{PearlPCH}
J.~Pearl, ``Causal diagrams for empirical research,'' \emph{Biometrika}, vol.~82, no.~4, pp. 669--688, 1995.

\bibitem{pearlCBM}
J.~Pearl \emph{et~al.}, ``Models, reasoning and inference,'' \emph{Cambridge, UK: CambridgeUniversityPress}, vol.~19, no.~2, 2000.

\bibitem{luo2020causal}
Y.~Luo, J.~Peng, and J.~Ma, ``When causal inference meets deep learning,'' \emph{Nat. Mach. Intell.}, vol.~2, no.~8, pp. 426--427, 2020.

\bibitem{Bengio2021Toward}
B.~Sch{\"o}lkopf, F.~Locatello, S.~Bauer, N.~R. Ke, N.~Kalchbrenner, A.~Goyal, and Y.~Bengio, ``Toward causal representation learning,'' \emph{Proc. of the IEEE}, vol. 109, no.~5, pp. 612--634, 2021.

\bibitem{CausalFairness}
D.~Plecko and E.~Bareinboim, ``Causal fairness analysis,'' \emph{arXiv:2207.11385}, 2022.

\bibitem{activelearning}
B.~Xie, L.~Yuan, S.~Li, C.~H. Liu, and X.~Cheng, ``Towards fewer annotations: Active learning via region impurity and prediction uncertainty for domain adaptive semantic segmentation,'' in \emph{Proc. IEEE Conf. Comput. Vis. Pattern Recognit.}, 2022, pp. 8068--8078.

\bibitem{AL_1}
O.~Ben-Eliezer, M.~Hopkins, C.~Yang, and H.~Yu, ``Active learning polynomial threshold functions,'' \emph{Int. Conf. Neural Inf. Process. Syst.}, 2022.

\bibitem{AL_2}
K.~Konyushkova, R.~Sznitman, and P.~Fua, ``Learning active learning from data,'' \emph{Int. Conf. Neural Inf. Process. Syst.}, vol.~30, 2017.

\bibitem{EDAL}
S.~Sun, S.~Zhi, J.~Heikkil{\"a}, and L.~Liu, ``Evidential uncertainty and diversity guided active learning for scene graph generation,'' in \emph{Proc. Int. Conf. Learn. Represent.}, 2023.

\bibitem{AlexNet}
A.~Krizhevsky, I.~Sutskever, and G.~E. Hinton, ``Imagenet classification with deep convolutional neural networks,'' in \emph{Int. Conf. Neural Inf. Process. Syst.}, vol.~25, 2012.

\bibitem{VGG}
K.~Simonyan and A.~Zisserman, ``Very deep convolutional networks for large-scale image recognition,'' in \emph{Proc. Int. Conf. Learn. Represent.}, 2014.

\bibitem{VG}
R.~Krishna, Y.~Zhu, O.~Groth, J.~Johnson, K.~Hata, J.~Kravitz, S.~Chen, Y.~Kalantidis, L.-J. Li, D.~A. Shamma \emph{et~al.}, ``Visual genome: Connecting language and vision using crowdsourced dense image annotations,'' \emph{Int. J. Comput. Vis.}, vol. 123, no.~1, pp. 32--73, 2017.

\bibitem{DT2}
A.~Desai, T.-Y. Wu, S.~Tripathi, and N.~Vasconcelos, ``Learning of visual relations: The devil is in the tails,'' in \emph{Proc. IEEE/CVF Int. Conf. Comput. Vis.}, 2021, pp. 15\,404--15\,413.

\bibitem{ML-MWN}
S.~Chen, Y.~Du, P.~Mettes, and C.~G. Snoek, ``Multi-label meta weighting for long-tailed dynamic scene graph generation,'' in \emph{Proc. ACM International Conference on Multimedia Retrieval}, 2023, pp. 39--47.

\bibitem{PENET}
C.~Zheng, X.~Lyu, L.~Gao, B.~Dai, and J.~Song, ``Prototype-based embedding network for scene graph generation,'' in \emph{Proc. IEEE Conf. Comput. Vis. Pattern Recognit.}, 2023, pp. 22\,783--22\,792.

\bibitem{DHL}
C.~Zheng, L.~Gao, X.~Lyu, P.~Zeng, A.~El~Saddik, and H.~T. Shen, ``Dual-branch hybrid learning network for unbiased scene graph generation,'' \emph{IEEE Trans. Circuits Syst. Video Technol.}, 2023.

\bibitem{MEET}
G.~Sudhakaran, D.~S. Dhami, K.~Kersting, and S.~Roth, ``Vision relation transformer for unbiased scene graph generation,'' in \emph{Proc. IEEE/CVF Int. Conf. Comput. Vis.}, 2023, pp. 21\,882--21\,893.

\bibitem{CaCao}
Q.~Yu, J.~Li, Y.~Wu, S.~Tang, W.~Ji, and Y.~Zhuang, ``Visually-prompted language model for fine-grained scene graph generation in an open world,'' in \emph{Proc. IEEE/CVF Int. Conf. Comput. Vis.}, 2023, pp. 21\,560--21\,571.

\bibitem{chen2023addressing}
G.~Chen, L.~Li, Y.~Luo, and J.~Xiao, ``Addressing predicate overlap in scene graph generation with semantic granularity controller,'' in \emph{Proc. IEEE Int. Conf. Multimedia Expo}.\hskip 1em plus 0.5em minus 0.4em\relax IEEE, 2023, pp. 78--83.

\bibitem{Wang}
X.~Wang, M.~Saxon, J.~Li, H.~Zhang, K.~Zhang, and W.~Y. Wang, ``Causal balancing for domain generalization,'' \emph{Proc. Int. Conf. Learn. Represent.}, 2023.

\bibitem{DisC}
S.~Fan, X.~Wang, Y.~Mo, C.~Shi, and J.~Tang, ``Debiasing graph neural networks via learning disentangled causal substructure,'' \emph{Int. Conf. Neural Inf. Process. Syst.}, 2022.

\bibitem{DSDI}
H.~Zhou, J.~Zhang, T.~Luo, Y.~Yang, and J.~Lei, ``Debiased scene graph generation for dual imbalance learning,'' \emph{IEEE Trans. Pattern Anal. Mach. Intell.}, vol.~45, no.~4, pp. 4274--4288, 2023.

\bibitem{TsCM}
S.~Sun, S.~Zhi, Q.~Liao, J.~Heikkil{\"a}, and L.~Liu, ``Unbiased scene graph generation via two-stage causal modeling,'' \emph{IEEE Trans. Pattern Anal. Mach. Intell.}, 2023.

\bibitem{lu2019bayes}
Y.~Lu, Y.-M. Cheung, and Y.~Y. Tang, ``Bayes imbalance impact index: A measure of class imbalanced data set for classification problem,'' \emph{IEEE Trans Neural Netw Learn Syst.}, vol.~31, no.~9, pp. 3525--3539, 2019.

\bibitem{fore-back}
Y.~Zhan, J.~Yu, T.~Yu, and D.~Tao, ``On exploring undetermined relationships for visual relationship detection,'' in \emph{Proc. IEEE Conf. Comput. Vis. Pattern Recognit.}, 2019, pp. 5128--5137.

\bibitem{SSRCNN-SGG}
Y.~Teng and L.~Wang, ``Structured sparse r-cnn for direct scene graph generation,'' in \emph{Proc. IEEE Conf. Comput. Vis. Pattern Recognit.}, 2022, pp. 19\,437--19\,446.

\bibitem{HetSGG}
K.~Yoon, K.~Kim, J.~Moon, and C.~Park, ``Unbiased heterogeneous scene graph generation with relation-aware message passing neural network,'' in \emph{Proc. AAAI Conf. Artif. Intell.}, vol.~37, no.~3, 2023, pp. 3285--3294.

\bibitem{SQUAT}
D.~Jung, S.~Kim, W.~H. Kim, and M.~Cho, ``Devil's on the edges: Selective quad attention for scene graph generation,'' in \emph{Proc. IEEE Conf. Comput. Vis. Pattern Recognit.}, 2023, pp. 18\,664--18\,674.

\bibitem{CV-SGG}
T.~Jin, F.~Guo, Q.~Meng, S.~Zhu, X.~Xi, W.~Wang, Z.~Mu, and W.~Song, ``Fast contextual scene graph generation with unbiased context augmentation,'' in \emph{Proc. IEEE Conf. Comput. Vis. Pattern Recognit.}, 2023, pp. 6302--6311.

\bibitem{FGPL-A}
X.~Lyu, L.~Gao, P.~Zeng, H.~T. Shen, and J.~Song, ``Adaptive fine-grained predicates learning for scene graph generation,'' \emph{IEEE Trans. Pattern Anal. Mach. Intell.}, 2023.

\bibitem{EICR}
Y.~Min, A.~Wu, and C.~Deng, ``Environment-invariant curriculum relation learning for fine-grained scene graph generation,'' in \emph{Proc. IEEE/CVF Int. Conf. Comput. Vis.}, 2023, pp. 13\,296--13\,307.

\bibitem{PSCV}
L.~Zhou, J.~Hu, Y.~Zhou, T.~L. Lam, and Y.~Xu, ``Peer learning for unbiased scene graph generation,'' \emph{arXiv:2301.00146}, 2023.

\bibitem{CFA}
L.~Li, G.~Chen, J.~Xiao, Y.~Yang, C.~Wang, and L.~Chen, ``Compositional feature augmentation for unbiased scene graph generation,'' in \emph{Proc. IEEE/CVF Int. Conf. Comput. Vis.}, 2023, pp. 21\,685--21\,695.

\bibitem{HiLo}
Z.~Zhou, M.~Shi, and H.~Caesar, ``Hilo: Exploiting high low frequency relations for unbiased panoptic scene graph generation,'' in \emph{Proc. IEEE/CVF Int. Conf. Comput. Vis.}, 2023, pp. 21\,637--21\,648.

\bibitem{NICEST}
L.~Li, L.~Chen, H.~Shi, H.~Zhang, Y.~Yang, W.~Liu, and J.~Xiao, ``Nicest: Noisy label correction and training for robust scene graph generation,'' \emph{arXiv:2207.13316}, 2022.

\bibitem{transformer}
A.~Vaswani, N.~Shazeer, N.~Parmar, J.~Uszkoreit, L.~Jones, A.~N. Gomez, {\L}.~Kaiser, and I.~Polosukhin, ``Attention is all you need,'' in \emph{Int. Conf. Neural Inf. Process. Syst.}, vol.~30, 2017.

\bibitem{GQAdataset}
D.~A. Hudson and C.~D. Manning, ``Gqa: A new dataset for real-world visual reasoning and compositional question answering,'' in \emph{Proc. IEEE Conf. Comput. Vis. Pattern Recognit.}, 2019, pp. 6700--6709.

\bibitem{openimages}
A.~Kuznetsova, H.~Rom, N.~Alldrin, J.~Uijlings, I.~Krasin, J.~Pont-Tuset, S.~Kamali, S.~Popov, M.~Malloci, A.~Kolesnikov \emph{et~al.}, ``The open images dataset v4: Unified image classification, object detection, and visual relationship detection at scale,'' \emph{Int. J. Comput. Vis.}, vol. 128, no.~7, pp. 1956--1981, 2020.

\bibitem{BGNN}
R.~Li, S.~Zhang, B.~Wan, and X.~He, ``Bipartite graph network with adaptive message passing for unbiased scene graph generation,'' in \emph{roc. IEEE Conf. Comput. Vis. Pattern Recognit.}, 2021, pp. 11\,109--11\,119.

\bibitem{PSG}
J.~Yang, Y.~Z. Ang, Z.~Guo, K.~Zhou, W.~Zhang, and Z.~Liu, ``Panoptic scene graph generation,'' in \emph{Proc. Eur. Conf. Comput. Vis.}, 2022, pp. 178--196.

\bibitem{resnet}
K.~He, X.~Zhang, S.~Ren, and J.~Sun, ``Deep residual learning for image recognition,'' in \emph{Proc. IEEE Conf. Comput. Vis. Pattern Recognit.}, 2016, pp. 770--778.

\bibitem{PCL}
L.~Tao, L.~Mi, N.~Li, X.~Cheng, Y.~Hu, and Z.~Chen, ``Predicate correlation learning for scene graph generation,'' \emph{IEEE Trans. Image Process.}, vol.~31, pp. 4173--4185, 2022.

\bibitem{Runet}
X.~Lin, C.~Ding, J.~Zhang, Y.~Zhan, and D.~Tao, ``Ru-net: Regularized unrolling network for scene graph generation,'' in \emph{Proc. IEEE Conf. Comput. Vis. Pattern Recognit.}, 2022, pp. 19\,457--19\,466.

\bibitem{LCS}
Y.~Shen, H.~Yun, Z.~C. Lipton, Y.~Kronrod, and A.~Anandkumar, ``Deep active learning for named entity recognition,'' \emph{arXiv preprint arXiv:1707.05928}, 2017.

\bibitem{MES}
W.~Luo, A.~Schwing, and R.~Urtasun, ``Latent structured active learning,'' in \emph{Int. Conf. Neural Inf. Process. Syst.}, vol.~26, 2013.

\bibitem{MS}
M.-F. Balcan, A.~Broder, and T.~Zhang, ``Margin based active learning,'' in \emph{International Conference on Computational Learning Theory}.\hskip 1em plus 0.5em minus 0.4em\relax Springer, 2007, pp. 35--50.

\bibitem{ADTrans}
L.~Li, W.~Ji, Y.~Wu, M.~Li, Y.~Qin, L.~Wei, and R.~Zimmermann, ``Panoptic scene graph generation with semantics-prototype learning,'' in \emph{Proc. AAAI Conf. Artif. Intell.}, vol.~38, no.~4, 2024, pp. 3145--3153.

\bibitem{SSR-CNN}
Y.~Teng and L.~Wang, ``Structured sparse r-cnn for direct scene graph generation,'' in \emph{Proc. IEEE Conf. Comput. Vis. Pattern Recognit.}, 2022, pp. 19\,437--19\,446.

\bibitem{EGTR}
J.~Im, J.~Nam, N.~Park, H.~Lee, and S.~Park, ``Egtr: Extracting graph from transformer for scene graph generation,'' in \emph{Proc. IEEE Conf. Comput. Vis. Pattern Recognit.}, 2024, pp. 24\,229--24\,238.

\bibitem{ST-SGG}
K.~Kim, K.~Yoon, Y.~In, J.~Moon, D.~Kim, and C.~Park, ``Adaptive self-training framework for fine-grained scene graph generation,'' in \emph{Proc. Int. Conf. Learn. Represent.}, 2020, pp. 1--25.

\bibitem{logitadjustment}
A.~K. Menon, S.~Jayasumana, A.~S. Rawat, H.~Jain, A.~Veit, and S.~Kumar, ``Long-tail learning via logit adjustment,'' in \emph{Proc. Int. Conf. Learn. Represent.}, 2020, pp. 1--27.

\end{thebibliography}

\clearpage

\appendices

\section*{Appendix file to ‘A Reverse Causal Framework to Mitigate Spurious Correlations for Debiasing Scene Graph Generation’}

This Appendix provides detailed theoretical proofs for the assumptions and theorems presented in the main paper. Additionally, further analysis and discussions on future work are included to offer a more comprehensive understanding of the topics covered. Unless otherwise noted, the notations and symbols used in this Appendix are consistent with those in the main paper, ensuring continuity and clarity in the mathematical discourse.

\section*{APPENDIX A: Proof of Spurious Correlations Between $X$ and $Y$ in SGG Task}

In this section, we provide the detailed mathematical proof that demonstrates the existence of spurious correlations between $X$ and $Y$ in the two-stage Scene Graph Generation (SGG) framework \cite{NICE,Neuralmotifs,VCtree,HML,FGPL,TransRwt,HiLo,MEET,PSCV}.

Specifically, we assume that the aggregated relationship features at the image level result from an unobserved confounder $Z$, with $X=a_1 Z+\epsilon_1$ and $Y=a_2 Z+\epsilon_2$, where $a_1$ and $a_2$ are constants, and $\epsilon_1$ and $\epsilon_2$ represent random noise terms. The Pearson correlation coefficient between X and Y is given by: 
\begin{equation}
\rho(X, Y)=\frac{\operatorname{Cov}(X, Y)}{\sqrt{\operatorname{Var}(X) \times \operatorname{Var}(Y)}},
\end{equation}
where 
\begin{equation}
\begin{aligned}
\operatorname{Cov}(X, Y) & = E[(a_1 Z+\epsilon_1)(a_2 Z+\epsilon_2)]-E[a_1 Z+\epsilon_1] E[a_2 Z+\epsilon_2] \\
& =a_1 a_2 E[Z^2]+a_1 E[Z \epsilon_2]+a_2 E[\epsilon_1 Z]+ \\ 
& \ \ \quad E[\epsilon_1 \epsilon_2] -a_1 a_2 E[Z]^2 \\
& = a_1 a_2 E[Z^2]-a_1 a_2 E[Z]^2 .
\end{aligned}
\end{equation}
We, therefore, have:
\begin{equation}
\rho(X, Y)=\frac{a_1 a_2 E[Z^2]-a_1 a_2 E[Z]^2}{\sqrt{({a_1}^2 E[Z^2]+\operatorname{Var}(\epsilon_1))({a_2}^2 E[Z^2]+\operatorname{Var}(\epsilon_2))}}.
\end{equation}
For the SGG task, we observe a strong co-occurrence of relationships at the image level, as illustrated in Figure 3 (b) in the main paper. For instance, the probability of "\textit{growing on}" co-occurring with "\textit{mounted on}" is much higher than with "\textit{flying on}" in a scene. Therefore, $E[Z^2] \neq E[Z]^2$, leading to $\rho(X, Y) \neq 0$. Yet, $f_c$ expects independent relationship features. Hence, we argue that spurious correlations exist between $X$ and $Y$.

\section*{APPENDIX B: Mathematical Foundations of Active Reverse Estimation (ARE)}

These section clarify how the effective parameter space $\mathcal{F}_{\widetilde{\mathbf{R}}}$, Spurious Correlation Error (SCE), Overlapping Error (OE), and their interplay enable the Bayesian optimal classifier $f_c^*$ to mitigate spurious correlations. The effective parameter space $\mathcal{F}_{\widetilde{\mathbf{R}}}$ is defined as:
\begin{equation}
\mathcal{F}_{\widetilde{\mathbf{R}}}=\{f_c:|SCE(f_c) -OE(f_c)| \leq \delta\},
\end{equation}
where $SCE(f_c)$ quantifies the errors caused by spurious correlations in the parameter space, $OE(f_c)$ measures the overlap between spurious correlations and residual classification errors, and $\delta$ is a threshold that balances spurious correlation mitigation with model flexibility. By constraining the difference between $SCE$ and $OE$, $\mathcal{F}_{\widetilde{\mathbf{R}}}$ ensures that the classifier focuses on causal relationships while reducing the influence of spurious correlations. The Spurious Correlation Error is calculated as:
\begin{align}
SCE(f_c) &= \sum_{k \in K} \int P(Y=k|X) \nonumber \\
&\times [1 - P(\hat{Y}=k|Y=k, X)] P(X) dX,
\end{align}
where $P(Y=k|X)$ represents the true label probability given input $X$, and $P(\hat{Y}=k|Y=k, X)$ is the model's predicted probability conditioned on the true label and input. This metric evaluates the discrepancies between true and predicted distributions under the influence of $X$.

The Overlapping Error measures the intersection of spurious correlations and causal signals:
\begin{align}
OE(f_c) &= \sum_{k \in K} \iint P(Y=k|R, X) \nonumber \\
&\times [1 - P(\hat{Y}=k|Y=k, R, X)] P(R, X) dR dX.
\end{align}
where $P(Y=k|R, X)$ denotes the conditional probability given relationship features $R$ and input $X$, and $P(\hat{Y}=k|Y=k, R, X)$ is the predicted probability conditioned on $R$, $X$, and the true labels. Minimizing $OE$ ensures that the parameter space prioritizes causal signals over spurious correlations.

To achieve the Bayesian optimal classifier $f_c^*$, the effective parameter space $\mathcal{F}_{\widetilde{\mathbf{R}}}$ must be optimized to minimize the residual classification error (RCE):
\begin{equation}
f_c^*=\operatorname{argmin}_{f_c \in \mathcal{F}_{\widetilde{\mathbf{R}}}} \ RCE(f_c),
\label{f_c^*_}
\end{equation}
This optimization aligns the model’s focus with causal patterns, mitigating the adverse effects of spurious correlations.

Active Reverse Estimation (ARE) is proposed to operationalize this theoretical framework. ARE estimates the reverse causality $R \leftarrow Y$ by designing query conditions and actively sampling high-information relationship features. This ensures that the classifier $f_c$ operates within $\mathcal{F}_{\widetilde{\mathbf{R}}}$, minimizing the impact of spurious correlations while enhancing the model's generalizability. Through ARE, the effective parameter space is aligned with causality, ultimately enabling robust and unbiased learning in SGG tasks.

\section*{APPENDIX C: Proof of Assumptions 1 and 2}
This section provides the detailed proofs of Assumptions 1 and 2, which form the theoretical basis of our analysis in the main paper.

\textbf{Assumption 1} (Spurious correlations in a single batch).
\textit{For any relationship feature $\mathbf{r}_{tb}^p$ in batch $\mathbf{B}_{tb}$, its outcome is influenced by another feature $\mathbf{r}_{tb}^q$ in the same batch. Formally, $Y(\mathbf{r}_{tb}^p) \neq Y(\mathbf{r}_{tb}^p \cup \mathbf{r}_{tb}^q)$, where $Y(\cdot)$ represents the prediction of the first instance in the given feature combination.}

\textbf{Assumption 2} (No spurious correlations across batches).
\textit{For any relationship feature $\mathbf{r}_{tb}^p$ in batch $\mathbf{B}_{tb}$, its outcome is unaffected by any feature $\mathbf{r}_{m n}^q$ from a different batch $\mathbf{B}_{m n}$. Formally, $Y(\mathbf{r}_{tb}^p) = Y(\mathbf{r}_{tb}^p \cup \mathbf{r}_{m n}^q)$, where $b \neq m$.}

\textbf{Proof of Assumptions 1.} We assume that the conditional probability output of the model is:
\begin{equation}
Y(X; \mathbf{\theta}) = \text{softmax}(\mathbf{W}X + \beta), 
\end{equation}
where $\mathbf{W}$ and $\beta$ are the model's weights and bias, respectively. Given that SGG is a classification task, we typically supervise the model's training with cross-entropy loss: 
\begin{equation}
L_{f_c}(\mathbf{B}_{tb} ; \theta)=-\frac{1}{|\mathbf{B}_{tb}|} \sum_{(X, Y) \in \mathbf{B}_{tb}} \log \frac{e^{(\mathbf{w}^{\top} X+b)}}{\sum_{k=1}^K e^{(\mathbf{w}^{\top} X +b)}},
\label{eq:loss}
\end{equation}
where $|\mathbf{B}_{tb}|$ is the size of batch $\mathbf{B}_{tb}$. The update of model parameters depends on the value of the loss function $L$; thus, for the claim $Y(\mathbf{r}_{tb}^p) \neq Y(\mathbf{r}_{tb}^p \cup \mathbf{r}_{tb}^q)$ in Assumption 1, we only need to prove $L_{f_c}(\mathbf{B}_{tb}; \mathbf{\theta}) \neq  L_{f_c}(\mathbf{B}_{tb} \backslash {\mathbf{r}_{tb}^q}; \mathbf{\theta})$. $\mathbf{B}_{tb} \backslash {\mathbf{r}_{tb}^q}$ signifies the removal of the $q$-th relationship category from batch $\mathbf{B}_{tb}$. Intuitively, assume $\mathbf{r}_{tb}$ is a head relationship with multiple instances in $\mathbf{B}_{tb}$; removing it would naturally enhance the learning of other categories, particularly tail categories. Even if $\mathbf{r}_{tb}$ has only a few instances in $\mathbf{B}_{tb}$, $L_{f_c}(\mathbf{B}_{tb} \backslash {\mathbf{r}_{tb}^q})$ would still change, thereby influencing parameter updates. Based on Equation (\ref{eq:loss}), we have:
\begin{equation}
\begin{aligned}
L_{f_c}(\mathbf{B}_{tb} \backslash {\mathbf{r}_{tb}^q} ; \theta)=-\frac{1}{|\mathbf{B}_{tb}| - |\mathbf{r}_{tb}^q|} & \sum_{(X, Y)  \in \mathbf{B}_{tb}}  ( \log \frac{e^{(\mathbf{w}^{\top} X+\beta)}}{\sum_{k=1}^K e^{(\mathbf{w}^{\top} X+\beta)}} \\
& -\log \frac{e^{(\mathbf{w}^{\top} \mathbf{r}_{tb}^q+\beta)}}{\sum_{k=1}^K e^{(\mathbf{w}^{\top} \mathbf{r}_{tb}^q+\beta)}}).
\label{eq:loss-}
\end{aligned}
\end{equation}
where $|\mathbf{r}_{tb}^q|$ denotes the number of the $q$-th relationship category in batch $\mathbf{B}_{tb}$. Regarding Equation (\ref{eq:loss-}), we see that it excludes components related to $\mathbf{r}_{tb}^q$ from the loss function of the entire batch $\mathbf{B}_{tb}$, altering the total sum of the loss function. Additionally, it is important to note that, compared to Equation (\ref{eq:loss}), the denominator changes from $|\mathbf{B}_{tb}|$ to $|\mathbf{B}_{tb}| - |\mathbf{r}_{tb}^q|$, affecting the average value of the loss. Therefore, the values of $L_{f_c}(\mathbf{B}_{tb} ; \theta)$ and  $L_{f_c}(\mathbf{B}_{tb} \backslash {\mathbf{r}_{tb}^q} ; \theta)$ cannot be the same unless $\mathbf{r}_{tb}^q$ completely offsets the exclusion terms, which is impractical as it would imply that $\mathbf{r}_{tb}^q$ is an infinitesimally small quantity. Thus, we arrive at the proof of Assumption 1.

\textbf{Proof of Assumptions 2.} Since the loss function for each batch is calculated independently, for $\mathbf{B}_{tb}$, we have $L_{f_c}(\mathbf{B}_{tb} ; \theta)$, as seen in Equation (\ref{eq:loss}); for $\mathbf{B}_{m n}$, we have:
\begin{equation}
L_{f_c}(\mathbf{B}_{m n} ; \theta)=-\frac{1}{|\mathbf{B}_{m n}|} \sum_{(X, Y) \in \mathbf{B}_{m n}} \log \frac{e^{(\mathbf{w}^{\top} X+\beta)}}{\sum_{k=1}^K e^{(\mathbf{w}^{\top} X +\beta)}},
\label{eq:loss_mn}
\end{equation}
Gradient updates for $\mathbf{B}_{tb}$ and $\mathbf{B}_{m n}$ are conducted separately, and the update rules ensure that data within each batch only affects its own parameter updates. Thus, the update of $\mathbf{B}_{tb}$ does not impact the loss calculation and parameter update for $\mathbf{B}_{m n}$, and vice versa. This leads us to the proof of Assumption 2.

\section*{APPENDIX D: Proof of Theorem 1}
This section provides the detailed proof for Theorem 1, which states that the model $f_c$ trained on a category-balanced dataset $\mathcal{D}$ is a Bayesian optimal classifier \cite{Wang,logitadjustment}, unaffected by spurious correlations.

\textbf{Theorem 1} (Bayesian optimal classifier induced by epoch balance). \textit {The model $f_c$ trained on $\mathcal{D}$ is a Bayesian optimal classifier $f_c^*$, \ie $\operatorname{argmax}_{y \in[K]} f_c(\mathbf{r})=\operatorname{argmax}_{y \in[K]} f_c^*(\mathbf{r})$, if the following two conditions hold: 1) $N_{\mathcal{D}} = N_{\mathbf{B}}$, 2) $|\mathbf{r}^{m}| = |\mathbf{r}^{n}|$ ($m \in [K], n \in [K], m \neq n$), where $\mathbf{r}^{m}$ is the feature of $m$-th relationship class in $\mathcal{D}$.} 

\textbf{Proof of Theorem 1.} Given that $N_{\mathcal{D}} = N_{\mathbf{B}}$, the loss $L$ for each epoch is: 
\begin{equation}
L_{f_c}(\mathcal{D} ; \theta)=-\frac{2}{N_{\mathcal{D}}(N_{\mathcal{D}}+1)} \sum_{i=1}^{N_{\mathcal{D}}} \sum_{j=1}^{N_i} \hat{Y}(\mathbf{r}_{tb}) \log (Y(\mathbf{r}_{tb})).
\end{equation}
Wherein, $\hat{Y}(\mathbf{r}_{tb})$ is the ground truth label for the $j$-th relationship feature in image $\mathbf{x}_{i}$. Then, given that $|\mathbf{R}_{im}|=|\mathbf{R}_{in}|$, we have: 
\begin{equation}
\mathbb{E}_{(\mathbf{r}^k,\hat{Y}(\mathbf{y}^k)) \sim \mathcal{D}} [L_{f_c}(\mathcal{D} ; \theta)] = \frac{1}{K} \sum_{k=1}^{K} \mathbb{E}_{(r,\hat{Y}(\mathbf{y}^k)=k) \sim \mathcal{D}} [-\log f_c(\mathbf{r}^k)],
\end{equation}
where $\mathbf{r}^k$ represents the relationship features of the $k$-th relationship category in $\mathcal{D}$, with $\hat{Y}(\mathbf{y}^k)$ and $f_c(\mathbf{r}^k)$ denoting its label and predicted probability, respectively. Under ideal training condition \renewcommand{\thefootnote}{\fnsymbol{footnote}} \footnote[4]{Ideal training condition refers to the optimizer used for training that can guide the model in reaching the global optimum. Unless specifically stated otherwise, all assumptions and theorems in this paper are based on this condition.}, it is easy to deduce that $\forall \epsilon>0$, $\exists N$ such that $\forall \text { epoch }>N$,
\begin{equation}
|L_{f_c}(\mathcal{D} ; \theta)-\min_{f_c} \mathbb{E}_{\hat{Y}(\mathbf{y}^k)|r} [-\log f_c(\mathbf{r}^k)]|<\epsilon. 
\end{equation}
Note that $\min_{f_c} \mathbb{E}_{\hat{Y}(\mathbf{y}^k)|r} [-\log f_c(\mathbf{r}^k)]$ minimizes the true expected loss, thus corresponding to the loss of the Bayesian optimal classifier $f_c^*$. As a result, Theorem 1 can be achieved.

\section*{APPENDIX E: Proof of Theorem 3}

Here we provide a detailed proof to demonstrate the validity of Theorem 3. Theorem 3 extends the idea presented in Theorem 2 by addressing the scenario where batch balance is only approximate rather than exact. This is a more realistic condition, as achieving perfect batch balance is often impractical due to the inherent distributional imbalances in real-world datasets.

\textbf{Theorem 3} (Bayesian optimal classifier induced by approximate batch balance). \textit {The model $f_c$ trained on $\mathcal{D}$ is a Bayesian optimal classifier, \ie $\operatorname{argmax}_{y \in[K]} f_c(\mathbf{r})=\operatorname{argmax}_{y \in[K]} f_c^*(\mathbf{r})$, if the following two conditions hold: 1) $N_{\mathcal{D}}/N_{\mathbf{B}} \in \{2, 3, 4 \cdots \}$, 2) $|\mathbf{r}^{bm}| \approx |\mathbf{r}^{bn}|$ ($b \in [N_{\mathcal{D}}/N_{\mathbf{B}}], m \in [K], n \in [K], m \neq n$).}

In the case of approximate batch balance, \ie, $|\mathbf{r}^{i m}| \approx |\mathbf{r}^{i n}|$, we posit the existence of an upper limit $\delta$ such that: 
\begin{equation}
|\nabla_\theta L_{f_c}(\mathbf{B}_i ; \theta) - \nabla_\theta L_{f_c^*}(\mathbf{B}_i ; \theta)| \leq \delta .
\end{equation} 
Throughout multiple batch training, these deviations accumulate. However, since they are bounded, the model's updates still approximately align in the direction of $f_c^*$. Considering the accumulated error after $n$ iterations: 
\begin{equation}
\sum_{i=1}^{n} |\nabla_\theta L_{f_c^i}(\mathbf{B}_i ; \theta) - \nabla_\theta L_{f_c^*}(\mathbf{B}_i ; \theta)| \leq n\delta .
\end{equation} 
It can be observed that as $n$ increases, $f_c$ tends to converge infinitely close to $f_c^*$; therefore, $f_c$ is a Bayesian optimal classifier in such case.

\section*{APPENDIX F: Detailed Discussion and Future Work}

\begin{figure}[ht]
	\footnotesize\centering
	\centerline{\includegraphics[width=0.9\linewidth]{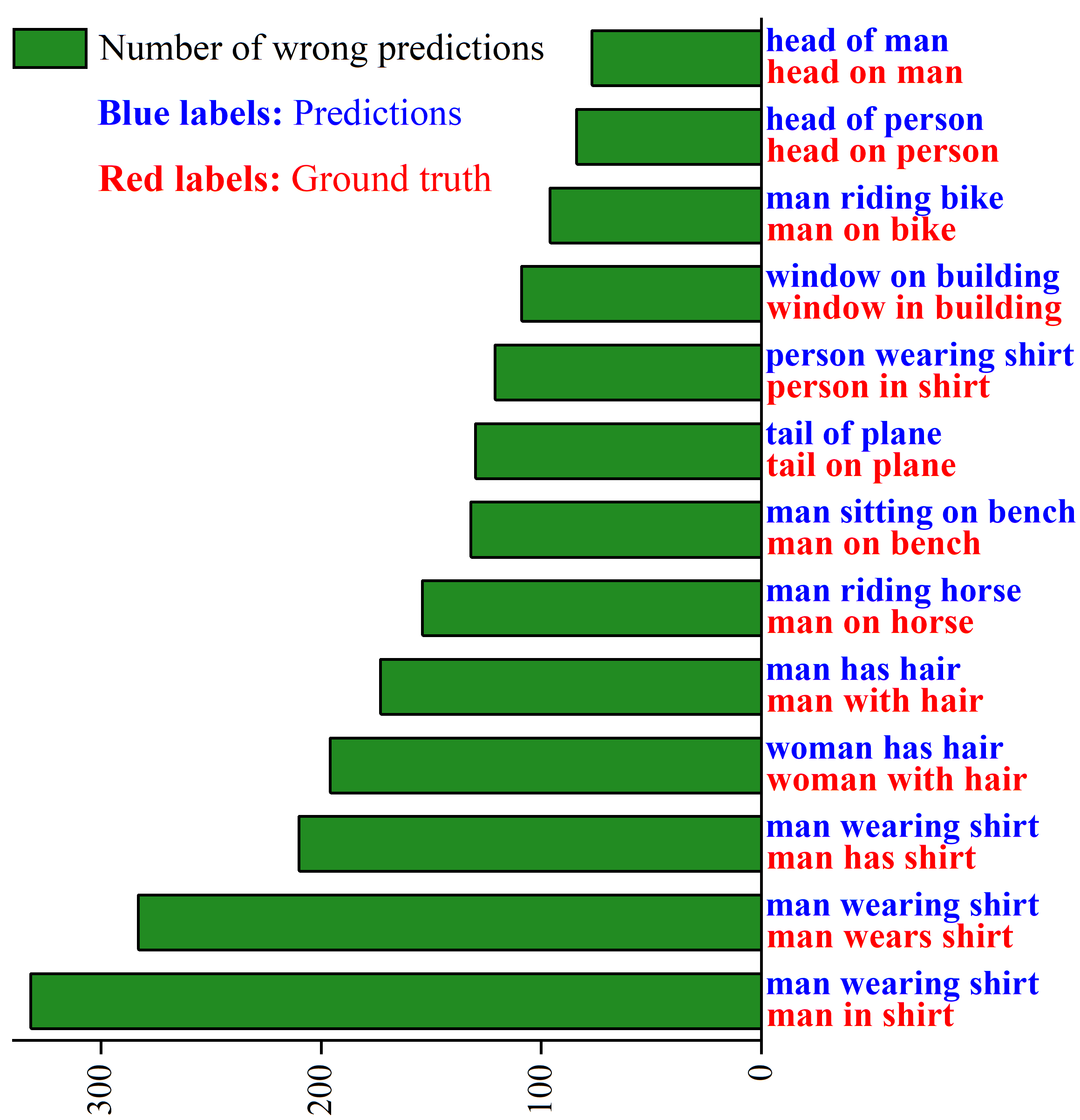}}
    \caption{The statistics of our method's wrong predictions on the test set. In the visualization, the red relationships signify the actual labels, while the blue ones represent the predictions. For example, the first bar indicates that our method predicted \textcolor{red}{\textit{\textless man, \textbf{in}, shirt\textgreater}} as \textcolor{blue}{\textit{\textless man, \textbf{wearing}, shirt\textgreater}}.}
	\label{feature_work}
\end{figure}

As depicted in Table 1 and 2 in the main text, our method achieves state-of-the-art results in terms of mR@K. However, compared to the baseline SGG frameworks, there is a sacrifice in R@K. We argue this can be primarily attributed to two reasons:

1) \textbf{Enhanced tail relationships.} By intervening in the relationship feature space, our method enhances the impact of tail categories on model training, thereby preventing the model from predominantly converging to head categories. Given this, we are not surprised by a decline in the model's performance for a handful of head categories. Considering that these categories collectively represent nearly $50\%$ of the total amount, even minor perturbations in these relationships can have a significant influence on the recall rate.

2) \textbf{Semantic confusion.} Some incorrect predictions arise from semantic confusion. For example, predicting \textit{\textless man, \textbf{wears}, shirt\textgreater} as \textit{\textless man, \textbf{wearing}, shirt\textgreater}. While this prediction is marked as incorrect, it is important to note that having \textit{wears}'' and \textit{wearing}'' as separate labels might be questionable since both could represent the same relationship feature. The choice between them could merely reflect annotator preference. This observation highlights a limitation of our method: it does not explicitly account for such synonymous labels, which may lead to mismatches with the original annotations. However, we argue that this issue stems primarily from the dataset itself rather than being an inherent flaw of our approach. Addressing this will require a more refined dataset annotation strategy in future work.

3) \textbf{Annotation granularity.} Our method demonstrates a strong ability to address noisy labels by predicting fine-grained tail relationships with higher information content. For instance, it tends to predict \textit{\textless man, \textbf{sitting on}, bench\textgreater} instead of the coarser \textit{\textless man, \textbf{on}, bench\textgreater}. This behavior directly mitigates the common noise issue in SGG datasets, as highlighted in \cite{NICE,Neuralmotifs}, where tail relationships with greater semantic specificity are often annotated as coarse-grained head relationships. By intervening in the relationship feature space, our method counteracts this tendency, effectively redistributing model attention towards more detailed and semantically rich relationships. This ability to predict fine-grained relationships reflects our method's robustness against annotation noise, as it prioritizes semantically accurate predictions over strict adherence to potentially ambiguous or imprecise annotations. Although this approach may lead to a decrease in R@K performance due to mismatched labels, it significantly enhances the model's capacity to capture nuanced relationships, as evidenced by the higher mR@K performance.

Inspired by the experiments presented in Figure \ref{feature_work}, our future endeavors will emphasize addressing the noise in the SGG dataset, particularly semantic confusion and ambiguous labels. For semantic confusion, we believe that label consolidation or multi-label prediction could serve as potential solutions. For ambiguous labels, constructing a hierarchical label representation appears promising. For instance, treating \textit{laying on}, \textit{standing on}, \textit{sitting on}, and \textit{walking on} as sub-labels of \textit{on}.

Moreover, the presence of noisy labels in the SGG dataset causes R@K and mR@K to exhibit an inversely proportional relationship. However, as detailed in Section 4.2 of the main text, by intervening in the background relationships and incorporating their associated outputs into evaluation, our approach outperforms the baseline for both R@K and mR@K. This improvement across dual metrics is indeed encouraging, and it provides valuable insights for our subsequent research endeavors.

\end{document}